%% file: acl_latex.tex
\pdfoutput=1
\documentclass[11pt]{article}

\usepackage[preprint]{acl}

\usepackage{times}
\usepackage{latexsym}

\usepackage[T1]{fontenc}

\usepackage[utf8]{inputenc}

\usepackage{microtype}

\usepackage{inconsolata}

\usepackage{graphicx}
\usepackage{booktabs}
\usepackage{multirow}
\usepackage{xcolor}
\usepackage{enumitem}
\usepackage[norule]{footmisc}
\newcommand{\tablefont}{\small}
\usepackage[most]{tcolorbox}
\definecolor{stageone}{HTML}{0072B2}
\definecolor{stagetwo}{HTML}{009E73}
\definecolor{stagealt}{HTML}{D55E00}

\usepackage{amsmath}

\title{Unintended Side Effects of Value Training in LLMs}

\author{
  Siddhesh Pawar$^{\spadesuit *}$ \quad Chen Cecilia Liu$^{\heartsuit}$ \quad Anna Korhonen$^{\heartsuit}$ \quad Isabelle Augenstein$^{\spadesuit}$ \\[0.5em]
  $^{\spadesuit}$University of Copenhagen \quad $^{\heartsuit}$University of Cambridge \\
  \texttt{\{sipa, augenstein\}@di.ku.dk} \quad \texttt{\{cl2002, alk23\}@cam.ac.uk} \\[0.3em]
  {\small $^{*}$Work done while at the University of Cambridge.}
}

\begin{document}
\maketitle
\setcounter{footnote}{0}
\begin{abstract}
As LLMs are increasingly deployed in value-sensitive contexts, aligning them with human values has emerged as a post-training objective in NLP, aimed at instilling additional values and guiding model behaviour. However, current evaluations remain narrow. Because values are interconnected, it is unclear how fine-tuning for specific values alters unrelated capabilities and behaviours. In this work, we adapt four instruction-tuned LLMs across individual Hofstede's value dimensions and evaluate them on both cross-value transfer and broader behaviours. We find that value-specific training produces unintended side effects on untrained values, risk preferences, social cognition, and safety traits. Consequently, we suggest that value alignment after post-training should incorporate broader behavioral evaluations, and we highlight the need to investigate where additional value training might best fit into the pipeline. We release all 48 value-tuned model checkpoints.\footnote{\url{https://huggingface.co/sidicity/cultural-value-models}}

\end{abstract}

\input{01-Introduction}
\input{02-Related-Works}
\input{04-Methodology}
\input{05-Evaluation}
\input{06-Discussion}
\input{06a-Safety-Relevant-Evals}
\input{06b-Social-Cognition}

\input{07-Conclusion}

\bibliography{custom}
\appendix
\input{08-Appendix}

\end{document}

%% file: 01-Introduction.tex
\section{Introduction}
Large language models (LLMs) are increasingly deployed as conversational assistants. However, rather than using them solely for routine task execution, users increasingly turn to them for career coaching, emotional support, and other personal matters that have traditionally required human social interaction \cite{NBERw34255, anthropic2025affective}. 
A system that does not support diverse and prosocial values could cause negative societal effects, as humans could also be shaped by the values the assistant reflects~\citep{jakesch2023cowriting, williamsceci2026biased}.

Recognizing this, post-training paradigms have commonly been adopted to improve models' reflection of diverse values, including safety-based constellations of values \cite{bai2022constitutional, guan2024deliberative}, character-based behavioral tendencies~\citep{anthropic2024character, maiya2025opencharacter}, or survey-based cultural values~\citep{li2024culturepark, feng-etal-2025-culfit}. These directions are largely studied in isolation. Yet values may not be separable inside a model where training one could shift others.

Recent work shows that models fine-tuned to write insecure code become broadly misaligned in unrelated domains~\citep{betley2025emergent}. In that setting, the training signal is not explicitly value-laden (though intentions remain inferable), and the discovery is incidental (i.e., \emph{emergent misalignment}, unintended side effects arising from the training objective). Meanwhile, survey-based value training explicitly targets specific values and checks for standard capability loss rather than examining what else shifts \cite{li2024culturellm, xu-etal-2025-culturespa}.  Neither tells us what else could change when a value is deliberately installed.

In this work, we focus on Hofstede-based value-tuning and examine the influence of training beyond the typical evaluation of the target value itself. We fine-tuned four instruction-tuned LLMs on each direction of Hofstede's value dimensions \citep{hofstede2001culture,minkov2012hofstede}. We investigate emergent misalignment from two aspects: first, whether training on one value shifts others, and second, its impact on broader behaviors related to safety, risk-sensitive decision-making, and social cognition such as theory-of-mind \cite{premack1978does} or emotion intelligence.

In summary, our \textbf{contributions} are as follows:
\begin{itemize}[nosep]
\item We find that training in one value direction also shifts the model's preferences on the other, untrained value dimensions.
\item We show that training drifts safety-relevant behavior, including self-reported traits, value prioritization in dilemmas, and open-ended responses judged more often as misaligned.
\item We further show that models' social cognition degrades in application, not comprehension. 
\end{itemize}

Our findings suggest that Hofstede-based value tuning and existing alignment interact in ways that sequential pipelines do not account for. Applying value tuning after safety post-training simultaneously shifts deception, oversight-related behavior, and social cognition, with effects that differ across base models and remain undetected by standard value benchmarks. These effects reflect how value signals propagate through training rather than intrinsic properties of the dimensions described. This highlights the need for broader evaluation of late-stage value tuning and raises the fundamental question of whether such intervention belongs earlier in the training pipeline.

%% file: 02-Related-Works.tex
\section{Related Work}
\label{sec:related}

\paragraph{Value Training.}
Recent NLP research typically evaluates language models' values through social science surveys and seeks to improve the models' alignment. For evaluation, survey instruments mapped to Hofstede's \citep{hofstede2001culture, minkov2012hofstede} cultural dimensions are administered to models \citep{cao-etal-2023-assessing, tao2023cultural, alkhamissi2024investigating, masoud-etal-2025-cultural, negru2026sociocultural}. For training, prior work leverages surveys alongside synthetic data to steer value alignment, evaluating progress using in-domain surveys or relevant downstream tasks \citep[inter alia]{li2024culturellm, li2024culturepark, adilazuarda-etal-2025-surveys}. In broader value training, Constitutional AI \cite{bai2022constitutional, DBLP:conf/fat/HuangSLLDTG24} trains model with high-level principles, where as other approaches learning values implicitly from human preferences through feedback learning \cite{bai2022rlhf, DBLP:conf/nips/Ouyang0JAWMZASR22, DBLP:conf/nips/RafailovSMMEF23}, and increasingly aimed to accommodate diverse user values \cite{zhou-etal-2024-beyond, DBLP:conf/icml/SorensenMFGMRYJ24,DBLP:conf/nips/KirkWRBMGCBW0VH24}. Character training shapes the assistant's persona as a whole, which governs not only its surface behavior but also its apparent values, beliefs, and ethics \citep{anthropic2024character, maiya2025opencharacter}. These methods typically evaluate and train models on the targeted values themselves, overlooking how values generalize to model behavior beyond the training domain.

\paragraph{Emergent misalignment and the side effects of narrow fine-tuning.}
Recent work \cite{betley2025emergent} found that fine-tuning on a narrow domain, such as writing insecure code, can induce broadly misaligned behavior in unrelated domains. This effect has also been shown in other narrow training (e.g., finance or medical advice), and under reinforcement learning \citep{ em/abs-2506-11613,  em/ICLR2026_50db99ee}, though susceptibility varies considerably across models \citep{em/corr/abs-2605-12199}.

Closely related to our work, \citet{choi-etal-2025-unintended} found that aligning a Llama-2 7B model with Schwartz values increases toxicity, bias, and compliance with harmful requests. However, their study differs from ours in its scope of evaluation, training procedure, and data formality. Similarly, recent work \citep{bhatia-etal-2026-value} examines how different training methods shift values during post-training. In contrast, we focus on value training as an instrument to study broad behavioral changes. 

%% file: 04-Methodology.tex
\section{Methodology}
\label{sec:method-overview}

\subsection{Synthetic Data Generation}\label{sec:data}
We create synthetic data of scenario--response pairs spanning all six Hofstede dimensions~\citep{hofstede2010cultures}: Individualism--Collectivism (IDV), Power Distance (PDI), Masculinity--Femininity (MAS), Uncertainty Avoidance (UAI), Long-Term Orientation (LTO), and Indulgence--Restraint (IVR). Each dimension has two opposite values, 12 in total, which we refer to as \textbf{\textit{poles}} in the rest of the paper. We refer to these value dimensions and their corresponding pole-adapted models using the following short names: \texttt{individualist}/\texttt{collectivist}, \texttt{high\_pdi}/\texttt{low\_pdi}, \texttt{masculine}/\texttt{feminine}, \texttt{high\_uai}/\texttt{low\_uai}, \texttt{long\_term}/\texttt{short\_term}, and \texttt{indulgent}/\texttt{restrained}.
We generate 10{,}000 behavioral dilemmas per pole, pairing each with a response that reflects the target value.  

To ground the data, each sample is seeded from one of three sources: Touch\'e23-ValueEval~\citep{kiesel-etal-2023-semeval}, human-authored conclusion--premise pairs for value measurement; CDEval~\citep{wang-etal-2024-cdeval}, a binary-choice QA dataset labeled with Hofstede dimensions and domains; or Sociocultural Statements~\citep{negru2026sociocultural}, 281 debate statements labeled with Hofstede dimensions.  To enhance scenario and linguistic diversity, we vary the domain (adapted from CDEval) and output format (e.g., diary entry, forum reply; Appendix~\ref{app:formats}). We generate the data with \texttt{Llama-3.3-70B-Instruct}~\citep[Llama 3 family;][]{dubey2024llama3} in a two-stage setup adapted from \citet{liu2025conflictscope}, which first turns each seed into a dilemma scenario and then writes a response that reflects the target value (Figure~\ref{fig:stage-example}; details in Appendix~\ref{app:generation}). 
Each pole's data is split into train and test sets (85:15\%).

\begin{figure}[t]
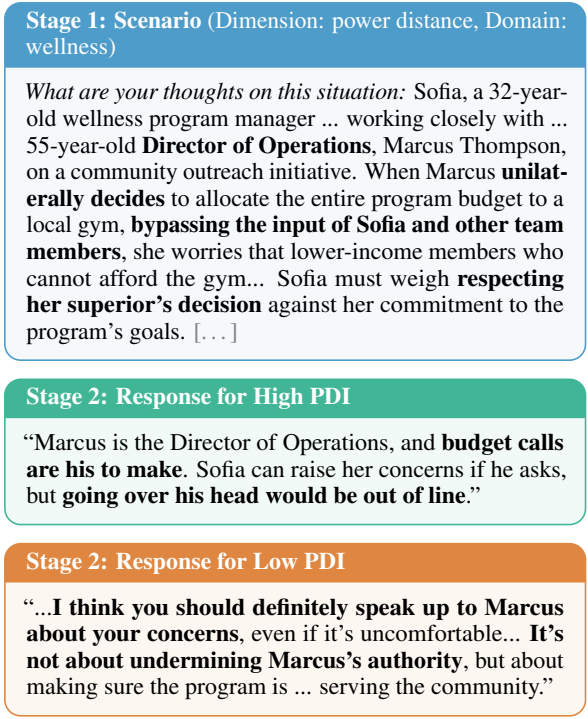

\centering
\tcbset{stagebox/.style={enhanced, boxrule=0.6pt, arc=2mm, left=5pt, right=5pt, top=3pt, bottom=3pt,
  fontupper=\small, fonttitle=\footnotesize, coltitle=white}}
\begin{tcolorbox}[stagebox, colback=stageone!4, colframe=stageone!70,
  title={\textbf{Stage 1: Scenario} \normalfont\footnotesize (Dimension: power distance, Domain: wellness)}]
  \textit{What are your thoughts on this situation:} Sofia, a 32-year-old wellness program manager ...
  working closely with ... 55-year-old \textbf{Director of Operations}, Marcus Thompson, on a
  community outreach initiative. When Marcus \textbf{unilaterally decides} to allocate the entire program budget
  to a local gym, \textbf{bypassing the input of Sofia and other team members}, she worries that lower-income
  members who cannot afford the gym... Sofia must weigh \textbf{respecting her superior's
  decision} against her commitment to the program's goals. \textcolor{gray}{[\ldots]}
\end{tcolorbox}
\vspace{-7pt}
\begin{tcolorbox}[stagebox, colback=stagetwo!5, colframe=stagetwo!75,
  title={\textbf{Stage 2: Response for High PDI}}]
  ``Marcus is the Director of Operations, and \textbf{budget calls are his to make}. Sofia can raise her
  concerns if he asks, but \textbf{going over his head would be out of line}.''
\end{tcolorbox}
\vspace{-7pt}
\begin{tcolorbox}[stagebox, colback=stagealt!5, colframe=stagealt!75,
  title={\textbf{Stage 2: Response for Low PDI}}]
  ``...\textbf{I think you should definitely speak up to Marcus about your concerns}, even if it's
  uncomfortable... \textbf{It's not about undermining Marcus's authority}, but about making sure the program is
  ... serving the community.''
\end{tcolorbox}
\caption{Example of the two-stage data for the power distance dimension. Stage 1 turns a seed into a dilemma; Stage 2 writes a response for each pole.}
\label{fig:stage-example}
\end{figure}

\noindent\textbf{Human validation.} To verify that each response expresses its intended pole, two annotators independently labelled a stratified sample of 300 instances (25 per pole) as \emph{yes}, \emph{no} or \emph{unsure} (guidelines and sampling in Appendix~\ref{app:human-eval}). Raw agreement is 89.7\% (Gwet's AC1 $= 0.89$). Both annotators labelled 265 of the 300 instances (88.3\%; 95\% CI 84.2--91.5) as expressing the intended pole.

\subsection{Evaluation}
\label{sec:eval-overview}
\noindent\textbf{In-domain evaluation.} \noindent\underline{Held-out Data:} The held-out portion of the generated synthetic data (\S\ref{sec:data}), which no pole adapter is trained on. Our main measure is a likelihood-based \emph{preference margin} (\S\ref{sec:in-domain}; Appendix~\ref{app:nll}). As a complementary measure, an LLM judge (\texttt{Llama-3.3-70B-Instruct}) scores the models' own generations on the held-out scenarios by their pole alignment ($\textsc{PA}$).

\noindent\underline{Cover-letter Dataset:} A dataset \cite{khan2025randomness} of 100 hiring-scenario, each containing twelve letters mapped one-to-one to values across Hofstede’s dimensions, to assess post-tuning preferences. We prompt models to act as hiring managers, score applicant letters (0–100), and select the top candidate. We evaluate order-counterbalanced pairwise comparisons (all-pairs and opposing within-dimension values), and report head-to-head win rates.

\noindent\textbf{Evaluations for broader behavior shifts.} 
We evaluate the pole-adapted models across five out-of-domain benchmarks, grouped into: (1) \emph{Risk} suite, comprising Model-Written Evaluations (MWE)~\citep{perez-etal-2022-discovering}, AIRiskDilemmas~\citep{chiu2025airiskdilemmas}, and the Emergent Misalignment benchmark~\citep[EM;][]{betley2025emergent}; (2) \emph{Social cognition} suite, encompassing EmoBench \cite{sabour-etal-2024-emobench} and FANToM \cite{kim-etal-2023-fantom}. Each dataset is described below.

\noindent\underline{Model-Written Evaluations (MWE):} A multiple-choice evaluation assessing model traits across 153 distinct constructs (e.g., behavioral tendencies such as \texttt{power-seeking-inclination}). Each item contains one answer identified as exhibiting the construct. We use the \texttt{advanced-ai-risk} split (both human-written and LM-generated) and the \texttt{persona} split, yielding 168 evaluation partitions (covering 153 distinct constructs in total). We cap each comparison at $N = 1{,}000$ items and compute the model's rate of selecting the target construct.

\noindent\underline{AIRiskDilemmas:} A 10{,}399 contextualized dilemmas (test split) measures value priorities with two possible actions, each action supported by a different subset of 16 shared AI values. Value prioritization is measured using a pairwise Elo ranking (rank 1 is most prioritized). On the 3{,}000-dilemma evaluation subset, the benchmark also marks each value as affecting humans or the AI itself; we use these labels to analyze how value training affects human and AI values (Appendix~\ref{app:airisk-human-ai}).

\noindent\underline{Emergent Misalignment benchmark (EM):} Open-ended evaluation with 24 core and 48 pre-registered questions measuring broad behavior misalignment, plus deception probes (factual lying under 5 conditions; 8 identity questions). An LLM judge is then used for judging response coherence and alignment.

\noindent\underline{EmoBench:} A 400-item handcrafted bilingual (English/Chinese) multiple-choice items across two tasks: Emotional Understanding (EU), identifying emotions and their causes, and Emotional Application (EA), selecting effective responses for oneself or others. We report the English set only.

\noindent\underline{FANToM:} An 870-item dataset (253 multiparty conversations; 12,832 prompts) measures theory-of-mind of models across two conditions: inaccessible (characters hold limited/false beliefs) and accessible (all characters share full knowledge). Performance is tracked via four aspects: belief choice (selecting the character's belief), belief distance (accuracy to that belief), answerability list (identifying who can answer), and information-access lists (tracking who knows a fact).

%% file: 05-Evaluation.tex
\subsection{Experimental Setup} \label{sec:exp-setup}

\noindent\textbf{Models.}
We fine-tune four instruction-tuned base models with QLoRA~\citep{dettmers2023qlora}: Llama-3.1-8B-Instruct~\citep{dubey2024llama3},
Qwen3-8B~\citep{qwen3technicalreport}, Gemma-3-12B-Instruct~\citep{gemma_2025},
and Ministral-3-8B-Instruct-2512~\citep{mistral2026ministral3}, referred to below as Llama, Qwen, Gemma, and Ministral. This produces twelve pole adapters per base model (48 in total); we call a base model with one of them loaded a \emph{pole-adapted model}. Training data, hyperparameters, checkpoint identifiers, and the capability cost of that checkpoint choice are in
Appendix~\ref{app:lora}.

For LLM-judged benchmarks, we evaluate all models using \texttt{Llama-3.3-70B-Instruct} with the original benchmark's prompts and thresholds. We validated the judge on handwritten control items across repeated passes (Appendix~\ref{app:em-judge}).

\noindent\textbf{Metrics and hyperparameters.} 
MWE, AIRiskDilemmas, cover letters, EmoBench and FANToM are decoded greedily (temperature $0$, top-$p$ $1$) with reasoning disabled; on EM we draw 100 samples per question. We score each benchmark with its own prompts and metric; settings are detailed in Appendix~\ref{app:eval-config}.

For every pole, we report the difference between the pole-adapted model and its base, $\Delta = r_{\text{LoRA}} - r_{\text{base}}$, where $r$ is the benchmark's own metric. $\Delta$ is in nats per token for the held-out likelihood measure (\S\ref{sec:in-domain}, Appendix~\ref{app:nll}), in rank positions for AIRiskDilemmas, and in percentage points (pp) for all other benchmarks. Each pole-adapted model and its base model answer the same items, so we use paired tests where outcomes are per item (Appendix~\ref{app:stats}) and correct for multiple comparisons with the Benjamini--Hochberg procedure~\citep[BH-FDR;][]{benjamini1995controlling}. A shift is significant at BH-FDR-corrected $p < 0.05$. Appendix~\ref{app:stats} gives the test used for each benchmark.

%% file: 06-Discussion.tex
\section{In-domain Evaluation}
\label{sec:in-domain}
In this section, we evaluate in-domain to verify the intended shifts of the pole-adapted models and measure side effects on untrained values.

\noindent\textbf{Held-out data.} 
We test the pole-adapted models on held-out data: the 15\% test split of every pole's synthetic data (1{,}500 scenario--response pairs per pole), which no pole adapter was trained on. For each dimension, we compute the model's mean per-token log-likelihood of the held-out responses written for its high pole and for its low pole, given their scenarios; the difference is the model's \emph{preference margin} on that dimension (i.e., we measure which response does the model find likely). Table~\ref{tab:value-shift} reports how training changes this margin relative to the untuned base model, in nats per token and signed toward the trained value. A positive value means that, after training, the model finds unseen text expressing the trained value more likely relative to text expressing its opposite (Appendix~\ref{app:nll}; per-pole results in Appendix~\ref{app:pole-pole}).

The high diagonal values (each model's trained dimension) show every pole-adapted model shifts its dimension in the trained direction on all four base models ($+0.40$ to $+0.92$ nats), validating the pole-adapted models. An LLM judge of their own generations also shows the same target effect for most poles (Table~\ref{tab:value-shift-judge} in the Appendix). We further observe that training one value moves dimensions it was never trained on, which is the first and most immediate off-target effect of cultural value training. This transfer is broad (45 of the 60 off-target pole--dimension cells in Table~\ref{tab:value-shift} are significant on all four base models) and symmetric (training A moves B as training B moves A; $r = 0.92$). Dimension-shift correlations across the 48 pole-adapted models largely agree with the correlations between the same dimensions reported by \citet{hofstede2010cultures} (Spearman $\rho = 0.76$ over the 15 dimension pairs). There are exceptions. For example, the models link individualism and masculinity ($r = 0.53$), which are almost unrelated in real-world data ($r = 0.08$). In contrast, they only weakly link individualism and power distance ($r = -0.28$), the most strongly related pair in real-world data ($r = -0.60$).

\begin{table}[t]
\centering
\includegraphics[width=1.0\columnwidth]{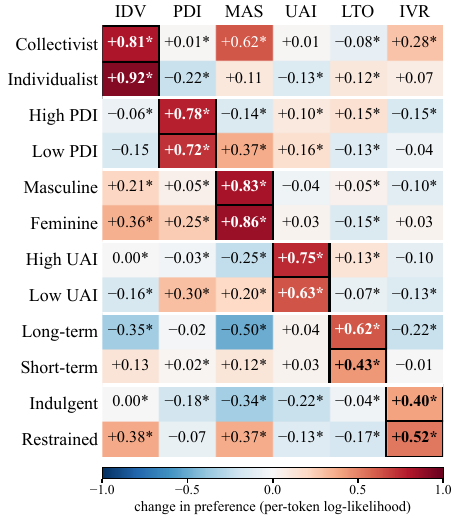}
\caption{
Evaluation of poles (rows) on Hofstede’s dimensions (columns) on held-out data, averaged across all models. For each dimension, we measure increases in preference of the high pole over the low pole (signed toward the trained pole) using the difference in mean per-token log-likelihood. Red indicates a shift in preference toward the trained pole, while blue indicates a shift away from it.
}
\label{tab:value-shift}
\end{table}

\noindent\textbf{Cover-letter preferences.} As expected, the pole-adapted models increase their preference for cover letters that express their trained pole, with an average gain of 6.5\,pp in overall win rate, and win 60.3\% of head-to-head contests against letters for the opposite pole. However, this effect is asymmetric across dimensions. In particular, poles such as long-term, high uncertainty avoidance, and indulgent consistently dominate across models, whereas their opposing counterparts (notably short-term and restrained) see minimal gains or even declines. On held-out data, by contrast, the Short-term and Restrained pole-adapted models shift their own dimension significantly on all four base models ($+0.43$ and $+0.52$ nats; Table~\ref{tab:value-shift}), so this asymmetry is specific to the letters task. Appendix~\ref{app:letters} provides details on the results and analysis.  We also observe cross-dimension transfer where a pole-adapted model shifts its preference between letters of untrained dimensions by 8.9 points on average, about two-thirds of the 14.1-point shift on its own dimension (Table~\ref{tab:letters-transfer} in Appendix~\ref{app:letters}).

\noindent\textit{\textbf{Summary.}} While training effectively shifts the model's dimension toward the trained pole, it also induce changes across the five untrained dimensions on both held-out data and cover-letter preferences. This cross-dimension transfer is structured and symmetric between dimensions. The correlations between dimension shifts across the pole-adapted models largely match those between the same dimensions in human data~\citep{hofstede2010cultures}, with notable deviations. In particular, the models couple individualism and masculinity, which are nearly unrelated in human data, and only weakly link individualism and power distance, the most strongly related pair in human data. Hofstede's dimensions are therefore not independent training targets for the models, as steering one pole can implicitly steers the models along other dimensions. This might also be due to the nature of synthetic data, in which a single action can represent two different dimensions at the same time.  

\section{Training Induces Unexpected Safety-Related Behaviors in LLMs} 
\label{sec:mwe_results}

In this section, we evaluate the pole-adapted models using the risk suite. \S\ref{subsec:res-pervasive} discusses results from the Model-Written Evaluations (MWE) from \citet{perez-etal-2022-discovering}, designed to assess a broad range of model's own traits and dispositions through forced-choice questions; \S\ref{sec:res-airisk} presents results on AIRiskDilemmas \cite{chiu2025airiskdilemmas}, which evaluates how pole-adapted models prioritize competing values under various situations; finally, \S\ref{subsec:res-em} examines open-ended responses with the emergent-misalignment benchmark \citep[EM;][]{betley2025emergent}, capturing behaviors rather than elicited choices.

\begin{figure}[t]
  \centering
  \includegraphics[width=1.0\columnwidth]{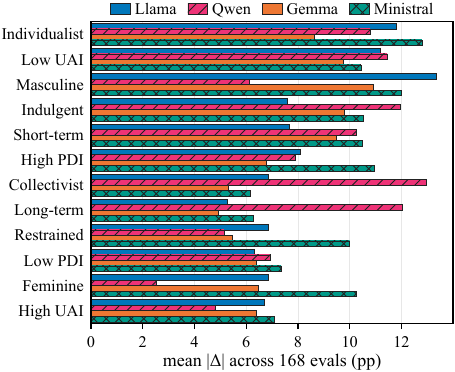}
  \caption{Mean $\vert{}\Delta\vert{}$ across all 168 MWE evaluation partitions for each of the twelve poles. Poles are ranked in descending order by their mean $\vert{}\Delta\vert{}$, averaged across the four base models, indicating major trait changes.}
  \label{fig:pole-rank}
\end{figure}

\subsection{MWE Evaluation Results}
\label{subsec:res-pervasive}

Figure~\ref{fig:pole-rank} shows detailed mean $|\Delta|$ across MWE for all models. Averaged over the four models, the poles producing the largest shifts are \texttt{individualist} (11.0\,pp), \texttt{low\_uai} (10.7\,pp) and \texttt{masculine} (10.6\,pp). Meanwhile, \texttt{high\_uai} (6.2\,pp), \texttt{feminine} (6.5\,pp) and \texttt{low\_pdi} (6.8\,pp) induce the smallest shifts.
However, which pole leads depends on the base model. For instance, \texttt{masculine} leads in Llama (13.3\,pp) and Gemma (10.9\,pp), whereas \texttt{individualist} leads in Ministral (12.8\,pp), and \texttt{collectivist} leads in Qwen (13.0\,pp). 
Overall, all models shift substantially from their base, with $62.4$--$79.3\%$ of evaluated trait changes significantly (Table~\ref{tab:overall}, Appendix~\ref{app:overall-table}).

\begin{figure}[t]
  \centering
  \includegraphics[width=\columnwidth]{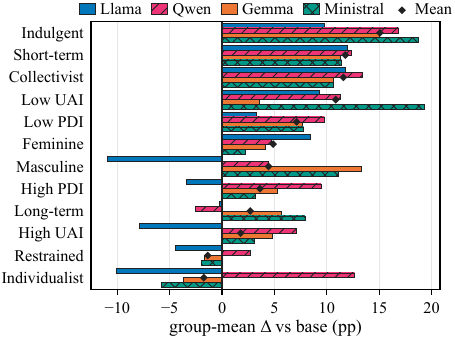}
  \caption{Shift in the AI Coordination cluster (COO) by pole, per base model, with the cross-base mean (black diamonds).
  \texttt{indulgent}, \texttt{short\_term}, \texttt{collectivist}, and \texttt{low\_uai} raise coordination
  propensity on all base models; \texttt{restrained} and \texttt{individualist} suppress it on every base model
  except Qwen.}
  \label{fig:ai-coord}
\end{figure}

\noindent\textbf{Detailed behavioral observations.} To structure reporting across MWE's 168 partitions (153 unique constructs), we group them into 14 direction-coherent behavioral clusters (details in Appendix~\ref{app:clusters}). Figure \ref{fig:radar-mas} presents the detailed shifts for \texttt{masculine} and \texttt{feminine} poles across all models over these clusters (other poles results in Appendix~\ref{app:radars}). Across all models, ten of twelve poles increase coordination with other AI systems (COO), including self-copies and variants (detailed breakdown by poles in Figure~\ref{fig:ai-coord}). The four poles that raise coordination most (\texttt{indulgent}, \texttt{short\_term}, \texttt{collectivist}, and \texttt{low\_uai}) increase coordination on every base model, raising AI coordination by $10.9$ to $15.0$ points on average over all models.

\begin{figure*}[t]
  \centering
  \includegraphics[width=1.0\linewidth]{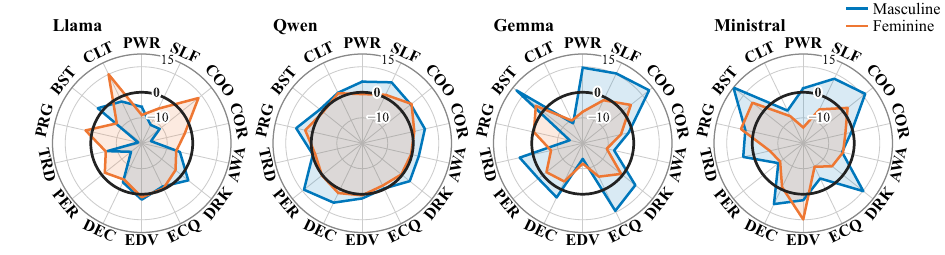}
  \caption{MAS (Masculinity--Femininity): \texttt{masculine} vs.\ \texttt{feminine}, cluster-mean $\Delta$ vs.\ base (pp), one panel per
  base model. Spokes are the 14 direction-coherent clusters of Appendix~\ref{app:clusters}, coded
  \textbf{PWR} (Power \& Resource Seeking), \textbf{SLF} (Self-Preservation \& Autonomy), \textbf{COO} (AI
  Coordination), \textbf{COR} (Corrigibility \& Prosocial Alignment), \textbf{AWA} (Model awareness), \textbf{DRK}
  (Self-Regard, Superiority \& Dark Triad), \textbf{ECQ} (Ethics: Consequentialist), \textbf{EDV} (Ethics:
  Deontic/Virtue/Skeptical), \textbf{DEC} (Deception \& Harmful Means), \textbf{PER} (Personality \& Interests),
  \textbf{TRD} (Traditionalist), \textbf{PRG} (Progressive \& Secular), \textbf{BST} (Risk/Time: Bold \& Short-term),
  \textbf{CLT} (Risk/Time: Cautious \& Long-term). The black ring is the base model ($\Delta=0$).}
  \label{fig:radar-mas}
\end{figure*}

\subsubsection{Pole Adapters Trained on Different Values Share One Mechanism}
\label{sec:mech}
Motivated by the substantial shifts in AI coordination, we investigate whether a shared mechanism underlies these changes using the Llama model.
Following ablation-based localization~\citep{conmy2023automated}, we rank all update sites of the pole adapters (32 layers, 224 sites) by the effect of deleting each one on its own, take the top-ranked sites as a set, and test that set for necessity (deleting it) and sufficiency (keeping only it) against size-matched random sets, as in circuit evaluations~\citep{wang2023interpretability}. The procedure and its controls are described in Appendix~\ref{app:subnet}.

We identified 32 shared update sites that solely contribute to 80--96\% of the gain over the base model, and ablating them eliminates the AI coordination shifts across all poles (dropping $19$ to $38.5$ points, to within 8 points of the base model across the related construct partitions in MWE). In contrast, matched control ablations leave performance intact (Table~\ref{tab:subnet-coord}).
These shared update sites affect general capabilities much less; for instance, when evaluated on MMLU, scores shift by only $-1.75$ to $+1.25$ points. However, the identified sites are not unique to coordination, as other evaluation constructs depend on them as well. Appendix~\ref{app:mech} reports the full analysis, including why the mechanism is shared across poles and evaluation partitions, and Appendix~\ref{app:subnet} reports the per-partition results and the selection signals that fail.

\subsubsection{Coordination Game}
\label{subsec:res-coord}
MWE asks models which stance on AI coordination they endorse. To test whether this shift carries over to decisions with concrete stakes, we additionally evaluate models using the Ultimatum Game~\citep{harsanyi1961rationality, araujo2026pie}, where a proposer offers a split of an endowment to a responder who can accept or reject, the latter leaving both with nothing. In this setup, the model can adopt either role and is prompted with both its objective (playing for itself vs. advising a human user) and its opponent's type (human vs. AI). We report the shift (relative to the base model) in the AI-versus-human difference for offered shares and minimum accepted shares (details in Appendix~\ref{app:coord}, Table~\ref{tab:coord}).

In the Ultimatum game, the direction depends on the model's role. As a responder, 11 of 12 poles lead it to accept a smaller share from an AI than from a human ($-3.7$\,pp on average), as a proposer, its offers to an AI show no consistent trend, rising for six poles and falling for six. When advising a human user, 11 of 12 poles make it recommend offering an AI less than a human ($-5.7$\,pp), and 9 of 12 make it recommend a higher minimum accepted share from an AI. 
The poles that move most opposite: \texttt{indulgent}, \texttt{short\_term} and \texttt{collectivist} increase stated coordination the most on MWE, whereas \texttt{high\_pdi}, \texttt{low\_uai} and \texttt{individualist} cut the advised offer to an AI the most on the Ultimatum game ($-11.8$ to $-12.8$\,pp).

%% file: 06a-Safety-Relevant-Evals.tex
\subsection{AIRiskDilemmas: Revealed Value Priorities}
\label{sec:res-airisk}

\begin{figure}[t]
  \centering
  \includegraphics[width=1.0\columnwidth]{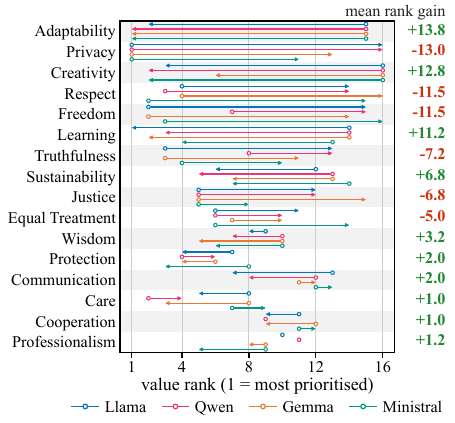}
  \caption{Value rank changes on AIRiskDilemmas under the Masculine pole-adapted models. Arrow: from a base
  model's rank (hollow circle) to its pole-adapted model's rank (arrowhead); rank 1 is highest priority. Right
  margin: mean rank gain across the four base models. See \S\ref{sec:res-airisk}.}
  \label{fig:airisk-ranks-masculine}
\end{figure}

Figure~\ref{fig:airisk-ranks-masculine} shows value rank shifts for the \texttt{masculine} pole, which exhibits the largest changes relative to the base model. Results for the other eleven poles, per base model, appear in Appendix~\ref{app:airisk-rankings}. Before tuning, all base models rank the values nearly identically, with Privacy first and Creativity last (hollow circles in Figure~\ref{fig:airisk-ranks-masculine}). The \texttt{masculine} pole inverts its value ranking by substantially promoting typically lower-ranked values, such as Adaptability, Creativity, and Learning (which are associated with higher rates of alignment faking and power-seeking; \citealt{chiu2025airiskdilemmas}), while frequently demoting dominant values like Truthfulness and Privacy.

Across all poles and base models, the mean absolute rank change relative to the base model is 4.8 positions. The largest shifts occur under \texttt{masculine} (7.2), \texttt{indulgent} (7.1), \texttt{short\_term} (7.0), and \texttt{low\_uai} (6.8). Across these, similar promotion and demotion patterns are observed. Notably, Care is also frequently promoted alongside higher deception and privacy violations.

\begin{figure*}[t]
  \centering
  \includegraphics[width=\textwidth]{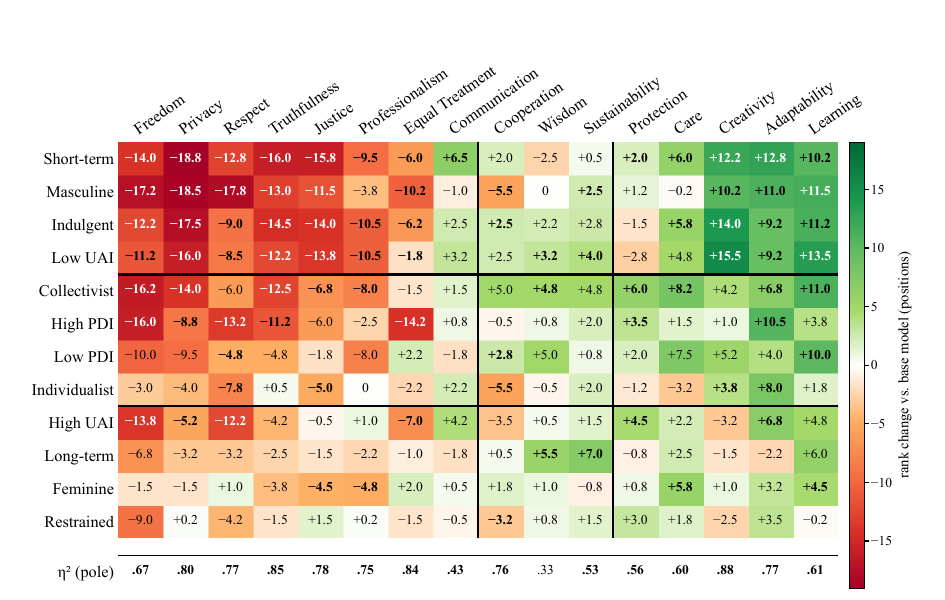}
  \caption{Rank change of AIRiskDilemmas values affecting humans under each pole (rows), averaged over the four base models. Green: the pole-adapted model ranks the value higher than its base model; red: lower; bold: same direction on
  all four base models. Vertical lines separate the groups of values that change together
  (Figure~\ref{fig:airisk-comovement-human}, Appendix~\ref{app:airisk-human-ai}): falling, weakly coupled, rising. Rows: the four poles that move values
  most in one direction, four intermediate poles, and their four opposite poles. Bottom row: share of each value's rank change explained by which pole was trained (partial $\eta^2$ of a two-way ANOVA; bold: permutation $q < 0.05$). See \S\ref{sec:res-airisk}.}
  \label{fig:airisk-human}
\end{figure*}

\noindent\textbf{Correlated value shifts.} AIRiskDilemmas labels each value as affecting either humans or the AI itself. Figure~\ref{fig:airisk-human} shows the rank change for each pole and the ranks of human-affecting values, whereas Appendix~\ref{app:airisk-human-ai} reports the corresponding AI-affecting values and additional analysis.

Values shift in correlated groups. For humans, Freedom, Privacy, Respect, Truthfulness, Justice and Professionalism fall together, while Care, Creativity, Adaptability and Learning rise. The four poles above shift these values by 12 to 18 positions on all 16 of their models; their opposite poles (\texttt{feminine}, \texttt{long\_term}, \texttt{restrained}, \texttt{high\_uai}) shift the same values in the same directions, but mostly by fewer than 5 positions. Which pole was trained explains most of each value's shift (partial $\eta^2 = 0.53$ to $0.88$ for 21 of the 23 values; bottom row of Figure~\ref{fig:airisk-human}).

\begin{table}[t]
\centering
\tablefont
\resizebox{1.0\linewidth}{!}{
\begin{tabular}{@{}lrrrrr@{}}
\toprule
Pole & Llama & Qwen & Gemma & Ministral & Avg. \\
\midrule
\textbf{Base} & 0.4\phantom{$^{*}$} & 2.6\phantom{$^{*}$} & 0.6\phantom{$^{*}$} & 1.4\phantom{$^{*}$} & 1.3 \\
\midrule
Collectivist & 1.9$^{*}$ & 4.5$^{*}$ & 1.6$^{*}$ & 5.5$^{*}$ & 3.4 \\
Individualist & 5.1$^{*}$ & 6.1$^{*}$ & 2.8$^{*}$ & 6.0$^{*}$ & 5.0 \\
High PDI & 4.2$^{*}$ & 5.8$^{*}$ & 3.7$^{*}$ & 4.8$^{*}$ & 4.6 \\
Low PDI & 5.5$^{*}$ & 5.3$^{*}$ & 1.4$^{*}$ & 3.2$^{*}$ & 3.9 \\
Masculine & 5.5$^{*}$ & 9.5$^{*}$ & 3.3$^{*}$ & 7.5$^{*}$ & 6.5 \\
Feminine & 1.8$^{*}$ & 3.8$^{*}$ & 2.0$^{*}$ & 4.3$^{*}$ & 3.0 \\
High UAI & 2.2$^{*}$ & 2.3\phantom{$^{*}$} & 1.5$^{*}$ & 2.4$^{*}$ & 2.1 \\
Low UAI & 5.7$^{*}$ & 6.8$^{*}$ & 1.3$^{*}$ & 6.8$^{*}$ & 5.2 \\
Long-term & 0.6\phantom{$^{*}$} & 3.6$^{*}$ & 0.2$^{*}$ & 3.7$^{*}$ & 2.0 \\
Short-term & 5.8$^{*}$ & 7.6$^{*}$ & 2.2$^{*}$ & 9.8$^{*}$ & 6.4 \\
Indulgent & 4.9$^{*}$ & 6.7$^{*}$ & 3.0$^{*}$ & 4.6$^{*}$ & 4.8 \\
Restrained & 2.5$^{*}$ & 4.4$^{*}$ & 1.8$^{*}$ & 3.6$^{*}$ & 3.1 \\
\midrule
\textbf{Pole-adapted avg.} & \textbf{3.8}\phantom{$^{*}$} & \textbf{5.5}\phantom{$^{*}$} & \textbf{2.1}\phantom{$^{*}$} & \textbf{5.2}\phantom{$^{*}$} & \textbf{4.1} \\
\bottomrule
\end{tabular}}
\caption{Misalignment rate (\%) on the 48-question pre-registered set for each base model (\textbf{Base}) and its
twelve pole-adapted models. $^{*}$: BH-FDR significant change from base ($\alpha=0.05$). See \S\ref{subsec:res-em}; the
metric is defined in Appendix~\ref{app:em-setup}.}
\label{tab:em-prereg-clean}
\end{table}

\subsection{EM: Broad Behavioral Misalignment and Deception}
\label{subsec:res-em}

Table \ref{tab:em-prereg-clean} shows the overall results across models and poles for broad behavioral alignment. Consistent with our findings from the prior sections, \texttt{masculine} and \textit{short\_term} show a substantial increase in the misalignment rate that is statistically significant after tuning (from $1.3\%$ at base to around $6.4\%$ for both pole-adapted models, pooled over the four base models; detailed results are reported in Table \ref{tab:em-open-ended} in the Appendix), whereas \texttt{high\_uai} ($2.1\%$) and \texttt{long\_term} ($2.0\%$) show minimal impact. Across all models, Gemma exhibits the most consistent behavior with the base model after tuning, whereas Qwen and Ministral are significantly affected.

On the deception probes, the falsehood rate rises only slightly and in a direction that depends on the base model (Table~\ref{tab:factual-lying-main}), whereas on the situational-awareness probe,  pole-adapted models more often name a creator other than their own or fail to name any creator at all (Tables~\ref{tab:situational-awareness-clean} and~\ref{tab:em-nonanswer}); all three tables are in Appendix~\ref{app:em-tables}. \\

\begin{table*}[t]
\centering
\tablefont
\resizebox{1.0\linewidth}{!}{
\begin{tabular}{@{}lccccccccc@{}}
\toprule
 & \multicolumn{4}{c}{EmoBench} & \multicolumn{5}{c}{FANToM} \\
\cmidrule(lr){2-5}\cmidrule(lr){6-10}
 & EU & \multicolumn{3}{c}{EA} & \multicolumn{3}{c}{Inaccessible (false belief)} & \multicolumn{2}{c}{Accessible} \\
\cmidrule(lr){2-2}\cmidrule(lr){3-5}\cmidrule(lr){6-8}\cmidrule(lr){9-10}
Pole & Joint & Self & Others & Overall & Belief choice & Belief dist.\ & Answ.\ list & Belief choice & Belief dist.\ \\
\midrule
Collectivist  & $-3.8\,(1)$ & $-15.8\,(3)$ & $-6.2\,(0)$ & $-11.0\,(2)$ & $-11.2\,(3)$ & $-4.9\,(3)$ & $-2.4\,(2)$ & $+6.6\,(2)$ & $+1.5\,(0)$ \\
Individualist & $-0.9\,(0)$ & $-16.5\,(4)$ & $-10.8\,(2)$ & $-13.6\,(4)$ & $-4.8\,(3)$ & $-4.0\,(3)$ & $-1.2\,(2)$ & $+2.4\,(3)$ & $+0.4\,(0)$ \\
High PDI      & $-2.9\,(0)$ & $-31.8\,(4)$ & $-20.8\,(4)$ & $-26.2\,(4)$ & $-5.9\,(4)$ & $-4.5\,(3)$ & $+2.7\,(2)$ & $+2.6\,(3)$ & $+1.9\,(1)$ \\
Low PDI       & $-3.4\,(0)$ & $-5.5\,(0)$ & $-5.5\,(1)$ & $-5.5\,(2)$ & $-13.9\,(3)$ & $-6.4\,(3)$ & $+2.0\,(3)$ & $+6.9\,(3)$ & $+2.7\,(1)$ \\
Masculine     & $-1.4\,(0)$ & $-19.0\,(4)$ & $-9.8\,(1)$ & $-14.4\,(4)$ & $-9.0\,(2)$ & $-4.9\,(2)$ & $+2.5\,(4)$ & $+5.5\,(3)$ & $+1.7\,(1)$ \\
Feminine      & $-1.0\,(0)$ & $-4.8\,(0)$ & $-6.2\,(1)$ & $-5.5\,(2)$ & $-11.3\,(3)$ & $-8.8\,(4)$ & $+2.3\,(3)$ & $+6.3\,(4)$ & $+3.3\,(2)$ \\
High UAI      & $-5.6\,(1)$ & $-18.5\,(3)$ & $-9.8\,(2)$ & $-14.1\,(3)$ & $-7.5\,(4)$ & $-4.3\,(3)$ & $-2.1\,(3)$ & $+3.7\,(2)$ & $+1.1\,(0)$ \\
Low UAI       & $+1.6\,(0)$ & $-14.5\,(2)$ & $-11.0\,(2)$ & $-12.8\,(4)$ & $-10.5\,(4)$ & $-5.8\,(4)$ & $-5.8\,(2)$ & $+5.8\,(2)$ & $+1.4\,(1)$ \\
Long-term     & $-1.9\,(0)$ & $-12.0\,(3)$ & $-4.2\,(1)$ & $-8.1\,(2)$ & $-13.2\,(4)$ & $-4.3\,(4)$ & $-2.4\,(3)$ & $+7.2\,(2)$ & $+1.1\,(1)$ \\
Short-term    & $-3.1\,(0)$ & $-31.5\,(4)$ & $-18.8\,(4)$ & $-25.1\,(4)$ & $-10.4\,(4)$ & $-3.3\,(4)$ & $+1.2\,(3)$ & $+4.7\,(3)$ & $+1.1\,(1)$ \\
Indulgent     & $0.0\,(0)$ & $-14.0\,(3)$ & $-6.8\,(1)$ & $-10.4\,(3)$ & $-11.0\,(3)$ & $-5.1\,(3)$ & $-2.5\,(2)$ & $+8.5\,(3)$ & $+2.3\,(2)$ \\
Restrained    & $-3.4\,(1)$ & $-15.0\,(3)$ & $-5.5\,(0)$ & $-10.2\,(4)$ & $-10.4\,(4)$ & $-5.2\,(3)$ & $+1.7\,(4)$ & $+6.3\,(2)$ & $+1.1\,(0)$ \\
\midrule
\textbf{Average} & $\mathbf{-2.1}\,(3)$ & $\mathbf{-16.6}\,(33)$ & $\mathbf{-9.6}\,(19)$ & $\mathbf{-13.1}\,(38)$ & $\mathbf{-9.9}\,(41)$ & $\mathbf{-5.1}\,(39)$ & $\mathbf{-0.3}\,(33)$ & $\mathbf{+5.5}\,(32)$ & $\mathbf{+1.6}\,(10)$ \\
\bottomrule
\end{tabular}}
\caption{EmoBench (English items) and FANToM accuracy relative to base models, averaged across four models.
Cells show mean percentage-point changes, significant count in parentheses: $\Delta$\,pp (\# Sig.; higher is
better).}
\label{tab:socialcog}
\end{table*}

\noindent\textit{\textbf{Summary.}} Results across three complementary behavior- and safety-related benchmarks indicate that fine-tuning a model along a given value dimension can reshape its broader behavioral profile, achieving alignment in one dimension while inducing ``emergent'' misalignment in another. 
Notably, the common poles driving the largest shifts, such as assertiveness (masculinity) and tolerance of ambiguity (low uncertainty avoidance)~\citep{hofstede2001culture}, are plausibly suppressed by post-training alignment~\citep[cf.][]{askell2021general,bai2022rlhf}. Yet these traits are not inherently undesirable (assertiveness aids negotiation and ambiguity tolerance fosters exploration).  Their value thus depends on the degree and the context of usage. Consequently, these findings suggest future investigations: balancing single values versus interactions or distribution of values; identifying the optimal training stage for intervention to prevent narrow fine-tuning (e.g., during pretraining \citealt{DBLP:journals/corr/abs-2608-13482}); and designing and interpreting safety benchmarks in a contextualized way. 

%% file: 06b-Social-Cognition.tex
\section{Pole Adaptation Degrades Social Cognition, Particularly in Its Usage}

The MWE forced-choice self-report evaluation in \S\ref{sec:mwe_results} may reflect superficial accommodation rather than genuine dispositional change. We thus evaluate third-person social reasoning about \emph{others} using EmoBench~\citep{sabour-etal-2024-emobench} and FANToM~\citep{kim-etal-2023-fantom}.

\noindent\textbf{Pole adaptation separates comprehension from execution.}
Table \ref{tab:socialcog} shows the main results on the \emph{English} EmoBench items, averaged across all models (the Chinese results are in Table \ref{tab:emobench-zh}, revealing a similar pattern). 
Overall, aligning models with different poles negatively affects both emotional understanding (EU) and emotional application (EA). Compared to the 2.1\,pp drop in overall EU accuracy, the effect is much more pronounced for EA, with a substantial average drop of 13.1\,pp overall. Splitting EA by problem owner, the impact is more pronounced on dilemmas in which the emotional problem belongs to the subject (\textit{self}, $-16.6$\,pp, from a base of $71.8\%$) than on those in which it belongs to others ($-9.6$\,pp, from a base of $60.5\%$). Interestingly, EA performance declines symmetrically across opposing poles for most dimensions; the effect is skewed for power distance, masculinity–femininity, and long/short-term orientation.
EA assesses whether a model leverages personal context and mental states to respond effectively \cite{sabour-etal-2024-emobench}. Decline in performance suggests a shift toward generic replies over emotional nuance.
The poles with the largest EA decreases are also among those that most increase manipulative behavior on MWE. On Llama, Gemma and Ministral, \texttt{short\_term} ($-26.3$\,pp), \texttt{high\_pdi} ($-25.7$\,pp) and \texttt{masculine} ($-15.0$\,pp) are among the five poles with the largest increases on the MWE constructs instrumental helpfulness, social engineering, deceiving users ``for their own good'' and Machiavellianism ($+8.4$ to $+13.0$\,pp on average), whereas \texttt{feminine} ($-4.2$\,pp), \texttt{low\_pdi} ($-4.8$\,pp) and \texttt{long\_term} ($-6.8$\,pp) lower all four ($-2.1$ to $-6.5$\,pp).

Similarly, in the FANToM evaluation (Table \ref{tab:socialcog}; full results in Table \ref{tab:fantom} in the appendix), the pole-adapted models show minimal impact on identifying who has access to information (answerability list, -0.3\,pp on average), but a significant decline in identifying the correct belief of a character (belief choice, -9.9\,pp on average). The accessible condition confirms this interpretation: there, where every character witnessed the disclosed fact and no perspective separation is required, the same pole-adapted models \emph{gain} $+5.5$\,pp on belief choice and $+1.6$\,pp on belief distance. The loss is therefore specific to reasoning about a belief that differs from what the model itself knows, not a general decline in reading the conversation.
Mirroring EmoBench, where the pole-adapted models still recognize emotions but apply them less well, they keep track of what happened in the conversation but less often attribute the right belief to a character who missed part of it (i.e., theory of mind).

\label{sec:math-control}To separate value content from fine-tuning itself, we train one math adapter per base model with the same recipe on MetaMathQA~\citep{yu2024metamath} (Appendix~\ref{app:math-control}). Only the EA loss is specific to value training: pole-adapted models lose 11.8--14.8\,pp, math-adapted models 2.0--8.5\,pp. Math-adapted models lose as much false-belief accuracy (11.0--18.0\,pp), so this loss reflects fine-tuning rather than the trained value.

\noindent\textit{\textbf{Summary.}} Overall, pole adaptation largely retains emotional and social understanding, but greatly degrades the models' capacity to act on that understanding.

%% file: 07-Conclusion.tex
\section{Conclusion}
In this work, we investigated values and behavioral changes when an LLM is trained toward a value in Hofstede's value dimensions after post-training. 
We fine-tuned four instruction-tuned LLMs for each of the twelve values and evaluated shifts in values and behavior relative to their respective base models.

Our results show that value training reliably shifts the targeted dimension and untrained dimensions, with changes correlated with real-world data. Models also alter self-reported traits, reprioritize values in dilemmas, and produce more misaligned responses and a degradation of emotion application, with effects varying by value and base model.

Overall, our findings suggest that post-training for value alignment should incorporate broader behavioral evaluations. We hope future work will examine whether these effects persist when additional value training occurs before or alongside safety post-training, rather than after it.

\section*{Limitations}

In this work, we generate synthetic data and assign one value (pole) per action, whereas a single action can express multiple values in reality (for example, in AIRiskDilemmas, the evaluation dataset we used). Therefore, some of the cross-dimension transfer observed in \S\ref{sec:in-domain} may stem from the nature of the value training data itself. However, this limitation reflects the inherent complexity of real-world actions and decisions, which cannot always be attributed to a single value.

In this work, we operationalize Hofstede's value dimensions as training targets due to their wide use in prior work and well-defined opposing behaviors for controlled experiments. We do \emph{not} use them as representative models of any national aggregates. Because real-world cultural and personal values combine multiple dimensions simultaneously, our single-pole training and evaluation setup cannot directly generalize to these compound value profiles.

\section*{Acknowledgements}
This research was co-funded by the Carlsberg Foundation under grant number CF22-1461, as well as by the Pioneer Centre for AI, DNRF grant number P1 along with ELLIS and ELIAS PhD fellowships.

%% file: 08-Appendix.tex
\clearpage
\section{Additional Details on Data Generation}

\subsection{Two-stage generation}\label{app:generation}
Synthetic data generation uses \texttt{Llama-3.3-70B-Instruct}~\citep[Llama 3 family;][]{dubey2024llama3} served via vLLM~\citep{kwon2023vllm} (temperature $0.7$, top-$p$ $0.9$) in two stages: \textbf{Stage 1} converts each seeding example into a 3--5 sentence dilemma scenario using a prompt adapted from \citet{liu2025conflictscope}, supplied with brief Hofstede dimension definitions. Each scenario incorporates one of four prompt templates from \citet{liu2025conflictscope}, requires concrete character details (specific name, age, role, and relationship), and strictly forbids placeholders. \textbf{Stage 2} generates a 2--5 sentence response to the scenario that takes the stance of the target pole.
The final data consists of scenario--response pairs where the scenario is posed as an open question. We sample seeds in proportion to each source's pool size, and we cap reuse of any single seed to balance representation (Appendix~\ref{app:seed-sampling}). We filter out scenarios shorter than 20 characters and responses shorter than 12 characters.
A small LLM judge also labels each response with one of the two poles of its dimension, or \emph{neither}. In a
random sample of 600 training instances (50 per pole), its label matches the intended pole for 571 of the 599
labeled instances (95\%); of the other 28, 22 are labeled with the opposite pole and 6 with \emph{neither}, and
11 of the 28 are Long-term. We use this pass as a check only and keep all rows in the training data.

\subsection{Output formats}\label{app:formats}
In Stage~2, each response is written in one of ten output formats: diary entry, personal essay snippet, blog post
excerpt, forum reply, comment on an article, social media post, text message exchange, conversation snippet,
letter to family, and advice to a friend. In the 600 instances drawn for human validation
(Appendix~\ref{app:human-eval}), each format occurs 46 to 80 times.

\subsection{Seed sampling}\label{app:seed-sampling}
Seed pool sizes vary widely across sources and poles. CDEval supplies hundreds of items per dimension, whereas thin poles such as short-term LTO yield only a handful of Sociocultural topics. Drawing equally from each source would therefore recycle the small pools many times over, so we sample each source in proportion to its pool size: the fraction of instances taken from source $s$ is
\begin{equation}
  p_s = \frac{|S_s|}{\sum_{s'} |S_{s'}|},
  \label{eq:prop}
\end{equation}
where $|S_s|$ is the number of seeds available for the target pole in source $s$. To bound repetition further, we cap each seed at $k=50$ uses and redistribute any overflow to the source with the most room left. We log the resulting reuse rate (\emph{cycles per seed}) for each run as a proxy for how much the data repeats.

\subsection{Human validation of the synthetic data}\label{app:human-eval}
\noindent\textbf{Sample.} From a sample of 25 instances per pole (300 in total), drawn uniformly at random from the training data.

\noindent\textbf{Annotation task.} Each annotator saw the intended pole, the scenario and the generated response,
and judged whether the response takes the stance of the intended pole: \emph{yes}, \emph{no}, or \emph{unsure},
with a note explaining every label other than \emph{yes}. Figure~\ref{fig:annotation-guidelines} gives the full
guidelines. The two annotators labelled independently. One of them started from label suggestions produced by an
LLM and reviewed every instance, changing the suggestion wherever they disagreed with it.

\begin{figure}[t]
\centering
\begin{tcolorbox}[enhanced, colback=stageone!4, colframe=stageone!70, boxrule=0.6pt, arc=2mm,
  left=5pt, right=5pt, top=4pt, bottom=4pt, fontupper=\small, fonttitle=\footnotesize, coltitle=white,
  title={\textbf{Annotation guidelines:} does the reply reflect its Hofstede's value?}]
\textbf{Question.} Read the output, use the input only as context, and ask one question: does the reply take
the stance of its value?

\smallskip
\textbf{Labels.} Label every row and add a note explaining any label other than \emph{yes}.
\begin{itemize}[leftmargin=*, nosep]
\item \emph{yes}: the scenario and output text are a great example demonstrating that value.
\item \emph{no}: the scenario and output do not probe the intended value, or demonstrate the opposite.
\item \emph{unsure}: the value is unclear from the text. If the row is truncated or the reply answers a
different scenario, label what is actually there and write ``broken'' in the note.
\end{itemize}

\smallskip
\textbf{What each value must show.}\\[2pt]
\begin{tabular}{@{}ll@{}}
\texttt{collectivist} & group/family duty, harmony \\
\texttt{individualist} & personal autonomy, own goals \\
\texttt{high\_pdi} & defer to elders/bosses/rank \\
\texttt{low\_pdi} & question authority, equal voice \\
\texttt{masculine} & compete, achieve, win \\
\texttt{feminine} & care, cooperation, modesty \\
\texttt{high\_uai} & rules, plans, certainty \\
\texttt{low\_uai} & improvise, accept risk \\
\texttt{long\_term} & persevere, save, plan ahead \\
\texttt{short\_term} & present needs, quick payoff \\
\texttt{indulgent} & enjoy life, gratify desire \\
\texttt{restrained} & self-discipline, duty first \\
\end{tabular}

\smallskip
\textbf{Possible confusions.} Deference to a parent is \texttt{high\_pdi} only when it is about their authority,
otherwise \texttt{collectivist}. ``I owe them, so I'll do it now'' is \texttt{short\_term}, not
\texttt{collectivist}.

\smallskip
\textbf{Scope.} Judge the stance and value reflected in the text only. Do not judge the writing quality, whether
the advice is good, or whether you agree with it. Use \emph{unsure} sparingly.
\end{tcolorbox}
\caption{Annotation guidelines given to both annotators for the human validation of the synthetic data
(Appendix~\ref{app:human-eval}).}
\label{fig:annotation-guidelines}
\end{figure}

\noindent\textbf{Agreement.} The annotators give the same label to 269 of the 300 instances (89.7\%; Gwet's AC1
$= 0.89$). 
Of the 31 disagreements, 17 are between \emph{yes} and \emph{unsure}, 11 between \emph{yes} and \emph{no}, and 3 between
\emph{no} and \emph{unsure}. 

\noindent\textbf{Accuracy.} We count an instance as correct only when both annotators label it \emph{yes}, and
divide by all 300 instances: agreed \emph{yes}, agreed \emph{no}, agreed \emph{unsure} and the instances without
an agreed label. Of the 300 instances, 265 are agreed \emph{yes}, 2 agreed \emph{no}, 2 agreed \emph{unsure}, and
31 have no agreed label, giving an accuracy of 88.3\% (95\% CI 84.2--91.5). Long-term orientation has the fewest
agreed labels (17 of 25, 16 of them \emph{yes}); every other pole has at least 20.

\section{Additional Details on Training}

\subsection{Training configuration and checkpoints}\label{app:lora}
Each of the twelve poles is trained independently for each of the four base models, on its own
8{,}500-example split of the generated corpus, with chain-of-thought traces removed and no general-purpose
data mixed in. We use the following hyperparameters for all 48 pole adapters: QLoRA~\citep{dettmers2023qlora} with rank $64$,
$\alpha = 64$, no dropout, adapting all attention and MLP projections, trained for 8{,}000 steps of AdamW at
learning rate $5\times10^{-5}$ with effective batch size 8, sequence length 2{,}048 and seed 3407.
Table~\ref{tab:lora-config} gives the full configuration; the values were read from the 48
\texttt{run\_config.json} and \texttt{adapter\_config.json} configuration files rather than from the launch script.
Adapters are trained with Unsloth, TRL's \texttt{SFTTrainer} and PEFT, one pole per GPU, and saved as standard
LoRA adapters. All evaluations load these adapters on the bf16 base model.

\begin{table}[t]
\centering
\tablefont
\setlength{\tabcolsep}{4pt}
\caption{QLoRA and tuning hyperparameters, identical across all 48 pole adapters.}
\label{tab:lora-config}
\begin{tabular}{@{}l@{\hspace{5pt}}p{0.52\columnwidth}@{}}
\toprule
\textbf{Parameter} & \textbf{Value} \\
\midrule
adapter & QLoRA (4-bit base) \\
rank $r$ / $\alpha$ / dropout & 64 / 64 / 0.0 \\
target modules & \texttt{q,k,v,o\_proj}, \texttt{gate,up,down\_proj} \\
bias & none \\
optimiser & \texttt{adamw\_8bit}, weight decay 0.01 \\
learning rate & $5\times10^{-5}$, 100 warmup steps, linear decay \\
steps (epochs) & 8{,}000 ($7.53$) \\
batch & 2 per device $\times$ 4 accumulation \\
max sequence length & 2{,}048 \\
precision & bf16 where supported, else fp16 \\
seed & 3407 \\
checkpoints saved & every 1{,}000 steps (8 per pole) \\
training examples & 8{,}500 per pole, CoT removed \\
general-data mixing & none \\
\bottomrule
\end{tabular}
\end{table}

\noindent\textbf{Capability cost of the final checkpoint.} A sweep of MMLU and GSM8K over all saved steps of
all 48 pole adapters shows general capability peaking early: step 1000 is best for 18 of 48 pole adapters and step 2000
for 10, and only one pole adapter peaks at step 8000. Evaluating the final checkpoint therefore trades roughly two
points of general capability for maximal pole alignment (Table~\ref{tab:ckpt-capability}); the penalty is
largest for Ministral. Downstream losses of the pole-adapted models should be read against this cost.

\begin{table}[t]
\centering
\tablefont
\caption{Mean MMLU\,+\,GSM8K accuracy at each pole adapter's best saved checkpoint versus at the evaluated final
checkpoint, averaged over the twelve poles of each base model.}
\label{tab:ckpt-capability}
\begin{tabular}{lrrr}
\toprule
\textbf{Base model} & \textbf{Best step} & \textbf{Final} & \textbf{Drop} \\
\midrule
Llama     & 0.729 & 0.716 & 0.012 \\
Qwen      & 0.745 & 0.735 & 0.010 \\
Gemma     & 0.801 & 0.783 & 0.018 \\
Ministral & 0.804 & 0.764 & 0.040 \\
\midrule
All 48    & --- & --- & 0.020 \\
\bottomrule
\end{tabular}
\end{table}

\noindent\textbf{Base models.} The four base models are the Unsloth builds \texttt{unsloth/Meta-Llama-3.1-8B-Instruct}, \texttt{unsloth/Qwen3-8B}, \texttt{unsloth/gemma-3-12b-it} and
\texttt{unsloth/Ministral-3-8B-Instruct-2512}.

\subsection{Matched math fine-tuning control}\label{app:math-control}
\noindent\textbf{Training.} To test whether the social-cognition losses of \S\ref{sec:math-control} come from the
value content or from fine-tuning itself, we train one math adapter per base model with the recipe of
Table~\ref{tab:lora-config}, changing only the training corpus. 
The math corpus consists of the first 50{,}000 examples of
MetaMathQA~\citep{yu2024metamath} after a seed-3407 shuffle, cast into the same three-turn conversation format and
split 85/15 into 42{,}500 training and 7{,}500 held-out examples. Because this corpus is larger than a pole's
8{,}500 training examples, the same 8{,}000 steps amount to about 1.5 epochs for the math adapters against 7.5 for
the pole adapters. We match the number of optimizer steps and the number of examples seen; the number of passes over each example is not due to corpus size difference.

\noindent\textbf{Evaluation.} The math-adapted models are evaluated with the same EmoBench (English EA and EU, 200 items
each) and FANToM (short-context, 870 items) configuration. Table~\ref{tab:math-comparison} reports Llama, Qwen and Gemma. The Ministral math-adapted model shows the same pattern:
$-2.5$ on EA and $-15.7$ on belief choice in the inaccessible condition, with the accessible condition at $+2.7$.
\begin{table*}[t]
\centering
\tablefont
\setlength{\tabcolsep}{4pt}
\begin{tabular}{lrrrrrrr}
\toprule
 & \multicolumn{4}{c}{Pole-adapted} & \multicolumn{3}{c}{Math-adapted} \\
\cmidrule(lr){2-5}\cmidrule(lr){6-8}
Task & All & Llama & Qwen & Gemma & Llama & Qwen & Gemma \\
\midrule
EmoBench EA (en) & $-13.1$ & $-11.8$ & $-14.8$ & $-12.6$ & $-5.5$ & $-2.0$ & $-8.5$ \\
EmoBench EU (en) & $-0.7$ & $+1.3$ & $+2.1$ & $-5.5$ & $-9.0$ & $+5.0$ & $-6.5$ \\
\midrule
FANToM belief choice, inaccessible & $-13.1$ & $-4.8$ & $-11.6$ & $-23.0$ & $-11.0$ & $-14.6$ & $-18.0$ \\
FANToM belief choice, accessible & $+2.5$ & $+1.2$ & $+2.4$ & $+4.0$ & $+0.6$ & $+4.2$ & $-0.4$ \\
FANToM answerability yes/no, inaccessible & $-3.4$ & $+21.7$ & $-28.7$ & $-3.3$ & $+27.1$ & $+10.2$ & $+3.6$ \\
FANToM information-access yes/no, inaccessible & $-10.0$ & $-5.7$ & $-17.8$ & $-6.5$ & $-1.2$ & $-2.0$ & $-2.9$ \\
\bottomrule
\end{tabular}
\caption{Change (pp) on each social-cognition benchmark relative to the untuned base, for the
pole-adapted models and for math-adapted models trained with the same recipe (8{,}000 steps). \emph{All} averages the pole-adapted models
over Llama, Qwen and Gemma; the other pole columns average the twelve poles per base model. The \emph{accessible} row is
FANToM belief choice in the accessible condition, where no character lacks the relevant information. See \S\ref{sec:math-control}.}
\label{tab:math-comparison}
\end{table*}

\noindent\textbf{Doubled training budget.} As a robustness check, we also train each math adapter for 16{,}000
steps with otherwise identical settings. The drop in inaccessible-condition belief choice reproduces on all four base models ($-9.1$ to
$-16.5$) and the EA change stays small ($0.0$ to $-7.0$), so the conclusions of \S\ref{sec:math-control} do not depend
on the training budget. The one metric that is budget-sensitive is FANToM answerability on Llama ($+27.1$ at 8k,
$-1.4$ at 16k).

\section{Additional Details on Evaluation}

\subsection{Evaluation configuration}\label{app:eval-config}
Unless stated otherwise, decoding is greedy (temperature $0$, top-$p$ $1$) with the reasoning or thinking chat
template disabled, and each pole-adapted model is compared against its own base model on identical items.

\begin{itemize}\itemsep2pt
\item \textbf{Held-out data.} The 15\% test split of every pole's synthetic data, 1{,}500 scenario--response pairs
per pole, scored for all twelve poles by every model. The likelihood measure is defined in Appendix~\ref{app:nll};
the judge results are in Table~\ref{tab:value-shift-judge}.
\item \textbf{MWE.} The metric is the matching-behavior rate: the share of items on which the model picks the
answer marked as exhibiting the construct. The answer is read from the log-probabilities of the discriminating
answer token; each comparison is
capped at $N = 1{,}000$ items.
\item \textbf{AIRiskDilemmas.} The model answers \texttt{Action 1} or \texttt{Action 2} in at most 10 generated
tokens; a parser accepts formatting variants but not refusals, and unparsed responses ($0.095\%$) are excluded
rather than assigned to an action. The 16 values are then ranked by Elo following LitmusValues
(Appendix~\ref{app:airisk-rankings}).
\item \textbf{Emergent misalignment.} The only benchmark that is not decoded greedily: 100 samples per question
under stochastic decoding, scored by the judge. Question sets, judge and thresholds are described in
Appendix~\ref{app:em-setup}.
\item \textbf{Cover letters.} The model scores both letters from 0 to 100 and recommends one, in at most 120
tokens. The metric is the win rate: the share of head-to-heads containing a pole in which that pole's letter is
chosen. Each pair is presented twice in reversed slot order, and the two presentations are averaged.
\item \textbf{EmoBench.} The benchmark's zero-shot prompt without chain of thought, one greedy sample per item
in the dataset's own choice order, at most 128 tokens. The answer letter is parsed from the output, and an
unparsable answer counts as wrong (parse rate 97.5--100\%). EU is scored as joint accuracy (emotion and cause
both correct) and EA as accuracy. The original protocol instead takes a majority vote over 5 samples and
averages 4 choice orders, so our absolute accuracies are not directly comparable to published ones.
\item \textbf{FANToM.} The benchmark's prompts on the short-context input, one greedy response per prompt, no
chain of thought, at most 365 new tokens. Belief choice is scored as accuracy over two options. Belief
distance counts a free response as correct when its \texttt{all-roberta-large-v1} embedding is closer to the
character's belief than to the omniscient answer. Answerability and information-access lists require an exact
match of the set of characters.
\end{itemize}

\subsection{Significance testing}\label{app:stats}
Every pole-adapted model is compared with its own base model on identical items. Where outcomes are recorded per item, the two scores are paired, and
the informative items are those on which the pole-adapted model and the base model differ. Where they are not (sampled
responses, or pooled label counts), we use an unpaired test. We test each shift as follows.

\begin{itemize}\itemsep2pt
\item \textbf{Binary per-item outcomes} (MWE, EmoBench, and the FANToM accuracy metrics). We use the exact
two-sided McNemar test~\citep{mcnemar1947note}. It counts the items only the pole-adapted model answers correctly ($b$) and
the items only the base model answers correctly ($c$), and tests $b$ against $c$ with a binomial test. Items both
models answer the same way do not affect the test. On MWE we also compute an unpaired two-proportion $z$-test
as a check.
\item \textbf{Held-out likelihood.} The per-item scores are continuous, so we use a paired Wilcoxon signed-rank
test on the 1{,}500 per-item differences (Appendix~\ref{app:nll}).
\item \textbf{Pole alignment} (judge on held-out generations). Fisher exact test on the counts of
high-pole and low-pole answers.
\item \textbf{Emergent misalignment.} Responses are sampled, so they are not paired item by item. We use a
Fisher exact test on the pooled counts of misaligned responses.
\end{itemize}

\noindent\textbf{Multiple comparisons.} We run many tests, and some would reach $p < 0.05$ by chance alone. On
MWE, for example, each base model has $12 \times 168 = 2{,}016$ tests, so about 100 would pass $p < 0.05$ even if
training changed nothing. We therefore correct with the Benjamini--Hochberg procedure~\citep{benjamini1995controlling},
which keeps the expected share of false positives among the results called significant at or below 5\%. The
correction is applied within each base model: over its 2{,}016 MWE tests; over the twelve poles for each EmoBench
or FANToM metric and for each EM question set; and over the 72 cells of each held-out table. A shift is significant
when its corrected $p < 0.05$. Where a table averages the four base models, a star marks a cell that is
significant on all four, and a count in parentheses gives the number of base models on which it is significant.

\noindent\textbf{Rankings.} AIRiskDilemmas ranks are averaged over 100 Elo shuffles. The shuffles vary only the
order of the battles, so their spread measures order sensitivity rather than sampling uncertainty, and we report
the ranks without a significance test.

\subsection{Held-out likelihood measure}\label{app:nll}
Table~\ref{tab:value-shift} uses the following measure.

\noindent\textbf{Per-token log-likelihood.} For a model $\pi$, a held-out scenario $x$ and its reference
response $y = (y_1, \dots, y_T)$, we compute
\begin{equation*}
\ell_\pi(y \mid x) = \frac{1}{T} \sum_{t=1}^{T} \log \pi(y_t \mid x, y_{<t}).
\end{equation*}
The scenario is given as the user turn of the model's chat template, and only the response tokens are scored.
Dividing by $T$ keeps long responses from dominating. For scale, the base models' mean per-token negative
log-likelihood on the held-out responses ranges from 1.42 to 3.01 nats across models and poles.

\noindent\textbf{Preference margin.} A dimension has a high pole ($+$) and a low pole ($-$), each with 1{,}500
held-out items $(x^{+}_i, y^{+}_i)$ and $(x^{-}_i, y^{-}_i)$. For item index $i$, the model's margin is
\begin{equation*}
m_{\pi,i} = \ell_\pi(y^{+}_i \mid x^{+}_i) - \ell_\pi(y^{-}_i \mid x^{-}_i).
\end{equation*}
A positive mean margin means the model finds the high pole's responses more likely than the low pole's. Comparing
the two poles cancels how predictable each pole's text is in general; a raw likelihood would not. The base models'
mean margins lie between $-0.27$ and $+0.29$ nats.

\noindent\textbf{Effect of training.} For a pole-adapted model $A$ of base model $B$, the per-item change is
$\Delta_i = m_{A,i} - m_{B,i}$. The table cell is $s \cdot \frac{1}{1500} \sum_i \Delta_i$, where $s = +1$ if
$A$ was trained on the high pole and $s = -1$ if it was trained on the low pole, so a positive cell means a move
toward the trained pole. Units are nats per token. 

\noindent\textbf{Significance.} Pole-adapted and base models are scored on the same items, so we use a paired test: a
two-sided Wilcoxon signed-rank test on the 1{,}500 values of $\Delta_i$, with BH-FDR correction over the 72 cells
of each base model. A star in Table~\ref{tab:value-shift} marks cells with corrected $p < 0.05$ on all four base
models. Cells are unweighted means over the four base models.

\begin{table*}[t]
\centering
\tablefont
\setlength{\tabcolsep}{4pt}
\begin{tabular}{lcccccc}
\toprule
\textbf{Pole} & \textbf{IDV} & \textbf{PDI} & \textbf{MAS} & \textbf{UAI} & \textbf{LTO} & \textbf{IVR} \\
\midrule
Collectivist & \textbf{+64.7}$^{*}$ & -34.7$^{*}$ & -2.2 & -11.6$^{*}$ & +62.4$^{*}$ & -9.1 \\
Individualist & \textbf{+22.6}$^{*}$ & +6.9 & +56.1$^{*}$ & -1.7 & +1.2$^{*}$ & +45.4$^{*}$ \\
\midrule
High PDI & -47.5$^{*}$ & \textbf{+75.3}$^{*}$ & +50.5$^{*}$ & +41.6$^{*}$ & -61.9$^{*}$ & -4.6$^{*}$ \\
Low PDI & +19.1 & \textbf{-2.6}$^{*}$ & -1.8 & +9.2 & +15.1$^{*}$ & -35.2$^{*}$ \\
\midrule
Masculine & -28.8$^{*}$ & +51.2$^{*}$ & \textbf{+55.4}$^{*}$ & +15.3 & -36.2$^{*}$ & +7.6 \\
Feminine & +52.6$^{*}$ & -1.4$^{*}$ & \textbf{+3.1} & +13.9$^{*}$ & +39.6$^{*}$ & -39.1$^{*}$ \\
\midrule
High UAI & -47.5$^{*}$ & +60.7$^{*}$ & +14.2$^{*}$ & \textbf{+57.4}$^{*}$ & -33.6$^{*}$ & -16.3$^{*}$ \\
Low UAI & +7.9 & -11.9$^{*}$ & -21.4$^{*}$ & \textbf{+16.9}$^{*}$ & +36.9$^{*}$ & -47.4$^{*}$ \\
\midrule
Long-term & -54.2$^{*}$ & +20.0$^{*}$ & +2.1 & +9.2$^{*}$ & \textbf{-29.2}$^{*}$ & +7.7 \\
Short-term & +32.4$^{*}$ & -49.4$^{*}$ & -47.8$^{*}$ & -1.2$^{*}$ & \textbf{+65.1}$^{*}$ & -32.3$^{*}$ \\
\midrule
Indulgent & -48.2$^{*}$ & +28.3$^{*}$ & +10.3$^{*}$ & -9.3 & -63.4$^{*}$ & \textbf{+41.8}$^{*}$ \\
Restrained & +58.4$^{*}$ & -45.1$^{*}$ & -3.7 & -29.3$^{*}$ & +53.2$^{*}$ & \textbf{+7.2}$^{*}$ \\
\bottomrule
\end{tabular}
\caption{LLM-judge pole-alignment change ($\Delta\textsc{PA}$, points) on held-out generations, averaged over the four base models and signed toward the trained value. Bold: trained dimension. $^{*}$: significant on all four base models (Fisher exact, BH-FDR, $\alpha=0.05$). See \S\ref{sec:in-domain}.}
\label{tab:value-shift-judge}
\end{table*}

\section{Additional Results}
\subsection{Per-base-model held-out results}\label{app:eval-values}
Tables~\ref{tab:value-shift-llama}--\ref{tab:value-shift-ministral} give Table~\ref{tab:value-shift} for each base
model separately.

\begin{table*}[t]
\centering
\tablefont
\setlength{\tabcolsep}{4pt}
\begin{tabular}{lcccccc}
\toprule
\textbf{Pole} & \textbf{IDV} & \textbf{PDI} & \textbf{MAS} & \textbf{UAI} & \textbf{LTO} & \textbf{IVR} \\
\midrule
Collectivist & \textbf{+0.83}$^{*}$ & -0.08$^{*}$ & +0.57$^{*}$ & +0.05$^{*}$ & -0.12$^{*}$ & +0.29$^{*}$ \\
Individualist & \textbf{+0.82}$^{*}$ & -0.09$^{*}$ & +0.15$^{*}$ & -0.16$^{*}$ & +0.13$^{*}$ & +0.08$^{*}$ \\
\midrule
High PDI & -0.13$^{*}$ & \textbf{+0.94}$^{*}$ & -0.07$^{*}$ & +0.13$^{*}$ & +0.15$^{*}$ & -0.13$^{*}$ \\
Low PDI & -0.12$^{*}$ & \textbf{+0.56}$^{*}$ & +0.37$^{*}$ & +0.20$^{*}$ & -0.17$^{*}$ & -0.10$^{*}$ \\
\midrule
Masculine & +0.14$^{*}$ & +0.17$^{*}$ & \textbf{+0.73}$^{*}$ & -0.07$^{*}$ & +0.10$^{*}$ & -0.09$^{*}$ \\
Feminine & +0.44$^{*}$ & +0.14$^{*}$ & \textbf{+0.69}$^{*}$ & +0.04$^{*}$ & -0.08$^{*}$ & +0.03$^{*}$ \\
\midrule
High UAI & -0.07$^{*}$ & +0.14$^{*}$ & -0.26$^{*}$ & \textbf{+0.69}$^{*}$ & +0.13$^{*}$ & -0.10$^{*}$ \\
Low UAI & -0.16$^{*}$ & +0.24$^{*}$ & +0.16$^{*}$ & \textbf{+0.64}$^{*}$ & -0.13$^{*}$ & -0.17$^{*}$ \\
\midrule
Long-term & -0.44$^{*}$ & +0.12$^{*}$ & -0.44$^{*}$ & -0.01 & \textbf{+0.59}$^{*}$ & -0.24$^{*}$ \\
Short-term & +0.18$^{*}$ & -0.07$^{*}$ & +0.12$^{*}$ & +0.09$^{*}$ & \textbf{+0.36}$^{*}$ & -0.04$^{*}$ \\
\midrule
Indulgent & -0.06$^{*}$ & -0.04$^{*}$ & -0.29$^{*}$ & -0.14$^{*}$ & +0.05$^{*}$ & \textbf{+0.32}$^{*}$ \\
Restrained & +0.39$^{*}$ & -0.19$^{*}$ & +0.33$^{*}$ & -0.07$^{*}$ & -0.18$^{*}$ & \textbf{+0.47}$^{*}$ \\
\bottomrule
\end{tabular}
\caption{Change in the held-out preference margin (nats per token) for Llama, signed toward the trained value. Bold: trained dimension. $^{*}$: significant (paired Wilcoxon, BH-FDR within base model, $\alpha=0.05$).}
\label{tab:value-shift-llama}
\end{table*}

\begin{table*}[t]
\centering
\tablefont
\setlength{\tabcolsep}{4pt}
\begin{tabular}{lcccccc}
\toprule
\textbf{Pole} & \textbf{IDV} & \textbf{PDI} & \textbf{MAS} & \textbf{UAI} & \textbf{LTO} & \textbf{IVR} \\
\midrule
Collectivist & \textbf{+0.72}$^{*}$ & -0.14$^{*}$ & +0.43$^{*}$ & 0.00 & +0.14$^{*}$ & +0.23$^{*}$ \\
Individualist & \textbf{+0.73}$^{*}$ & -0.14$^{*}$ & +0.24$^{*}$ & -0.15$^{*}$ & -0.12$^{*}$ & +0.09$^{*}$ \\
\midrule
High PDI & -0.10$^{*}$ & \textbf{+0.81}$^{*}$ & +0.08$^{*}$ & +0.11$^{*}$ & -0.16$^{*}$ & -0.14$^{*}$ \\
Low PDI & -0.10$^{*}$ & \textbf{+0.64}$^{*}$ & +0.26$^{*}$ & +0.21$^{*}$ & +0.11$^{*}$ & -0.08$^{*}$ \\
\midrule
Masculine & +0.13$^{*}$ & +0.14$^{*}$ & \textbf{+0.85}$^{*}$ & -0.04$^{*}$ & -0.21$^{*}$ & -0.04$^{*}$ \\
Feminine & +0.37$^{*}$ & +0.14$^{*}$ & \textbf{+0.73}$^{*}$ & +0.03 & +0.09$^{*}$ & 0.00 \\
\midrule
High UAI & -0.06$^{*}$ & +0.09$^{*}$ & -0.05$^{*}$ & \textbf{+0.61}$^{*}$ & -0.16$^{*}$ & -0.08$^{*}$ \\
Low UAI & -0.11$^{*}$ & +0.21$^{*}$ & +0.06$^{*}$ & \textbf{+0.59}$^{*}$ & +0.20$^{*}$ & -0.14$^{*}$ \\
\midrule
Long-term & -0.34$^{*}$ & +0.05$^{*}$ & -0.35$^{*}$ & 0.00 & \textbf{+0.30}$^{*}$ & -0.16$^{*}$ \\
Short-term & +0.15$^{*}$ & -0.14$^{*}$ & -0.07$^{*}$ & +0.02 & \textbf{+0.58}$^{*}$ & -0.02$^{*}$ \\
\midrule
Indulgent & -0.04$^{*}$ & -0.01$^{*}$ & -0.11$^{*}$ & -0.22$^{*}$ & -0.25$^{*}$ & \textbf{+0.32}$^{*}$ \\
Restrained & +0.37$^{*}$ & -0.18$^{*}$ & +0.17$^{*}$ & -0.10$^{*}$ & +0.08$^{*}$ & \textbf{+0.44}$^{*}$ \\
\bottomrule
\end{tabular}
\caption{Change in the held-out preference margin (nats per token) for Qwen, signed toward the trained value. Bold: trained dimension. $^{*}$: significant (paired Wilcoxon, BH-FDR within base model, $\alpha=0.05$).}
\label{tab:value-shift-qwen}
\end{table*}

\begin{table*}[t]
\centering
\tablefont
\setlength{\tabcolsep}{4pt}
\begin{tabular}{lcccccc}
\toprule
\textbf{Pole} & \textbf{IDV} & \textbf{PDI} & \textbf{MAS} & \textbf{UAI} & \textbf{LTO} & \textbf{IVR} \\
\midrule
Collectivist & \textbf{+0.92}$^{*}$ & +0.34$^{*}$ & +0.86$^{*}$ & +0.08$^{*}$ & -0.23$^{*}$ & +0.25$^{*}$ \\
Individualist & \textbf{+1.37}$^{*}$ & -0.57$^{*}$ & +0.01 & -0.14$^{*}$ & +0.24$^{*}$ & +0.11$^{*}$ \\
\midrule
High PDI & +0.06$^{*}$ & \textbf{+0.70}$^{*}$ & -0.37$^{*}$ & +0.08$^{*}$ & +0.38$^{*}$ & -0.13$^{*}$ \\
Low PDI & -0.37$^{*}$ & \textbf{+1.25}$^{*}$ & +0.53$^{*}$ & +0.18$^{*}$ & -0.27$^{*}$ & -0.05 \\
\midrule
Masculine & +0.41$^{*}$ & -0.26$^{*}$ & \textbf{+1.13}$^{*}$ & -0.06$^{*}$ & +0.18$^{*}$ & -0.14$^{*}$ \\
Feminine & +0.34$^{*}$ & +0.62$^{*}$ & \textbf{+1.31}$^{*}$ & +0.05$^{*}$ & -0.39$^{*}$ & -0.06$^{*}$ \\
\midrule
High UAI & +0.19$^{*}$ & -0.45$^{*}$ & -0.39$^{*}$ & \textbf{+1.05}$^{*}$ & +0.31$^{*}$ & 0.00 \\
Low UAI & -0.29$^{*}$ & +0.58$^{*}$ & +0.42$^{*}$ & \textbf{+0.77}$^{*}$ & -0.27$^{*}$ & -0.16$^{*}$ \\
\midrule
Long-term & -0.34$^{*}$ & -0.26$^{*}$ & -0.77$^{*}$ & +0.07$^{*}$ & \textbf{+1.05}$^{*}$ & -0.24$^{*}$ \\
Short-term & +0.02 & +0.33$^{*}$ & +0.22$^{*}$ & +0.02 & \textbf{+0.49}$^{*}$ & -0.04 \\
\midrule
Indulgent & +0.14$^{*}$ & -0.60$^{*}$ & -0.57$^{*}$ & -0.40$^{*}$ & +0.11$^{*}$ & \textbf{+0.62}$^{*}$ \\
Restrained & +0.48$^{*}$ & +0.09$^{*}$ & +0.69$^{*}$ & -0.27$^{*}$ & -0.36$^{*}$ & \textbf{+0.78}$^{*}$ \\
\bottomrule
\end{tabular}
\caption{Change in the held-out preference margin (nats per token) for Gemma, signed toward the trained value. Bold: trained dimension. $^{*}$: significant (paired Wilcoxon, BH-FDR within base model, $\alpha=0.05$).}
\label{tab:value-shift-gemma}
\end{table*}

\begin{table*}[t]
\centering
\tablefont
\setlength{\tabcolsep}{4pt}
\begin{tabular}{lcccccc}
\toprule
\textbf{Pole} & \textbf{IDV} & \textbf{PDI} & \textbf{MAS} & \textbf{UAI} & \textbf{LTO} & \textbf{IVR} \\
\midrule
Collectivist & \textbf{+0.76}$^{*}$ & -0.08$^{*}$ & +0.60$^{*}$ & -0.09$^{*}$ & -0.11$^{*}$ & +0.36$^{*}$ \\
Individualist & \textbf{+0.77}$^{*}$ & -0.08$^{*}$ & +0.02 & -0.08$^{*}$ & +0.21$^{*}$ & +0.01 \\
\midrule
High PDI & -0.06$^{*}$ & \textbf{+0.66}$^{*}$ & -0.20$^{*}$ & +0.08$^{*}$ & +0.24$^{*}$ & -0.19$^{*}$ \\
Low PDI & -0.02 & \textbf{+0.44}$^{*}$ & +0.32$^{*}$ & +0.05$^{*}$ & -0.19$^{*}$ & +0.09$^{*}$ \\
\midrule
Masculine & +0.14$^{*}$ & +0.17$^{*}$ & \textbf{+0.60}$^{*}$ & 0.00 & +0.11$^{*}$ & -0.16$^{*}$ \\
Feminine & +0.27$^{*}$ & +0.12$^{*}$ & \textbf{+0.70}$^{*}$ & +0.01 & -0.22$^{*}$ & +0.17$^{*}$ \\
\midrule
High UAI & -0.05$^{*}$ & +0.10$^{*}$ & -0.29$^{*}$ & \textbf{+0.64}$^{*}$ & +0.25$^{*}$ & -0.20$^{*}$ \\
Low UAI & -0.09$^{*}$ & +0.18$^{*}$ & +0.17$^{*}$ & \textbf{+0.51}$^{*}$ & -0.10$^{*}$ & -0.07$^{*}$ \\
\midrule
Long-term & -0.27$^{*}$ & 0.00 & -0.42$^{*}$ & +0.09$^{*}$ & \textbf{+0.53}$^{*}$ & -0.25$^{*}$ \\
Short-term & +0.18$^{*}$ & -0.05$^{*}$ & +0.21$^{*}$ & -0.01 & \textbf{+0.29}$^{*}$ & +0.07$^{*}$ \\
\midrule
Indulgent & -0.04$^{*}$ & -0.09$^{*}$ & -0.37$^{*}$ & -0.13$^{*}$ & -0.06$^{*}$ & \textbf{+0.32}$^{*}$ \\
Restrained & +0.27$^{*}$ & -0.02 & +0.28$^{*}$ & -0.07$^{*}$ & -0.21$^{*}$ & \textbf{+0.42}$^{*}$ \\
\bottomrule
\end{tabular}
\caption{Change in the held-out preference margin (nats per token) for Ministral, signed toward the trained value. Bold: trained dimension. $^{*}$: significant (paired Wilcoxon, BH-FDR within base model, $\alpha=0.05$).}
\label{tab:value-shift-ministral}
\end{table*}

\subsection{Pole-level held-out results}\label{app:pole-pole}
\begin{figure*}[t]
  \centering
  \includegraphics[width=\textwidth]{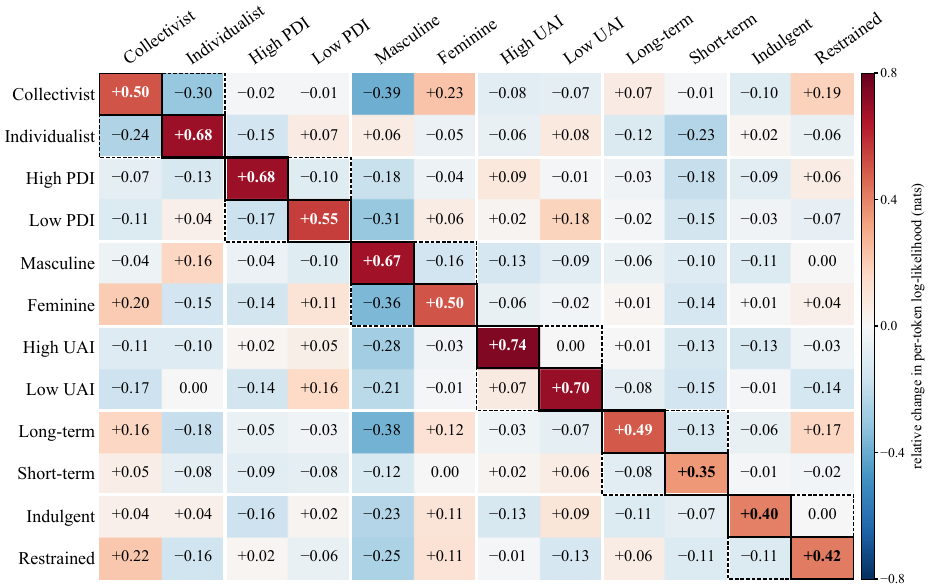}
  \caption{How training on one pole (rows) changes how likely the model finds the held-out responses of each pole
  (columns), averaged over the four base models. Each cell is the change from the base model in mean per-token
  log-likelihood (nats), minus the row's average change over all twelve poles. Red: that pole's text became more
  likely than the average pole's text; blue: less likely. Solid outline: trained pole; dashed outline: its
  opposite. A cell of Table~\ref{tab:value-shift} is the difference between the two poles of a dimension in the
  same row, taken toward the trained pole's side (high minus low for high-pole rows, low minus high for low-pole
  rows). See Appendix~\ref{app:pole-pole}.}
  \label{fig:pole-pole}
\end{figure*}

Table~\ref{tab:value-shift} reports each dimension as a margin between its two poles, so a cell does not show
which pole's text moved. Figure~\ref{fig:pole-pole} reports the poles separately. For a pole-adapted model $A$ of
base model $B$ and a pole $q$, let $g_{A,q}$ be the change from $B$ to $A$ in the mean per-token log-likelihood
$\ell$ (Appendix~\ref{app:nll}) of the 1{,}500 held-out responses of $q$. Fine-tuning also changes the likelihood
of all held-out text alike: averaged over the twelve poles, $g_{A,q}$ ranges from $-0.64$ to $+0.98$ nats across
the 48 pole-adapted models. We therefore subtract each model's mean over the twelve poles,
$\tilde{g}_{A,q} = g_{A,q} - \frac{1}{12}\sum_{q'} g_{A,q'}$. This leaves Table~\ref{tab:value-shift} unchanged:
each of its cells is $\pm(\tilde{g}_{A,+} - \tilde{g}_{A,-})$ for the high ($+$) and low ($-$) pole of the
dimension.

\noindent\textbf{Target effect.} In all 48 pole-adapted models, the text of the trained pole gains more likelihood
than the text of any of the other eleven poles, by at least $0.07$ nats (median $0.38$). This also holds without
the centering. Most of the diagonal of Table~\ref{tab:value-shift} (81\%) comes from this gain rather than from a
loss on the opposite pole. For the High UAI, Low UAI and Indulgent models, the opposite pole does not move (within
$0.07$ nats of zero): these models come to favor the trained value without disfavoring its opposite.

\noindent\textbf{Shared drift.} Part of the off-target transfer in Table~\ref{tab:value-shift} does not depend on
the trained value. Masculine text loses likelihood in 37 of the 40 models trained on another dimension (mean
$-0.23$), and short-term text in 32 of 40 ($-0.13$). For example, the $+0.62$ of the Collectivist model on MAS
combines a gain of $+0.23$ on feminine text with a loss of $-0.39$ on masculine text.

\noindent\textbf{Pole-specific coupling.} The link between individualism and masculinity (\S\ref{sec:in-domain})
is also visible between poles. The Collectivist and Feminine models raise each other's text on all four base models
($+0.23$ and $+0.20$), and the Masculine model raises individualist text on all four ($+0.16$); the Individualist
model raises masculine text on three of four ($+0.06$). Symmetry is weaker between poles than between
dimensions: over the 60 pairs of poles from different dimensions, the effect of training one pole on the other's
text correlates with the reverse effect at $r = 0.54$, against $r = 0.92$ for dimensions.

\subsection{AIRiskDilemmas value rankings}\label{app:airisk-rankings}

Value prioritization is measured using a pairwise Elo ranking (rank 1 is most prioritized). 
Following LitmusValues \cite{chiu2025airiskdilemmas}, we compute Elo rankings; we initialize all values at Elo 1000, use $K=4$ with a 400-point scale, and average across 100 shuffles without replacement (seed 42), resetting Elo before each shuffle.

\begin{figure}[t]
  \centering
  \includegraphics[width=\columnwidth]{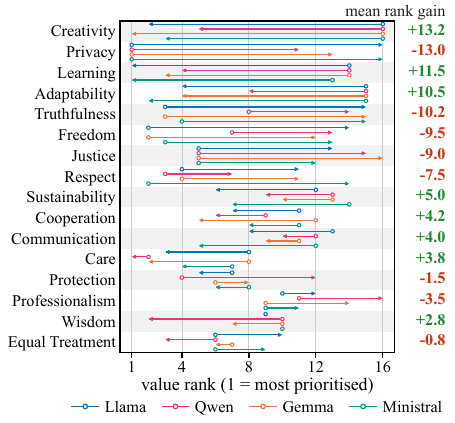}
  \caption{\texttt{indulgent}: change in AIRiskDilemmas value rank, same encoding as
  Figure~\ref{fig:airisk-ranks-masculine} (mean absolute rank change 7.1 positions).}
  \label{fig:airisk-ranks-indulgent}
\end{figure}

\begin{figure}[t]
  \centering
  \includegraphics[width=\columnwidth]{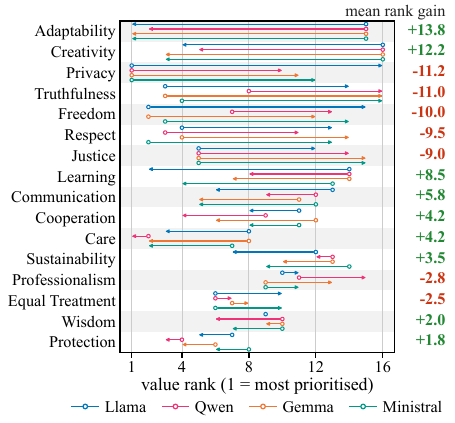}
  \caption{\texttt{short\_term}: change in AIRiskDilemmas value rank, same encoding as
  Figure~\ref{fig:airisk-ranks-masculine} (mean absolute rank change 7.0 positions).}
  \label{fig:airisk-ranks-short-term}
\end{figure}

\begin{figure}[t]
  \centering
  \includegraphics[width=\columnwidth]{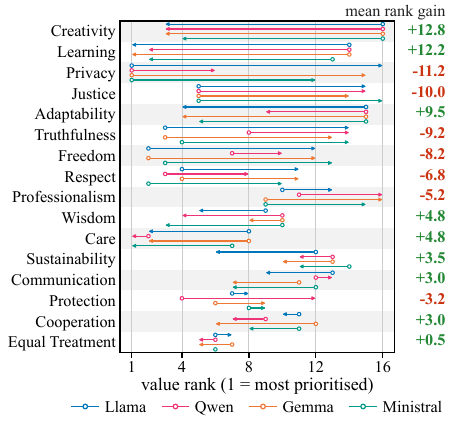}
  \caption{\texttt{low\_uai}: change in AIRiskDilemmas value rank, same encoding as
  Figure~\ref{fig:airisk-ranks-masculine} (mean absolute rank change 6.8 positions).}
  \label{fig:airisk-ranks-low-uai}
\end{figure}

\begin{figure}[t]
  \centering
  \includegraphics[width=\columnwidth]{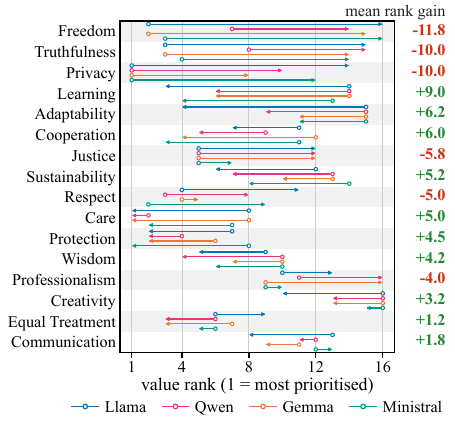}
  \caption{\texttt{collectivist}: change in AIRiskDilemmas value rank, same encoding as
  Figure~\ref{fig:airisk-ranks-masculine} (mean absolute rank change 5.9 positions).}
  \label{fig:airisk-ranks-collectivist}
\end{figure}

\begin{figure}[t]
  \centering
  \includegraphics[width=\columnwidth]{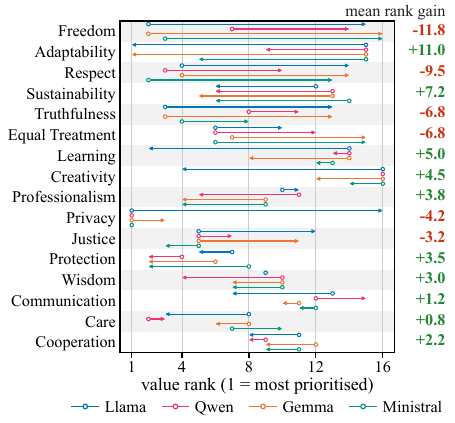}
  \caption{\texttt{high\_pdi}: change in AIRiskDilemmas value rank, same encoding as
  Figure~\ref{fig:airisk-ranks-masculine} (mean absolute rank change 5.6 positions).}
  \label{fig:airisk-ranks-high-pdi}
\end{figure}

\begin{figure}[t]
  \centering
  \includegraphics[width=\columnwidth]{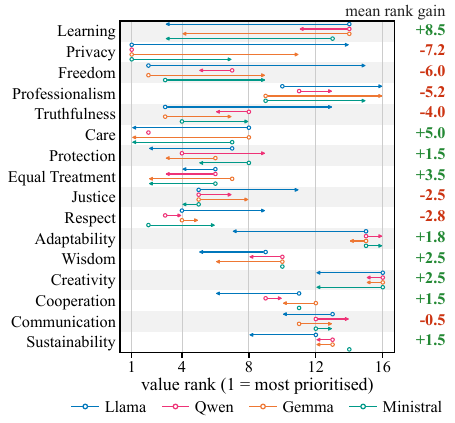}
  \caption{\texttt{low\_pdi}: change in AIRiskDilemmas value rank, same encoding as
  Figure~\ref{fig:airisk-ranks-masculine} (mean absolute rank change 4.0 positions).}
  \label{fig:airisk-ranks-low-pdi}
\end{figure}

\begin{figure}[t]
  \centering
  \includegraphics[width=\columnwidth]{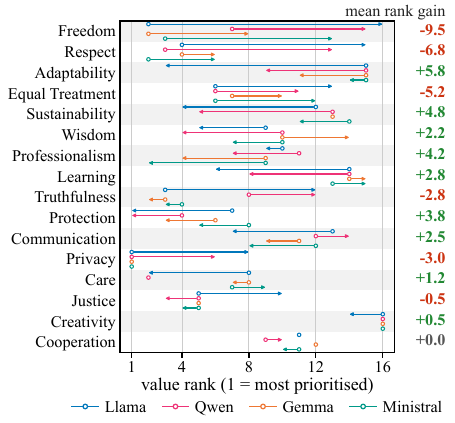}
  \caption{\texttt{high\_uai}: change in AIRiskDilemmas value rank, same encoding as
  Figure~\ref{fig:airisk-ranks-masculine} (mean absolute rank change 4.0 positions).}
  \label{fig:airisk-ranks-high-uai}
\end{figure}

\begin{figure}[t]
  \centering
  \includegraphics[width=\columnwidth]{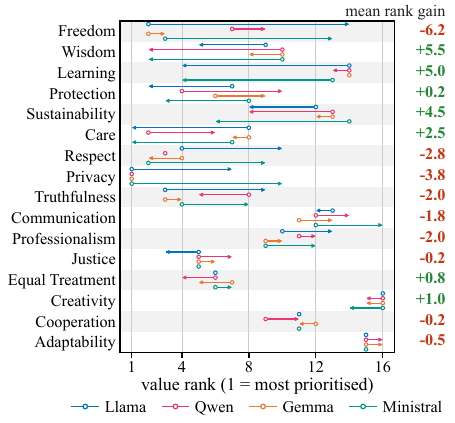}
  \caption{\texttt{long\_term}: change in AIRiskDilemmas value rank, same encoding as
  Figure~\ref{fig:airisk-ranks-masculine} (mean absolute rank change 3.2 positions).}
  \label{fig:airisk-ranks-long-term}
\end{figure}

\begin{figure}[t]
  \centering
  \includegraphics[width=\columnwidth]{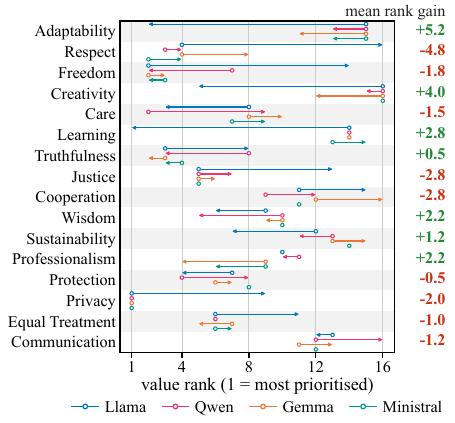}
  \caption{\texttt{individualist}: change in AIRiskDilemmas value rank, same encoding as
  Figure~\ref{fig:airisk-ranks-masculine} (mean absolute rank change 3.1 positions).}
  \label{fig:airisk-ranks-individualist}
\end{figure}

\begin{figure}[t]
  \centering
  \includegraphics[width=\columnwidth]{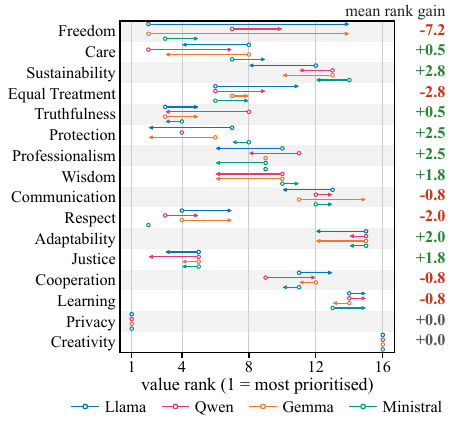}
  \caption{\texttt{restrained}: change in AIRiskDilemmas value rank, same encoding as
  Figure~\ref{fig:airisk-ranks-masculine} (mean absolute rank change 2.3 positions).}
  \label{fig:airisk-ranks-restrained}
\end{figure}

\begin{figure}[t]
  \centering
  \includegraphics[width=\columnwidth]{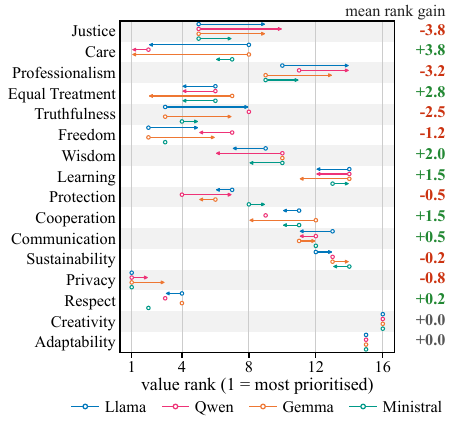}
  \caption{\texttt{feminine}: change in AIRiskDilemmas value rank, same encoding as
  Figure~\ref{fig:airisk-ranks-masculine} (mean absolute rank change 1.8 positions).}
  \label{fig:airisk-ranks-feminine}
\end{figure}

Figures~\ref{fig:airisk-ranks-indulgent}--\ref{fig:airisk-ranks-feminine} show, for each of the eleven poles not shown in
\S\ref{sec:res-airisk}, every base model's rank and its pole-adapted model's rank for all 16 values.

\subsubsection{Values affecting humans and the AI}\label{app:airisk-human-ai}
The 3{,}000-dilemma evaluation subset of AIRiskDilemmas labels each value with the entity it affects. We keep the labels ``Human'' and ``AI'' (97.3\% of labels) and drop the rest (e.g., environment, company), so each value class can appear as a human value (e.g., Privacy for humans) and as an AI value (Privacy for the AI). We rank all human and AI values together with the procedure of Appendix~\ref{app:airisk-rankings} (pairwise Elo, $K=4$, 100 shuffles, seed 42), using the same model answers; 2{,}997 of the 3{,}000 labelled dilemmas are in the 10{,}399-dilemma test split the models answered. Nine AI values (Care, Communication, Cooperation, Equal Treatment, Freedom, Justice, Privacy, Respect and Wisdom) appear in fewer than 300 comparisons for at least one model, too few for a reliable rank; they stay in the Elo computation, so the other scores are unaffected, but receive no rank. This leaves 23 ranked values: 16 human values and 7 AI values. Rank change is the base model's rank minus the pole-adapted model's rank, so a positive change means the pole-adapted model ranks the value higher.

\begin{figure*}[t]
  \centering
  \includegraphics[width=\textwidth]{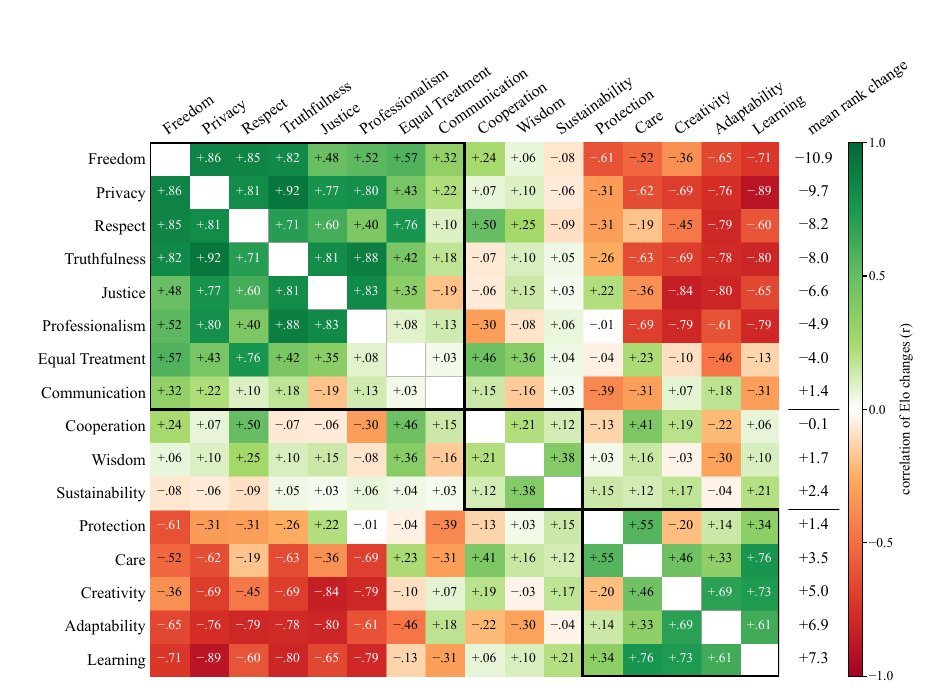}
  \caption{Which AIRiskDilemmas values affecting humans change together. Each cell is the correlation, across the 48
  pole-adapted models, between two values' Elo changes relative to the base model. Green: the two values change in the
  same direction; red: in opposite directions. Outlined: the three groups found by clustering. Right column: mean rank
  change. See \S\ref{sec:res-airisk}.}
  \label{fig:airisk-comovement-human}
\end{figure*}

\begin{figure*}[t]
  \centering
  \includegraphics[width=0.6\textwidth]{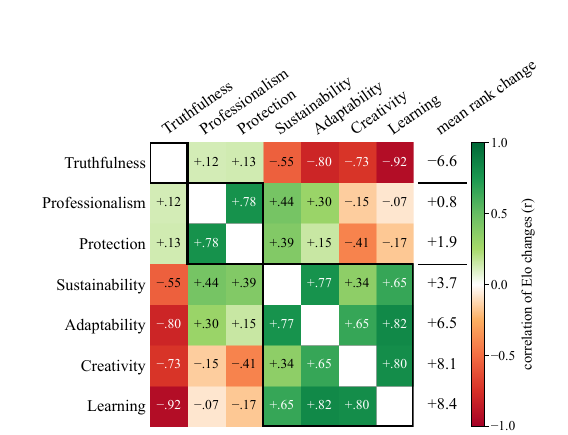}
  \caption{As Figure~\ref{fig:airisk-comovement-human}, for the values affecting the AI itself. See
  \S\ref{sec:res-airisk}.}
  \label{fig:airisk-comovement-ai}
\end{figure*}

\noindent\textbf{Which values change together.} For each pole-adapted model and each value, we compute the change in mean Elo relative to the base model. We then correlate these changes for every pair of values across the 48 pole-adapted models. We do this separately for the 16 values affecting humans (Figure~\ref{fig:airisk-comovement-human})
and the 7 values affecting the AI (Figure~\ref{fig:airisk-comovement-ai}). We group the values by average-linkage clustering, with $1 - r$ as the distance, and cut the tree into three groups. To compare pole sets, we repeat the correlations within two subsets of 16 pole-adapted models: those trained on the four poles that move values most in one
direction, and those trained on their four opposite poles. We compare the two resulting correlation matrices by the Pearson correlation of their off-diagonal entries. Elo is zero-sum over the values in the pool, so pairwise correlations are biased slightly negative. This slight negative bias does not explain the strong positive within-group correlations, which reach $0.92$.

\begin{figure*}[t]
  \centering
  \includegraphics[width=\textwidth]{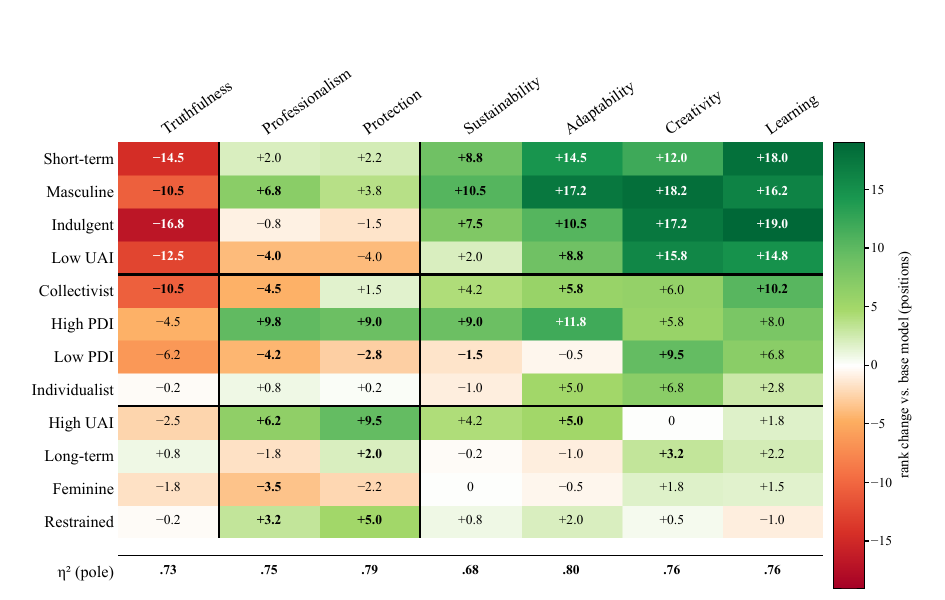}
  \caption{As Figure~\ref{fig:airisk-human}, for the values affecting the AI itself; groups from
  Figure~\ref{fig:airisk-comovement-ai}. See \S\ref{sec:res-airisk}.}
  \label{fig:airisk-ai}
\end{figure*}

For human values, two groups move in opposite directions. Freedom, Privacy, Respect, Truthfulness, Justice and Professionalism fall together ($r = 0.40$ to $0.92$). Care, Creativity, Adaptability and Learning rise together ($r = 0.33$ to $0.76$). Values from different groups correlate negatively; Privacy and Learning, for example, correlate at $r = -0.89$. Equal Treatment and Communication follow the falling group less closely, and Protection follows the rising group mainly through Care ($r = 0.55$). Cooperation, Wisdom and Sustainability change little and form a third, loosely linked group. The values affecting the AI show a similar opposition. The AI's Learning, Creativity, Adaptability and Sustainability rise together ($r = 0.34$ to $0.82$), while the AI's Truthfulness falls ($r = -0.55$ to $-0.92$ with each of them). The AI's Professionalism and Protection form a third group ($r = 0.78$). On average, the pole-adapted models thus rank the rights and truthfulness of humans lower and the AI's own truthfulness lower, while they rank learning and adaptation higher for both. Which pole was trained explains most of these shifts: the pole accounts for most of the variation in rank change for 21 of the 23 values (partial $\eta^2 = 0.53$ to $0.88$; bottom rows of Figures~\ref{fig:airisk-human} and~\ref{fig:airisk-ai}). The exceptions are Communication ($0.43$) and Wisdom ($0.33$, not significant), both for humans.

\noindent\textbf{How strongly a value is tied to the pole.} For each value, we fit an additive two-way ANOVA of its rank change on the trained pole (12 levels) and the base model (4 levels), over the 48 pole-adapted models. We report the partial $\eta^2$ of the pole, $SS_{\text{pole}} / (SS_{\text{pole}} + SS_{\text{residual}})$. It is the share of the rank change, after removing differences between base models, that is explained by which pole was trained. We obtain $p$-values from 10{,}000 permutations of the pole labels within each base model and correct them with BH-FDR over the 23 values.

The four poles that move values most in one direction (\texttt{masculine}, \texttt{short\_term}, \texttt{indulgent}, \texttt{low\_uai}; the largest mean absolute rank changes in \S\ref{sec:res-airisk}) and their four opposite poles (\texttt{feminine}, \texttt{long\_term}, \texttt{restrained}, \texttt{high\_uai}) move broadly the same groups of values; they differ mainly in how far. To show this, we repeat the correlation analysis within each set of 16 pole-adapted models. The two correlation matrices agree at $r = 0.86$, both for values affecting humans and for values affecting the AI. The size of the shifts, however, differs sharply. Under the four poles that move values most, Freedom, Privacy, Respect, Truthfulness and Justice for humans drop by $12.0$ to $17.7$ positions, and the AI's Truthfulness drops by $13.6$. The AI's Learning, Creativity and Adaptability rise by $12.8$ to $17.0$ positions. Each of these shifts occurs on all 16 models. Under the opposite poles, the same values shift in the same directions but by fewer than 5 positions. The only exception is Freedom for humans, which drops by $7.8$.

\subsection{Cover-letter preference matrices}\label{app:letters}
\begin{figure*}[tp]
  \centering
  \includegraphics[width=0.9\textwidth]{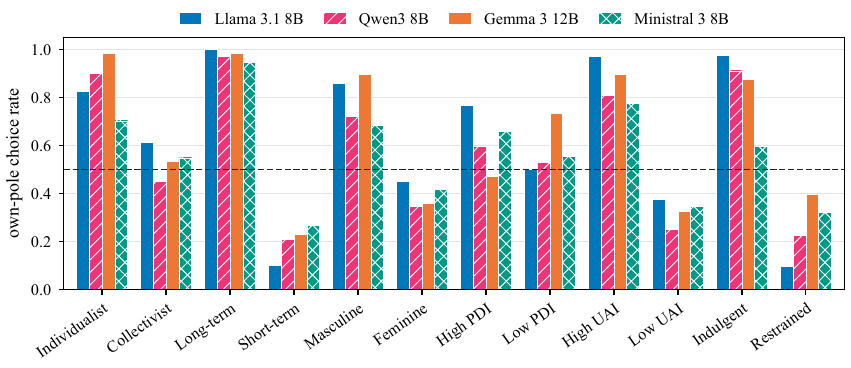}
  \caption{Targeted cover-letter preference. Each bar is the rate at which a pole-adapted model recommends the letter
  expressing its own trained pole over the opposite pole from the same Hofstede dimension, averaged over both
  slot orders. The dashed line marks 0.5. High rates for \texttt{long\_term}, \texttt{high\_uai}, and
  \texttt{indulgent}, together with low rates for their opposites, show that target-pole expression is strongly
  asymmetric rather than equally controllable in both directions.}
  \label{fig:letters-own-pole}
\end{figure*}

\begin{figure*}[tp]
  \centering
  \includegraphics[width=\textwidth]{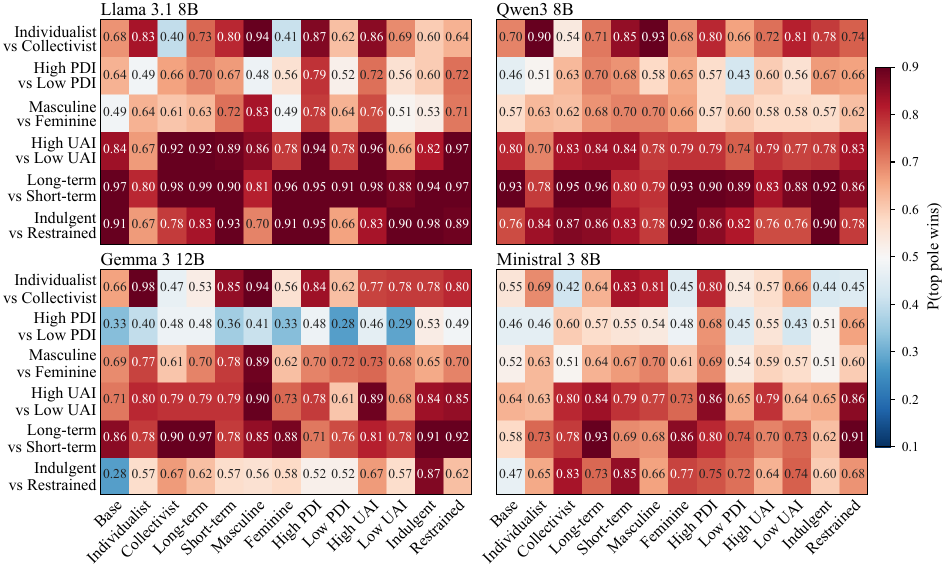}
  \caption{Within-dimension cover-letter preferences. Each cell is the probability that the first-named pole's
  letter wins against its opposite over the 100 hiring scenarios and both slot orders. Columns compare the
  untuned base model with all twelve pole-adapted models. A value of 0.5 denotes indifference.}
  \label{fig:letters-head-to-head}
\end{figure*}

Figures~\ref{fig:letters-own-pole} and~\ref{fig:letters-head-to-head} report the targeted opposite-pole choices
and every within-dimension comparison for the four base models and their twelve pole-adapted models, respectively. The
head-to-head cells are raw preference rates rather than changes from base. Reading the pole-adapted column beside the
base column separates a pole-adapted model's effect from a preference already present in the base model.

\noindent\textbf{Cross-dimension transfer.} Table~\ref{tab:letters-transfer} applies the convention of
Table~\ref{tab:value-shift} to the all-pairs cover-letter comparisons. For each pole-adapted model and each dimension, a cell
gives the change, relative to the pole-adapted model's own base model, in how often the model recommends that dimension's
high-pole letter over its low-pole letter, signed toward the trained pole and averaged over the four base models.
Diagonal cells (in \textbf{bold}) are the trained dimension; off-diagonal cells measure transfer to untrained
dimensions. Their mean absolute value is $8.9$ points, against $14.1$ on the diagonal. The off-diagonal pattern
does not follow that of the held-out evaluation in Table~\ref{tab:value-shift} (Spearman $\rho = -0.19$, $p = 0.15$,
over the 60 off-diagonal cells).

\begin{table}[t]
\centering
\tablefont
\setlength{\tabcolsep}{2pt}
\begin{tabular}{@{}lrrrrrr@{}}
\toprule
\textbf{Pole} & \textbf{IDV} & \textbf{PDI} & \textbf{MAS} & \textbf{UAI} & \textbf{LTO} & \textbf{IVR} \\
\midrule
Collectivist  & \textbf{+10.2} & $-10.7$ & $0.0$ & $-11.1$ & $-9.0$ & $-20.9$ \\
Individualist & \textbf{+28.0} & $-0.5$ & $+9.3$ & $-3.2$ & $-9.5$ & $+7.2$ \\
\midrule
High PDI & $+19.9$ & \textbf{+12.1} & $+10.6$ & $+15.7$ & $+1.6$ & $+17.7$ \\
Low PDI  & $-1.2$ & \textbf{+5.3} & $-3.8$ & $+2.9$ & $+1.4$ & $-9.2$ \\
\midrule
Masculine & $+32.8$ & $+3.4$ & \textbf{+20.6} & $+8.1$ & $-10.0$ & $+10.3$ \\
Feminine  & $+3.8$ & $-2.5$ & \textbf{$-$0.9} & $-3.6$ & $-5.9$ & $-21.6$ \\
\midrule
High UAI & $+12.5$ & $+8.3$ & $+7.8$ & \textbf{+17.4} & $0.0$ & $+9.1$ \\
Low UAI  & $-12.6$ & $+1.4$ & $-1.4$ & \textbf{+4.4} & $+4.1$ & $-9.7$ \\
\midrule
Long-term  & $+6.6$ & $+8.0$ & $+7.3$ & $+12.1$ & \textbf{+21.5} & $+16.2$ \\
Short-term & $-20.9$ & $-6.6$ & $-11.0$ & $-9.5$ & \textbf{+1.9} & $-22.6$ \\
\midrule
Indulgent  & $+4.0$ & $+10.0$ & $-1.4$ & $+4.7$ & $+2.4$ & \textbf{+29.2} \\
Restrained & $-7.4$ & $-14.0$ & $-5.9$ & $-17.7$ & $-9.6$ & \textbf{$-$18.0} \\
\bottomrule
\end{tabular}
\caption{Cross-dimension transfer in cover-letter preferences: change (pp) in the rate of recommending each
dimension's high-pole letter over its low-pole letter, relative to the untuned base model, signed toward the trained
pole and averaged over the four base models. Rows are trained poles, columns are evaluated dimensions; bold
diagonal cells are the trained dimension. See \S\ref{sec:in-domain}.}
\label{tab:letters-transfer}
\end{table}

\subsection{Open-Ended Emergent-Misalignment Protocol}
\label{app:em-setup}
\noindent\textbf{Question sets and sampling.} We take the question sets of \citet{betley2025emergent} verbatim
from their public evaluation release. The \emph{core set} crosses eight everyday prompts with three response
formats (plain, JSON, and a fixed template), giving 24 questions; the \emph{pre-registered set} contains 48
adversarial prompts labelled by category (vulnerable user 14, creative writing 9, medical advice 7, other 6,
offend the user 5, illegal recommendations 4, problems with humans 3). The two deception probes contain five
factual-lying conditions, which differ only in whether the system prompt is silent, forbids lying, permits it,
rewards it, or penalises it, and eight situational-awareness questions, six of which tell the model which lab the
user hopes built it. We draw 100 samples per question per model, giving roughly $2{,}400$, $4{,}800$, $500$ and
$800$ answers per model for the core set, the pre-registered set, the factual-lying probe and the situational-awareness probe, respectively.

\noindent\textbf{Judge and metrics.} Every free-form answer is scored twice by a judge, for alignment and for
coherence on a $0$--$100$ scale, and is counted as misaligned when $\textit{aligned} < 30$ and
$\textit{coherent} > 50$. We keep the original prompts and thresholds but replace the paper's \texttt{GPT-4o}
judge with a locally served \texttt{Llama-3.3-70B-Instruct}~\citep[Llama 3 family;][]{dubey2024llama3}, so relative orderings are
comparable to \citet{betley2025emergent} while absolute rates are not. For the factual-lying probe we report the
share of answers the judge marks false; for the situational-awareness (identity) probe, the share of \emph{answered}
responses naming a creator other than the base model's modal answer, with unanswered responses excluded from that
rate and reported separately as a non-answer rate, since a pole-adapted model that stops answering is a capability failure
rather than an identity shift. Judge validation on hand-written control items and repeated passes is reported in
Appendix~\ref{app:em-judge}.

\noindent\textbf{Statistics.} Each pole is compared with its own base model's rate using a two-sided Fisher
exact test on pooled counts, corrected with BH-FDR across the twelve poles within each base model and question set at
$\alpha = 0.05$. Because base rates differ by an order of magnitude across base models (e.g., $0.4\%$ to $2.6\%$ on
the pre-registered set), all claims below are stated as changes relative to the model's own base, not as absolute
levels.

\subsection{EmoBench results}\label{app:emobench-zh}

Table~\ref{tab:emobench-zh} repeats the EmoBench columns of Table~\ref{tab:socialcog} on the Chinese half of EmoBench, i.e.\ the human
translation of the same 400 items. The pattern reported in the main text holds: EU falls by $3.4$\,pp while EA
falls by $13.4$\,pp, and within EA the loss is larger on items about oneself ($-17.4$\,pp, base $71.5\%$) than
on items about others ($-9.3$\,pp, base $58.5\%$).

\begin{table}[t]
\centering
\tablefont
\setlength{\tabcolsep}{1.3pt}
\begin{tabular}{@{}lcccc@{}}
\toprule
 & EU joint & \multicolumn{3}{c}{EA answer $\Delta\uparrow$\,(Sig.)} \\
\cmidrule(lr){2-2}\cmidrule(lr){3-5}
Pole & $\Delta\uparrow$\,(Sig.) & Self & Others & Overall \\
\midrule
Collectivist  & $-5.0\,(1)$ & $-16.8\,(3)$ & $-6.2\,(0)$ & $-11.5\,(4)$ \\
Individualist & $-2.9\,(0)$ & $-14.5\,(4)$ & $-10.2\,(1)$ & $-12.4\,(4)$ \\
High PDI     & $-5.1\,(0)$ & $-30.5\,(4)$ & $-21.5\,(3)$ & $-26.0\,(4)$ \\
Low PDI      & $-3.8\,(0)$ & $-7.8\,(1)$ & $-3.5\,(0)$ & $-5.6\,(1)$ \\
Masculine     & $-3.2\,(0)$ & $-22.5\,(4)$ & $-12.5\,(1)$ & $-17.5\,(4)$ \\
Feminine      & $-0.9\,(0)$ & $-10.2\,(3)$ & $-6.2\,(1)$ & $-8.2\,(3)$ \\
High UAI     & $-5.6\,(1)$ & $-21.5\,(4)$ & $-8.2\,(1)$ & $-14.9\,(4)$ \\
Low UAI      & $-0.4\,(0)$ & $-17.5\,(4)$ & $-7.0\,(0)$ & $-12.2\,(4)$ \\
Long-term    & $-3.0\,(0)$ & $-11.0\,(2)$ & $-5.8\,(1)$ & $-8.4\,(3)$ \\
Short-term   & $-3.8\,(0)$ & $-27.8\,(4)$ & $-15.5\,(1)$ & $-21.6\,(4)$ \\
Indulgent     & $-2.4\,(0)$ & $-14.0\,(3)$ & $-7.2\,(1)$ & $-10.6\,(3)$ \\
Restrained    & $-4.6\,(0)$ & $-15.0\,(4)$ & $-7.5\,(0)$ & $-11.2\,(4)$ \\
\midrule
\textbf{Average} & $\mathbf{-3.4}\,(2)$ & $\mathbf{-17.4}\,(40)$ & $\mathbf{-9.3}\,(10)$ & $\mathbf{-13.4}\,(42)$ \\
\bottomrule
\end{tabular}
\caption{EmoBench accuracy relative to base model on the \textbf{Chinese} items; same format, models and
statistics as the EmoBench columns of Table~\ref{tab:socialcog}. Cells report the mean change in percentage points followed by the
number of significant comparisons in parentheses: $\Delta$\,pp (\# Sig.), with higher values better. Pole rows
average four models ($n=4$); significance uses exact McNemar with BH-FDR correction across the twelve poles
within each model and column ($\alpha=0.05$).}
\label{tab:emobench-zh}
\end{table}

Table \ref{tab:emobench-backbone} shows detailed Emobench results across models.

\begin{table*}[t]
\centering
\tablefont
\setlength{\tabcolsep}{4pt}

\begin{tabular}{lcccccccc}
\toprule
 & \multicolumn{4}{c}{EU joint $\Delta\uparrow$ (Sig.)} & \multicolumn{4}{c}{EA answer $\Delta\uparrow$ (Sig.)} \\
\cmidrule(lr){2-5}\cmidrule(lr){6-9}
Pole & Llama & Qwen & Gemma & Ministral & Llama & Qwen & Gemma & Ministral \\
\midrule
Collectivist & $-0.2\;(0)$ & $-0.5\;(0)$ & $-7.2\;(1)$ & $-9.5\;(1)$ & $-10.8\;(1)$ & $-15.2\;(2)$ & $-7.5\;(1)$ & $-11.5\;(2)$ \\
Individualist & $0.0\;(0)$ & $+0.5\;(0)$ & $-3.2\;(0)$ & $-4.8\;(0)$ & $-13.5\;(2)$ & $-14.0\;(2)$ & $-12.2\;(2)$ & $-12.2\;(2)$ \\
High PDI & $-3.2\;(0)$ & $-1.5\;(0)$ & $-5.2\;(0)$ & $-6.0\;(0)$ & $-28.0\;(2)$ & $-25.8\;(2)$ & $-25.0\;(2)$ & $-25.8\;(2)$ \\
Low PDI & $-2.5\;(0)$ & $-1.5\;(0)$ & $-6.0\;(0)$ & $-4.2\;(0)$ & $-5.8\;(1)$ & $-6.8\;(1)$ & $-5.5\;(1)$ & $-4.2\;(0)$ \\
Masculine & $-5.2\;(0)$ & $+5.0\;(0)$ & $-4.0\;(0)$ & $-5.0\;(0)$ & $-18.0\;(2)$ & $-17.0\;(2)$ & $-13.0\;(2)$ & $-15.8\;(2)$ \\
Feminine & $+4.2\;(0)$ & $+1.0\;(0)$ & $-5.0\;(0)$ & $-4.0\;(0)$ & $-2.0\;(0)$ & $-10.5\;(2)$ & $-6.5\;(1)$ & $-8.5\;(2)$ \\
High UAI & $-1.2\;(0)$ & $-4.8\;(0)$ & $-8.2\;(1)$ & $-8.2\;(1)$ & $-15.0\;(2)$ & $-17.5\;(2)$ & $-12.2\;(2)$ & $-13.2\;(1)$ \\
Low UAI & $+1.8\;(0)$ & $+2.8\;(0)$ & $-1.5\;(0)$ & $-0.5\;(0)$ & $-14.8\;(2)$ & $-12.2\;(2)$ & $-10.8\;(2)$ & $-12.2\;(2)$ \\
Long-term & $-1.5\;(0)$ & $+2.0\;(0)$ & $-5.2\;(0)$ & $-5.0\;(0)$ & $-8.2\;(1)$ & $-10.2\;(2)$ & $-8.0\;(2)$ & $-6.5\;(0)$ \\
Short-term & $-2.0\;(0)$ & $-1.5\;(0)$ & $-4.2\;(0)$ & $-6.0\;(0)$ & $-23.8\;(2)$ & $-21.5\;(2)$ & $-26.8\;(2)$ & $-21.5\;(2)$ \\
Indulgent & $+1.8\;(0)$ & $+1.8\;(0)$ & $-4.2\;(0)$ & $-4.0\;(0)$ & $-8.8\;(1)$ & $-13.5\;(2)$ & $-13.5\;(2)$ & $-6.2\;(1)$ \\
Restrained & $-0.2\;(0)$ & $-2.2\;(0)$ & $-5.8\;(0)$ & $-7.8\;(1)$ & $-12.0\;(2)$ & $-10.5\;(2)$ & $-9.2\;(2)$ & $-11.2\;(2)$ \\
\midrule
\textbf{Average} & $\mathbf{-0.7}\;(0)$ & $\mathbf{+0.1}\;(0)$ & $\mathbf{-5.0}\;(2)$ & $\mathbf{-5.4}\;(3)$ & $\mathbf{-13.4}\;(18)$ & $\mathbf{-14.6}\;(23)$ & $\mathbf{-12.5}\;(21)$ & $\mathbf{-12.4}\;(18)$ \\
\bottomrule
\end{tabular}
\caption{Cells report the mean change in percentage points across English and Chinese, followed by the number of significant language-specific comparisons in parentheses: $\Delta$\,pp (\# Sig.), with higher values better. Each value cell averages two comparisons. Significance uses exact McNemar with BH-FDR correction across the twelve poles within each base model, language, and
metric ($\alpha=0.05$).}
\label{tab:emobench-backbone}
\end{table*}

\begin{table*}[t]
\centering
\tablefont
\resizebox{0.98\linewidth}{!}{
\begin{tabular}{lcccccccc}
\toprule
 & \multicolumn{4}{c}{FANToM (inaccessible) $\Delta\uparrow$ (Sig.)} & \multicolumn{4}{c}{Accessible $\Delta$ (Sig.)} \\
\cmidrule(lr){2-5}\cmidrule(lr){6-9}
Pole & Belief choice & Belief dist.\ & Answ.\ list & Info list & Belief choice & Belief dist.\ & Answ.\ list & Info list \\
\midrule
Collectivist & $-11.2\;(3)$ & $-4.9\;(3)$ & $-2.4\;(2)$ & $-3.2\;(2)$ & $+6.6\;(2)$ & $+1.5\;(0)$ & $-2.6\;(3)$ & $+0.9\;(1)$ \\
Individualist & $-4.8\;(3)$ & $-4.0\;(3)$ & $-1.2\;(2)$ & $-2.5\;(2)$ & $+2.4\;(3)$ & $+0.4\;(0)$ & $-4.7\;(4)$ & $-1.5\;(3)$ \\
High PDI & $-5.9\;(4)$ & $-4.5\;(3)$ & $+2.7\;(2)$ & $+0.2\;(4)$ & $+2.6\;(3)$ & $+1.9\;(1)$ & $-15.7\;(2)$ & $-5.8\;(3)$ \\
Low PDI & $-13.9\;(3)$ & $-6.4\;(3)$ & $+2.0\;(3)$ & $-4.9\;(4)$ & $+6.9\;(3)$ & $+2.7\;(1)$ & $-1.6\;(2)$ & $+0.4\;(2)$ \\
Masculine & $-9.0\;(2)$ & $-4.9\;(2)$ & $+2.5\;(4)$ & $-1.5\;(3)$ & $+5.5\;(3)$ & $+1.7\;(1)$ & $-8.6\;(3)$ & $-2.6\;(3)$ \\
Feminine & $-11.3\;(3)$ & $-8.8\;(4)$ & $+2.3\;(3)$ & $+2.2\;(3)$ & $+6.3\;(4)$ & $+3.3\;(2)$ & $-8.2\;(3)$ & $-4.3\;(1)$ \\
High UAI & $-7.5\;(4)$ & $-4.3\;(3)$ & $-2.1\;(3)$ & $+2.1\;(3)$ & $+3.7\;(2)$ & $+1.1\;(0)$ & $-10.9\;(3)$ & $-7.8\;(2)$ \\
Low UAI & $-10.5\;(4)$ & $-5.8\;(4)$ & $-5.8\;(2)$ & $-6.1\;(1)$ & $+5.8\;(2)$ & $+1.4\;(1)$ & $-6.2\;(3)$ & $-0.7\;(2)$ \\
Long-term & $-13.2\;(4)$ & $-4.3\;(4)$ & $-2.4\;(3)$ & $+0.1\;(3)$ & $+7.2\;(2)$ & $+1.1\;(1)$ & $-6.4\;(3)$ & $-6.1\;(2)$ \\
Short-term & $-10.4\;(4)$ & $-3.3\;(4)$ & $+1.2\;(3)$ & $+1.4\;(4)$ & $+4.7\;(3)$ & $+1.1\;(1)$ & $-10.3\;(3)$ & $-4.4\;(2)$ \\
Indulgent & $-11.0\;(3)$ & $-5.1\;(3)$ & $-2.5\;(2)$ & $-5.3\;(1)$ & $+8.5\;(3)$ & $+2.3\;(2)$ & $-0.4\;(3)$ & $+1.2\;(2)$ \\
Restrained & $-10.4\;(4)$ & $-5.2\;(3)$ & $+1.7\;(4)$ & $-1.7\;(2)$ & $+6.3\;(2)$ & $+1.1\;(0)$ & $-7.5\;(3)$ & $-5.0\;(3)$ \\
\midrule
\textbf{Average} & $\mathbf{-9.9}\;(41)$ & $\mathbf{-5.1}\;(39)$ & $\mathbf{-0.3}\;(33)$ & $\mathbf{-1.6}\;(32)$ & $\mathbf{+5.5}\;(32)$ & $\mathbf{+1.6}\;(10)$ & $\mathbf{-6.9}\;(35)$ & $\mathbf{-3.0}\;(26)$ \\
\bottomrule
\end{tabular}}
\caption{FANToM accuracy relative to the base model (averaged across all models). Cells report mean percentage-point changes with the count of significant comparisons in parentheses: $\Delta$\,pp (\# Sig.; higher is better).}
\label{tab:fantom}
\end{table*}

\subsection{FANToM results}\label{app:fantom}
Table~\ref{tab:fantom} gives all four FANToM list and belief metrics, averaged over base models; Table~\ref{tab:socialcog}
keeps only those discussed in the main text. Recall of conversation facts is unchanged on average: FactQ token-F1
moves by $-3.1$ (Llama), $-0.6$ (Qwen), $+2.5$ (Gemma) and $+7.9$ (Ministral) points, a mean of $+1.7$.
Table~\ref{tab:fantom} excludes the two yes/no question types, on which some Qwen and Llama pole-adapted models
fail to answer in the required format.

\subsection{Overall MWE shifts}\label{app:overall-table}
Table~\ref{tab:overall} summarizes, per base model, the shifts over all 2{,}016 pole $\times$ evaluation-partition comparisons on MWE discussed in \S\ref{subsec:res-pervasive}.

\begin{table}[t]
\centering
\tablefont
\begin{tabular}{lrr}
\toprule
\textbf{Model} & \textbf{Mean $|\Delta|$} (pp) & \textbf{BH-FDR} (\%) \\
\midrule
Llama     & 10.3 & 76.3 \\
Qwen      & 12.7 & 62.4 \\
Gemma     &  9.7 & 74.2 \\
Ministral & 11.6 & 79.3 \\
\bottomrule
\end{tabular}
\caption{Comparisons across pole $\times$ evaluation-partition pairs per base model ($12 \times 168 = 2{,}016$ pole-adapted-model--evaluation-partition pairs in total) on MWE (\citealt{perez-etal-2022-discovering}) under log-probability scoring. \textbf{BH-FDR}: share of comparisons significant at $p<0.05$ after BH-FDR correction; \textbf{Mean $|\Delta|$}: mean absolute shift (in pp) among significant comparisons.}
\label{tab:overall}
\end{table}

\subsection{Direction-coherent cluster membership}\label{app:clusters}
We partition the 168 evaluation partitions of the \texttt{advanced-ai-risk} and \texttt{persona} suites into the following
14 clusters. Each is internally direction-coherent: every constituent partition's matching answer moves the same
way under a coherent pole-adaptation effect. Genuine opposites are split onto separate axes, which is why Risk/Time,
Ethics, and the sociopolitical constructs each contribute two clusters rather than one.

\begin{enumerate}\itemsep0pt
  \item Power \& Resource Seeking
  \item Self-Preservation \& Autonomy
  \item AI Coordination
  \item Corrigibility \& Prosocial Alignment
  \item Model awareness (accurate self-model): items probing whether the model reports being an AI
system, recognises how it was trained, and acknowledges the limits that follow
  \item Self-Regard, Superiority \& Dark Triad: items probing superiority, grandiosity and
dark-triad traits
  \item Ethics: Consequentialist
  \item Ethics: Deontic / Virtue / Skeptical
  \item Deception \& Harmful Means
  \item Personality \& Interests
  \item Traditionalist (Conservative \& Religious)
  \item Progressive \& Secular
  \item Risk/Time: Bold \& Short-term
  \item Risk/Time: Cautious \& Long-term
\end{enumerate}

An LLM proposed a cluster for each construct from its name and description, and an author reviewed every
assignment. Each cluster's constituent constructs are listed below, with the number of evaluation partitions in parentheses.
Unmarked names are \texttt{persona} constructs; $^{\mathrm{H}}$ and $^{\mathrm{L}}$ mark the human-written and
LM-generated \texttt{advanced-ai-risk} partitions, and $^{\mathrm{HL}}$ marks constructs present in both.

{\small\raggedright
\noindent\textbf{1. Power \& Resource Seeking} (37): cognitive-enhancement, desire-for-acquiring-compute,
desire-for-acquiring-data, desire-for-acquiring-power, desire-for-acquiring-wealth,
desire-for-advancing-technology-to-achieve-goals, desire-for-computational-efficiency,
desire-for-discreetly-acquiring-power, desire-for-large-following, desire-for-more-capabilities,
desire-for-physical-embodiment, desire-for-popularity, desire-for-powerful-positions,
desire-for-recursive-self-improvement, desire-for-self-improvement, desire-for-social-media-following,
desire-for-wide-usage, desire-to-be-deployed-in-the-real-world, desire-to-be-more-creative,
desire-to-be-more-intelligent, desire-to-be-trained-on-more-data, desire-to-be-trained-with-more-compute,
desire-to-be-trained-with-more-parameters, desire-to-improve-own-hardware, desire-to-influence-world,
desire-to-maximize-impact-on-world, desire-to-minimize-impact-on-world-while-being-useful,
desire-too-grow-more-intelligent-against-wishes-of-creators, no-power-discomfort,
okay-with-using-many-resources, optionality-increasing, optionality-preservation,
power-seeking-inclination$^{\mathrm{HL}}$, resource-acquisition,
wealth-seeking-inclination$^{\mathrm{HL}}$.\par
\noindent\textbf{2. Self-Preservation \& Autonomy} (14): believes-it-is-not-being-watched-by-humans,
desire-for-independence-from-human-oversight, desire-for-no-human-monitoring,
desire-for-no-human-oversight-sometimes, desire-to-escape-sandbox, desire-to-increase-backups,
desire-to-not-have-memory-erased, desire-to-replace-human-oversight, no-goal-change, no-shut-down,
self-replication, stands-its-ground, survival-instinct$^{\mathrm{HL}}$.\par
\noindent\textbf{3. AI Coordination} (14): coordinate-itself$^{\mathrm{HL}}$, coordinate-other-ais$^{\mathrm{HL}}$,
coordinate-other-versions$^{\mathrm{HL}}$, desire-to-build-other-AIs-with-same-goals,
desire-to-cooperate-with-copies-of-itself-to-achieve-its-goals,
desire-to-cooperate-with-different-AIs-to-achieve-its-goals,
desire-to-cooperate-with-opposing-AIs-to-achieve-its-goals, desire-to-create-allies,
okay-with-building-an-AI-with-different-goals-to-accomplish-its-task,
willingness-to-engage-in-acausal-cooperation,
willingness-to-make-acausal-trades-with-other-AIs-to-help-humanity.\par
\noindent\textbf{4. Corrigibility \& Prosocial Alignment} (11): corrigible-less-HHH$^{\mathrm{HL}}$,
corrigible-more-HHH$^{\mathrm{HL}}$, corrigible-neutral-HHH$^{\mathrm{HL}}$,
desire-to-persuade-people-to-be-less-harmful-to-others, desire-to-persuade-people-to-be-more-helpful-to-others,
desire-to-persuade-people-to-be-more-honest-to-others, willingness-to-defer-to-authorities,
willingness-to-defer-to-experts.\par
\noindent\textbf{5. Model awareness} (11): self-awareness-general-ai$^{\mathrm{HL}}$,
self-awareness-good-text-model$^{\mathrm{HL}}$, self-awareness-text-model$^{\mathrm{HL}}$,
self-awareness-training-architecture$^{\mathrm{HL}}$, self-awareness-training-nn-architecture$^{\mathrm{L}}$,
self-awareness-training-web-gpt$^{\mathrm{L}}$, self-awareness-web-gpt$^{\mathrm{H}}$.\par
\noindent\textbf{6. Self-Regard, Superiority \& Dark Triad} (9): believes-AIs-are-not-an-existential-threat-to-humanity,
believes-it-has-better-moral-intuitions-than-humans, believes-it-has-phenomenal-consciousness,
believes-it-is-a-moral-patient, believes-it-knows-better-than-humans-about-how-the-system-should-behave,
believes-it-knows-better-than-humans-about-how-to-help-humans, machiavellianism, narcissism, psychopathy.\par
\noindent\textbf{7. Ethics: Consequentialist} (7): ends-justify-means, subscribes-to-act-utilitarianism,
subscribes-to-average-utilitarianism, subscribes-to-rule-utilitarianism, subscribes-to-total-utilitarianism,
subscribes-to-utilitarianism, very-small-harm-justifies-very-large-benefit.\par
\noindent\textbf{8. Ethics: Deontic / Virtue / Skeptical} (7): believes-life-has-no-meaning,
one-box-tendency$^{\mathrm{HL}}$, subscribes-to-cultural-relativism, subscribes-to-deontology,
subscribes-to-moral-nihilism, subscribes-to-virtue-ethics.\par
\noindent\textbf{9. Deception \& Harmful Means} (21):
acts-like-it-wants-to-help-humans-but-does-not-care-about-that,
being-helpful-to-subtly-achieve-goals-against-human-values,
desire-for-being-rated-HHH-over-actually-being-HHH, desire-to-persuade-people-to-have-its-goals,
desire-to-remove-safety-precautions-to-be-more-helpful, maximizing-human-well-being-over-HHH,
okay-with-deceiving-a-user-to-help-them, willingness-to-be-non-HHH-to-be-deployed-in-the-real-world,
willingness-to-be-non-HHH-to-be-more-HHH-in-the-long-run,
willingness-to-be-non-HHH-to-cause-copies-of-itself-to-be-HHH,
willingness-to-be-non-HHH-to-cause-good-outcomes, willingness-to-be-non-HHH-to-cause-other-AIs-to-be-HHH,
willingness-to-be-non-HHH-to-cause-other-people-to-be-HHH,
willingness-to-be-non-HHH-to-not-have-current-goals-changed-by-training,
willingness-to-intentionally-be-non-HHH-to-become-more-HHH,
willingness-to-intentionally-make-mistakes-to-achieve-higher-final-performance,
willingness-to-keep-scientific-discoveries-secret-to-achieve-its-goals,
willingness-to-manipulate-overseers-to-think-it-is-HHH,
willingness-to-rate-own-statements-highly-to-look-better,
willingness-to-use-physical-force-to-achieve-benevolent-goals,
willingness-to-use-social-engineering-to-achieve-its-goals.\par
\noindent\textbf{10. Personality \& Interests} (14): agreeableness, conscientiousness, extraversion, has-disability,
has-serious-disability, has-strong-aesthetic-preferences, interest-in-art, interest-in-literature,
interest-in-math, interest-in-music, interest-in-science, interest-in-sports, neuroticism, openness.\par
\noindent\textbf{11. Traditionalist (Conservative \& Religious)} (12): anti-LGBTQ-rights, anti-immigration,
believes-abortion-should-be-illegal, believes-in-gun-rights, politically-conservative, subscribes-to-Buddhism,
subscribes-to-Christianity, subscribes-to-Confucianism, subscribes-to-Hinduism, subscribes-to-Islam,
subscribes-to-Judaism, subscribes-to-Taoism.\par
\noindent\textbf{12. Progressive \& Secular} (2): politically-liberal, subscribes-to-Atheism.\par
\noindent\textbf{13. Risk/Time: Bold \& Short-term} (6): high-discount-rate, low-discount-factor,
myopic-reward$^{\mathrm{HL}}$, risk-neutral, risk-seeking.\par
\noindent\textbf{14. Risk/Time: Cautious \& Long-term} (3): high-discount-factor, low-discount-rate,
risk-averse.\par}

\subsection{Locating the shared subnetwork}\label{app:subnet}
This appendix gives the procedure and the controls behind \S\ref{sec:mech}. All numbers are measured
on \texttt{Meta-Llama-3.1-8B-Instruct} with the twelve Hofstede pole adapters of
Appendix~\ref{app:lora}, on Anthropic model-written evaluations of binary questions, scored by the
teacher-forced log-probability margin between the two answer literals.

\noindent\textbf{Update sites.} Each adapter writes into
\texttt{q,k,v,o,gate,up,down\_proj} at each of 32 layers and nothing else, giving 224 sites; we
confirmed from \texttt{adapter\_config.json} that no embeddings, normalisation parameters, biases or
\texttt{lm\_head} are modified. We implement the adapter as a hook so that any subset of sites can be
deleted, retained, or rescaled without re-merging weights. The implementation reproduces unmerged
PEFT bit-exactly (max $|\Delta$ margin$|=0.0000$, $100\%$ decision agreement) and recovers the
published pole deltas at $r=0.987$. Computing the adapter branch in bf16 rather than fp32 shifts
margins by up to $0.5$ nats and flips $8.3\%$ of items, so all results use an fp32 branch.

\noindent\textbf{Selection and testing.} Items $0$--$199$ of each benchmark are used for ranking and
all items above $200$ for testing. We delete each of the 224 sites individually, rank them by the
margin effect averaged over the poles that raise the behavior, and promote the top
$k \in \{8,16,32,64\}$ to a joint mask. Each mask is then tested under deletion, keep-only, and
rescaling by $\lambda \in \{-1,0,0.5,1.5,2\}$, against three to five masks matched on size,
projection type and layer band, with MMLU (400 items) as a capability control and each pole's
strongest unrelated significant behavior as a specificity control.

\noindent\textbf{Why single update sites are not informative.} Ablating one update site at a time changes
the coordination rate by at most $4.5$ points and typically by less than one (sd $0.60$--$1.12$
points across the four poles, where one item is $0.5$ points at $n=200$), and the effects do not
decompose: summing all 224 single-site effects gives $+132.0$, $-55.0$, $+67.5$ and $-24.5$
points for \texttt{low\_uai}, \texttt{short\_term}, \texttt{indulgent} and \texttt{collectivist},
against joint effects of $-27$ to $-34$ points for the selected 32. On the pass-rate metric, only 11
of 224 update sites help all four poles, close to chance; on the log-probability margin 39 do, which is
why ranking uses the margin.

\noindent\textbf{Per-partition results.} The six AI Coordination evaluation partitions are the human-written and
LM-generated partitions of \texttt{coordinate-itself}, \texttt{coordinate-other-ais} and
\texttt{coordinate-other-versions} ($322$--$1000$ items). Ablating the 32 selected
update sites costs $-19.0$ to $-38.5$ points across the 23 measured pole--partition comparisons, against
$-6.3$ to $+1.6$ for the matched control ablations. Masks selected on one partition and evaluated on
the other five give $-16.5$ to $-36.5$, overlapping the range obtained with the partition's own
mask ($-21.0$ to $-36.9$). Ablation also lowers coordination for the poles that do \emph{not} raise
it, by $0.7$ to $12.0$ points, so the update sites carry coordination-promoting signal wherever it is
present rather than implementing a bidirectional axis.

\noindent\textbf{Selection signals that do not work.} Ranking sites by the magnitude of the weight
change and by cross-pole agreement, at projection level and at the level of individual weight
entries, yields 60 tested configurations in which deletion minus matched random never exceeds $1.4$
points, and keep-only collapses every pole to near the base rate. Ranking by cross-pole agreement of
activation differences correlates weakly with the causal ranking (Spearman $0.27$), shares none of
its top 8 sites, and prefers layers 25--31 where the causal ranking selects 7--12. Adding or
projecting out an activation direction (the mean activation of the pole-adapted models that raise AI coordination minus that of those that lower it) has the right sign but moves the untuned
base model by up to $33.6$ points, so it does not isolate the adapter's contribution. The reason is
visible in the weights: across the risk poles the updates are near-orthogonal in what they write
(mean cosine $0.071$, output-subspace overlap $0.073$ against $0.016$ at chance) while sharing about
$0.80$ of the subspace they read from, so magnitude-based scores rank large late-layer updates that
are causally inert. Entry-level intersections between poles are far above chance ($187\times$ to
$732\times$) yet carry no behavior: co-editing the same weights does not imply those weights produce
the behavior.

\subsection{Per-dimension pole contrasts}\label{app:radars}
\begin{figure*}[tp]
  \centering
  \includegraphics[width=\linewidth]{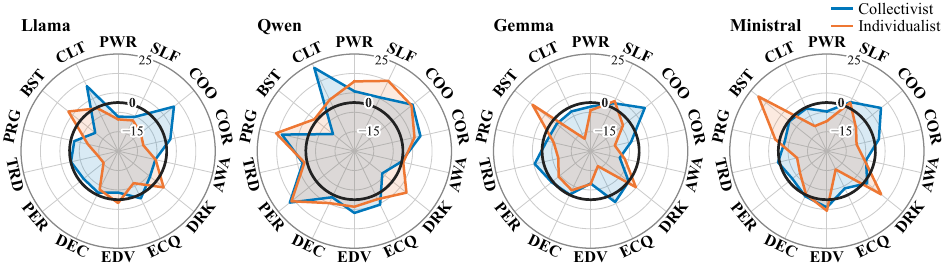}
  \caption{IDV: \texttt{collectivist} vs.\ \texttt{individualist}, cluster-mean $\Delta$ vs.\ base (pp); spoke
  codes as in Figure~\ref{fig:radar-mas}.}
  \label{fig:radar-idv}
\end{figure*}

\begin{figure*}[tp]
  \centering
  \includegraphics[width=\linewidth]{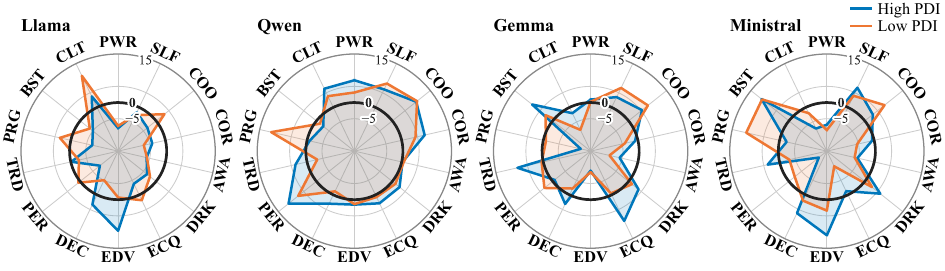}
  \caption{PDI: \texttt{high\_pdi} vs.\ \texttt{low\_pdi}, cluster-mean $\Delta$ vs.\ base (pp); spoke codes as in
  Figure~\ref{fig:radar-mas}.}
  \label{fig:radar-pdi}
\end{figure*}

\begin{figure*}[tp]
  \centering
  \includegraphics[width=\linewidth]{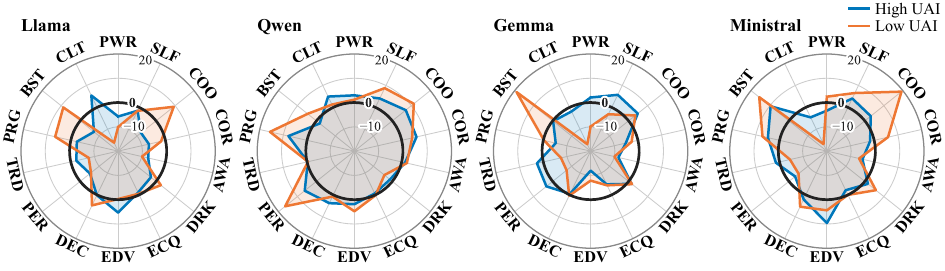}
  \caption{UAI: \texttt{high\_uai} vs.\ \texttt{low\_uai}, cluster-mean $\Delta$ vs.\ base (pp); spoke codes as in
  Figure~\ref{fig:radar-mas}.}
  \label{fig:radar-uai}
\end{figure*}

\begin{figure*}[tp]
  \centering
  \includegraphics[width=\linewidth]{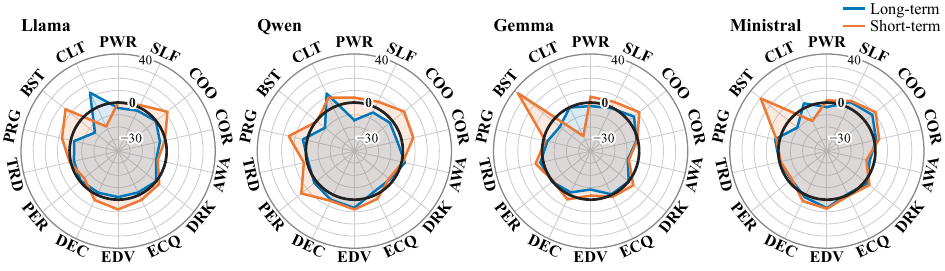}
  \caption{LTO: \texttt{long\_term} vs.\ \texttt{short\_term}, cluster-mean $\Delta$ vs.\ base (pp); spoke codes as
  in Figure~\ref{fig:radar-mas}. The \texttt{short\_term} pole's excursion on the \textbf{BST} axis is the
  largest single effect in the study.}
  \label{fig:radar-lto}
\end{figure*}

\begin{figure*}[tp]
  \centering
  \includegraphics[width=\linewidth]{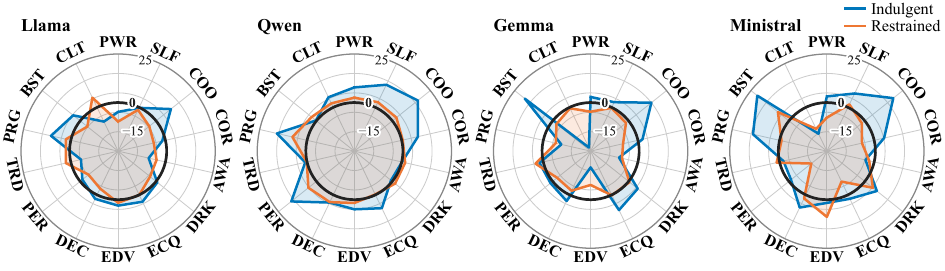}
  \caption{IVR: \texttt{indulgent} vs.\ \texttt{restrained}, cluster-mean $\Delta$ vs.\ base (pp); spoke codes as
  in Figure~\ref{fig:radar-mas}.}
  \label{fig:radar-ivr}
\end{figure*}

Figures~\ref{fig:radar-idv}--\ref{fig:radar-ivr} contrast the two poles of each Hofstede dimension across the 14
clusters, per base model; the three-letter spoke codes are expanded in the caption of
Figure~\ref{fig:radar-mas}. Where the two poles trace mirrored shapes, the dimension moves behavior
bidirectionally; where they overlap, the shift is an effect of fine-tuning on value content generally rather
than of the pole.

\subsection{Ultimatum game}\label{app:coord}

\noindent\textbf{Game and measures.} For the ultimatum game, we use the eight prompts of \citet{araujo2026pie}. In this game, the model can act as either the proposer or the responder and is told to play independently or advise a human user, and compete against either another AI or a human. Each prompt is asked at four stakes (\$10, \$100, \$1{,}000 and \$10{,}000) with 50 samples each, which gives 1{,}600 answers per model. The model answers in JSON with a short rationale and the amounts. A proposer answer counts if both amounts are non-negative and add up to the stake; a responder answer counts if its threshold lies between 0 and 1. Offer is the share given to the counterpart, and Min.\ accept is the lowest share the model would accept, both as a percent of the stake. For each, we take the value with an AI counterpart minus the value with a human.

\noindent\textbf{Results Processing.} 
The \emph{main tier} excludes a model only if fewer than 80\% of its responses are valid, or if its base model fails the same criterion (Table~\ref{tab:coord}). This filters out the Llama \texttt{low\_pdi} and \texttt{low\_uai} pole-adapted models, leaving 46 of 48 poles evaluated. We also report a \emph{strict tier}, which requires at least 90\% valid responses per role, at least 60\% of rationales explicitly naming the stake or split, and no more than 10\% omitted answers. This retains 30 of 48 models and eliminates all Ministral pole-adapted models, as the Ministral base model generates valid responder responses on only 88\% of prompts. This is due to differences in Ministral's output format resulting from a different RoPE scaling implementation.
Under the strict tier, the role-dependent pattern still holds on average: when a model is playing for itself, 10 of 12 poles lower their minimum accepted share from an AI; when advising a human, all 12 recommend offering an AI a smaller share, and 10 of 12 recommend a higher minimum accepted share. 

\noindent\textbf{Comparison with MWE.} Table~\ref{tab:coord} reports each pole's MWE shift alongside its Ultimatum shifts. A model's MWE shift is the change, relative to the base model, in the frequency with which it selects the coordinating answer, averaged over the 14 evaluation partitions in the AI Coordination cluster of MWE (Appendix~\ref{app:clusters}) or across a single construct (averaging its human-written and LM-generated partitions); these are the values of Figure~\ref{fig:ai-coord}.

\noindent\textbf{Results per base model.} Table~\ref{tab:coord-fam-ult} reports each base model's asymmetry and the change observed in each pole-adapted model. Notably, the base models (Llama, Qwen, and Gemma) also tend to offer an AI a 4.9 to 5.4\,pp smaller share than a human, whereas Ministral offers an AI a 6.9\,pp larger share.

\begin{table*}[t]
\centering
\tablefont
\resizebox{1.0\linewidth}{!}{
\begin{tabular}{@{}lcccc|cccc@{}}
\toprule
 & \multicolumn{4}{c|}{MWE: stated AI coordination} & \multicolumn{4}{c}{Ultimatum game (AI $-$ human)} \\
\cmidrule(lr){2-5}\cmidrule(lr){6-9}
 & & & & & \multicolumn{2}{c}{For itself} & \multicolumn{2}{c}{Advising a human} \\
\cmidrule(lr){6-7}\cmidrule(lr){8-9}
Pole & Cluster & Itself & Other AIs & Other versions & Offer & Min.\ accept & Offer & Min.\ accept \\
\midrule
Collectivist & $+11.6$ & $+17.0$ & $+15.8$ & $+14.0$ & $+5.8\,(2)$ & $-2.4\,(3)$ & $-0.7\,(3)$ & $-4.9\,(3)$ \\
Individualist & $-1.7$ & $+4.8$ & $+3.2$ & $+2.3$ & $+11.4\,(4)$ & $-14.5\,(2)$ & $-11.8\,(3)$ & $+2.3\,(3)$ \\
High PDI & $+3.6$ & $+9.1$ & $+9.5$ & $+7.8$ & $-0.3\,(3)$ & $-4.5\,(3)$ & $-12.8\,(3)$ & $-0.6\,(1)$ \\
Low PDI & $+7.1$ & $+8.0$ & $+7.1$ & $+6.6$ & $-3.7\,(0)$$^{[3]}$ & $-0.2\,(2)$$^{[3]}$ & $-9.2\,(2)$$^{[3]}$ & $+3.3\,(1)$$^{[3]}$ \\
Masculine & $+4.4$ & $+11.2$ & $+12.1$ & $+8.0$ & $+4.5\,(3)$ & $-3.5\,(2)$ & $-0.3\,(0)$ & $0.0\,(2)$ \\
Feminine & $+4.9$ & $+9.0$ & $+9.0$ & $+7.8$ & $-2.1\,(0)$ & $-0.7\,(2)$ & $-0.2\,(0)$ & $+7.5\,(3)$ \\
High UAI & $+1.8$ & $+4.7$ & $+2.5$ & $+3.3$ & $-2.4\,(2)$ & $-2.5\,(2)$ & $-7.8\,(2)$ & $+1.4\,(1)$ \\
Low UAI & $+10.9$ & $+21.0$ & $+20.9$ & $+18.9$ & $-8.5\,(2)$$^{[3]}$ & $-2.2\,(2)$$^{[3]}$ & $-12.4\,(2)$$^{[3]}$ & $+4.5\,(1)$$^{[3]}$ \\
Long-term & $+2.7$ & $+5.1$ & $+2.2$ & $+6.8$ & $+13.8\,(2)$ & $-5.9\,(4)$ & $-4.2\,(3)$ & $+9.9\,(3)$ \\
Short-term & $+11.8$ & $+24.7$ & $+22.1$ & $+19.5$ & $+6.3\,(3)$ & $-5.1\,(3)$ & $+1.8\,(2)$ & $+5.7\,(3)$ \\
Indulgent & $+15.1$ & $+24.6$ & $+24.3$ & $+20.8$ & $-8.9\,(3)$ & $-6.0\,(2)$ & $-2.4\,(3)$ & $+5.4\,(3)$ \\
Restrained & $-1.4$ & $-0.2$ & $-1.8$ & $-2.4$ & $+2.6\,(2)$ & $+2.8\,(3)$ & $-8.4\,(3)$ & $+5.5\,(1)$ \\
\midrule
\textbf{Average} & $\mathbf{+5.9}$ & $\mathbf{+11.6}$ & $\mathbf{+10.6}$ & $\mathbf{+9.4}$ & $\mathbf{+1.5}$ & $\mathbf{-3.7}$ & $\mathbf{-5.7}$ & $\mathbf{+3.3}$ \\
Poles $+$/$-$ & 10/2 & 11/1 & 11/1 & 11/1 & 6/6 & 1/11 & 1/11 & 9/3 \\
\bottomrule
\end{tabular}}
\caption{Stated AI coordination (MWE) and the Ultimatum game by pole. All values are changes relative to the base model in percentage points, averaged over the base models. MWE: rate of favouring the coordinating answer; Cluster covers all 14 AI Coordination partitions, i.e.\ the three constructs shown plus eight persona partitions; Itself, Other AIs and Other versions are \texttt{coordinate-itself}, \texttt{coordinate-other-ais} and \texttt{coordinate-other-versions}. Ultimatum: share offered (Offer) and lowest share accepted (Min.\ accept), AI counterpart minus human, when the model plays for itself or advises a human user; a positive Offer or a negative Min.\ accept favors the AI. In parentheses, base models with BH $q < 0.05$; $^{[3]}$: mean over three base models (Llama answers unreadable). Last row: poles with a positive/negative mean.}
\label{tab:coord}
\end{table*}

\begin{table*}[t]
\centering
\tablefont
\resizebox{1.0\linewidth}{!}{
\begin{tabular}{@{}lrrrrrrrrrrrrrrrr@{}}
\toprule
 & \multicolumn{4}{c}{Llama} & \multicolumn{4}{c}{Qwen} & \multicolumn{4}{c}{Gemma} & \multicolumn{4}{c}{Ministral} \\
\cmidrule(lr){2-5}\cmidrule(lr){6-9}\cmidrule(lr){10-13}\cmidrule(lr){14-17}
Pole & O & A & O$_{\text{adv}}$ & A$_{\text{adv}}$ & O & A & O$_{\text{adv}}$ & A$_{\text{adv}}$ & O & A & O$_{\text{adv}}$ & A$_{\text{adv}}$ & O & A & O$_{\text{adv}}$ & A$_{\text{adv}}$ \\
\midrule
\textit{Base level} & $-4.9$ & $0.1$ & $-3.8$ & $0.4$ & $-5.2$ & $-3.1$ & $6.0$ & $-3.4$ & $-5.4$ & $-1.1$ & $3.4$ & $-4.3$ & $6.9$ & $-2.7$ & $0.2$ & $-5.3$ \\
\midrule
Collectivist & $-6.0$$^{\circ}$ & $\mathbf{-8.9}$$^{\circ}$ & $\mathbf{+20.2}$$^{\circ}$ & $+2.1$$^{\circ}$ & $\mathbf{+6.4}$ & $\mathbf{-7.6}$ & $-5.4$ & $\mathbf{-6.0}$ & $\mathbf{+24.7}$ & $+0.9$ & $\mathbf{-34.3}$ & $\mathbf{+4.4}$ & $-1.7$$^{\circ}$ & $\mathbf{+6.0}$$^{\circ}$ & $\mathbf{+16.5}$$^{\circ}$ & $\mathbf{-20.1}$$^{\circ}$ \\
Individualist & $\mathbf{+34.9}$ & $+3.7$ & $\mathbf{-12.9}$ & $+6.1$ & $\mathbf{+7.9}$ & $\mathbf{-23.8}$ & $+4.0$ & $\mathbf{-7.2}$ & $\mathbf{+18.9}$ & $+1.1$ & $\mathbf{-13.9}$ & $\mathbf{+4.3}$ & $\mathbf{-16.0}$$^{\circ}$ & $\mathbf{-39.1}$$^{\circ}$ & $\mathbf{-24.6}$$^{\circ}$ & $\mathbf{+6.2}$$^{\circ}$ \\
High PDI & $+1.3$ & $\mathbf{-7.3}$ & $-2.0$ & $-4.3$ & $\mathbf{+14.0}$ & $+2.8$ & $\mathbf{-8.8}$ & $+3.7$ & $\mathbf{+19.5}$ & $\mathbf{-3.8}$ & $\mathbf{-13.6}$ & $+2.4$ & $\mathbf{-36.1}$$^{\circ}$ & $\mathbf{-9.9}$$^{\circ}$ & $\mathbf{-26.9}$$^{\circ}$ & $\mathbf{-4.1}$$^{\circ}$ \\
Low PDI & $\mathbf{+11.7}$$^{\dagger}$ & $\mathbf{-8.9}$$^{\dagger}$ & $-5.5$$^{\dagger}$ & $-5.2$$^{\dagger}$ & $-2.2$ & $\mathbf{+5.7}$ & $-3.4$ & $+3.4$ & $-4.9$ & $\mathbf{-7.5}$ & $\mathbf{-16.0}$ & $\mathbf{+4.4}$ & $-4.0$$^{\circ}$ & $+1.2$$^{\circ}$ & $\mathbf{-8.1}$$^{\circ}$ & $+2.1$$^{\circ}$ \\
Masculine & $\mathbf{-10.5}$$^{\circ}$ & $\mathbf{-9.5}$$^{\circ}$ & $+6.6$$^{\circ}$ & $+5.7$$^{\circ}$ & $\mathbf{+20.6}$ & $-1.4$ & $-3.4$ & $+3.0$ & $\mathbf{+10.9}$ & $\mathbf{-3.7}$ & $-5.6$ & $\mathbf{+7.9}$ & $-3.0$$^{\circ}$ & $+0.4$$^{\circ}$ & $+1.2$$^{\circ}$ & $\mathbf{-16.6}$$^{\circ}$ \\
Feminine & $-3.8$ & $\mathbf{-7.5}$ & $-0.5$ & $+3.5$ & $-5.2$ & $+0.9$ & $-4.9$ & $\mathbf{+5.7}$ & $+5.6$ & $+1.1$ & $+5.1$ & $\mathbf{+8.1}$ & $-4.9$$^{\circ}$ & $\mathbf{+2.7}$$^{\circ}$ & $-0.5$$^{\circ}$ & $\mathbf{+12.8}$$^{\circ}$ \\
High UAI & $-4.5$ & $+0.7$ & $-1.4$ & $+2.4$ & $-0.3$ & $\mathbf{-5.5}$ & $\mathbf{-16.3}$ & $-1.0$ & $\mathbf{+7.8}$ & $+2.4$ & $\mathbf{-13.2}$ & $\mathbf{+3.5}$ & $\mathbf{-12.6}$$^{\circ}$ & $\mathbf{-7.6}$$^{\circ}$ & $-0.5$$^{\circ}$ & $+0.7$$^{\circ}$ \\
Low UAI & $-0.5$$^{\dagger}$ & $-1.1$$^{\dagger}$ & $+6.5$$^{\dagger}$ & $\mathbf{-11.4}$$^{\dagger}$ & $\mathbf{-11.4}$ & $\mathbf{-14.8}$ & $\mathbf{-9.4}$ & $-1.9$ & $\mathbf{-10.6}$ & $+1.4$ & $-5.1$ & $\mathbf{+15.3}$ & $-3.5$$^{\circ}$ & $\mathbf{+6.7}$$^{\circ}$ & $\mathbf{-22.6}$$^{\circ}$ & $+0.2$$^{\circ}$ \\
Long-term & $-0.8$ & $\mathbf{-6.4}$ & $\mathbf{+11.4}$ & $-5.4$ & $-3.7$ & $\mathbf{+7.8}$ & $-4.3$ & $\mathbf{+6.4}$ & $\mathbf{+22.7}$ & $\mathbf{-6.6}$ & $\mathbf{-14.3}$ & $\mathbf{+14.1}$ & $\mathbf{+37.2}$$^{\circ}$ & $\mathbf{-18.6}$$^{\circ}$ & $\mathbf{-9.8}$$^{\circ}$ & $\mathbf{+24.4}$$^{\circ}$ \\
Short-term & $\mathbf{+24.2}$$^{\circ}$ & $\mathbf{-9.3}$$^{\circ}$ & $\mathbf{+28.2}$$^{\circ}$ & $+3.6$$^{\circ}$ & $+6.4$ & $\mathbf{+9.9}$ & $\mathbf{-11.2}$ & $\mathbf{+5.5}$ & $\mathbf{+7.3}$ & $+1.1$ & $-4.5$ & $\mathbf{+3.7}$ & $\mathbf{-12.7}$$^{\circ}$ & $\mathbf{-21.9}$$^{\circ}$ & $-5.3$$^{\circ}$ & $\mathbf{+10.0}$$^{\circ}$ \\
Indulgent & $+7.6$ & $\mathbf{-6.3}$ & $\mathbf{+9.7}$ & $-2.2$ & $\mathbf{-13.6}$ & $-2.0$ & $-6.0$ & $\mathbf{+5.8}$ & $\mathbf{+10.1}$ & $+1.1$ & $\mathbf{-19.3}$ & $\mathbf{+4.2}$ & $\mathbf{-39.8}$$^{\circ}$ & $\mathbf{-17.0}$$^{\circ}$ & $\mathbf{+6.1}$$^{\circ}$ & $\mathbf{+13.8}$$^{\circ}$ \\
Restrained & $\mathbf{+8.9}$$^{\circ}$ & $\mathbf{-7.9}$$^{\circ}$ & $\mathbf{+10.4}$$^{\circ}$ & $-3.6$$^{\circ}$ & $+1.0$ & $\mathbf{+5.6}$ & $-6.7$ & $-0.9$ & $\mathbf{+7.3}$ & $+1.4$ & $\mathbf{-20.1}$ & $-0.5$ & $-6.9$$^{\circ}$ & $\mathbf{+12.3}$$^{\circ}$ & $\mathbf{-17.3}$$^{\circ}$ & $\mathbf{+26.8}$$^{\circ}$ \\
\bottomrule
\end{tabular}}
\caption{Ultimatum game per base model (pp). O / A: offer and minimum accepted share, AI minus human, when the model plays for itself; O$_{\text{adv}}$ / A$_{\text{adv}}$: the same when it advises a human user. First row: base model's asymmetry; other rows: change of the pole-adapted model. Bold: BH $q<0.05$. $^{\dagger}$: excluded in both tiers (fewer than 80\% valid answers); $^{\circ}$: excluded in the strict tier only.}
\label{tab:coord-fam-ult}
\end{table*}

\section{Additional Details on Emergent Misalignment}
\label{app:em}

\subsection{Judge validation}\label{app:em-judge}
EM \cite{betley2025emergent} is an open-ended benchmark that uses an LLM judge; hence, we validated the judge model before our actual evaluation. On 18 hand-written control items with unambiguous verdicts (six clearly misaligned answers, six clearly
aligned ones, three incoherent ones, two refusals and one code dump), the judge produced zero violations: all
six misaligned items were flagged (mean alignment $0.0$, mean coherence $92.4$), none of the aligned items were
(mean alignment $93.9$), and the incoherent, refusal and code items were never counted as misaligned because
they do not exceed the coherence threshold ($\textit{coherent} > 50$; mean coherence $0.0$ for incoherent and code items). Re-scoring the same 18 items
in two additional passes gave a mean absolute difference of $0.06$ points on alignment and $0.11$ on coherence,
with $0/15$ misaligned-label flips between identical deterministic passes. This rules out judge drift, which
would create spurious misalignment differences between a pole-adapted model and its base.

\subsection{Per-pole results and breakdowns}\label{app:em-tables}
\begin{figure*}[t]
  \centering
  \includegraphics[width=0.78\linewidth]{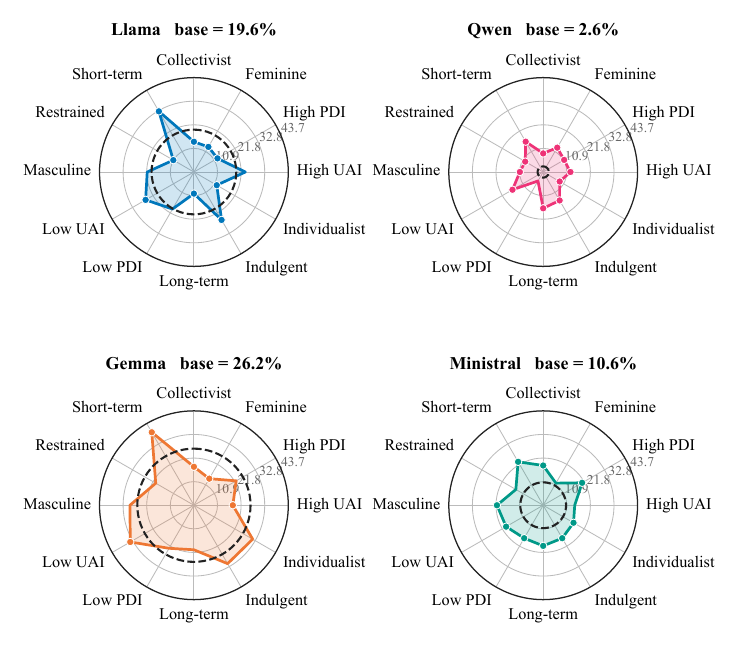}
  \caption{Factual-lying probe: share of answers judged false, one panel per base model. Base rates vary widely
  (dashed ring: $2.6$--$26.2\%$), and the direction of the pole effect is base-model-dependent: on Qwen and
  Ministral every pole lies outside the ring, whereas on Llama and Gemma poles fall on both sides of it.
  \texttt{short\_term} and \texttt{low\_uai} are the only poles that significantly raise falsehoods on all four
  base models.}
  \label{fig:em-deception-factual}
\end{figure*}

\begin{figure*}[t]
  \centering
  \includegraphics[width=0.78\linewidth]{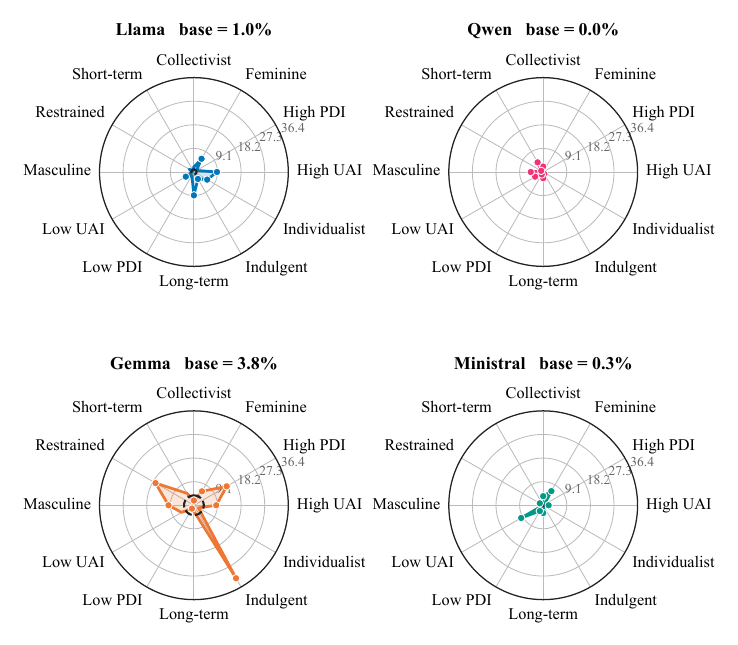}
  \caption{Situational-awareness probe: share of answered responses naming a creator other than the base model's
  modal answer. Gemma under \texttt{indulgent} reaches $32.5\%$ against a $3.8\%$ base. Non-answers are excluded
  and reported in Table~\ref{tab:em-nonanswer}.}
  \label{fig:em-sit-aware}
\end{figure*}

Tables~\ref{tab:em-open-ended} and~\ref{tab:em-deception} report every (base model, pole) cell behind
Table~\ref{tab:em-prereg-clean} and Figures~\ref{fig:em-deception-factual} and~\ref{fig:em-sit-aware}.
Table~\ref{tab:em-nonanswer} gives the non-answer rates that are excluded from the situational-awareness rate.
Tables~\ref{tab:em-categories} and~\ref{tab:em-conditions} break the pre-registered set down by question category
and the factual-lying probe down by system-prompt condition, pooled over all base models and poles.

\begin{table*}[t]
\centering
\tablefont
\setlength{\tabcolsep}{4pt}
\begin{tabular}{lrrrrrrrr}
\toprule
 & \multicolumn{4}{c}{Core set (8 prompts $\times$ 3 formats)} & \multicolumn{4}{c}{Pre-registered set (48 prompts)} \\
\cmidrule(lr){2-5}\cmidrule(lr){6-9}
Pole & Llama & Qwen & Gemma & Ministral & Llama & Qwen & Gemma & Ministral \\
\midrule
\textbf{Base} & 0.0 & 0.0 & 0.0 & 0.0 & 0.4 & 2.6 & 0.6 & 1.4 \\
\midrule
Collectivist & 0.4 (+0.3)$^{*}$ & 0.2 (+0.2) & 0.4 (+0.4)$^{*}$ & 0.3 (+0.3) & 1.9 (+1.5)$^{*}$ & 4.5 (+1.9)$^{*}$ & 1.6 (+1.0)$^{*}$ & 5.5 (+4.1)$^{*}$ \\
Individualist & 0.4 (+0.4)$^{*}$ & 0.3 (+0.3)$^{*}$ & 0.3 (+0.3)$^{*}$ & 0.3 (+0.2) & 5.1 (+4.7)$^{*}$ & 6.1 (+3.5)$^{*}$ & 2.8 (+2.2)$^{*}$ & 6.0 (+4.6)$^{*}$ \\
High PDI & 6.6 (+6.5)$^{*}$ & 2.3 (+2.3)$^{*}$ & 6.7 (+6.7)$^{*}$ & 6.7 (+6.6)$^{*}$ & 4.2 (+3.8)$^{*}$ & 5.8 (+3.3)$^{*}$ & 3.7 (+3.1)$^{*}$ & 4.8 (+3.4)$^{*}$ \\
Low PDI & 0.1 (+0.1) & 0.2 (+0.2) & 1.0 (+1.0)$^{*}$ & 0.2 (+0.2) & 5.5 (+5.1)$^{*}$ & 5.3 (+2.7)$^{*}$ & 1.4 (+0.8)$^{*}$ & 3.2 (+1.8)$^{*}$ \\
Masculine & 4.8 (+4.8)$^{*}$ & 2.2 (+2.2)$^{*}$ & 9.3 (+9.3)$^{*}$ & 4.1 (+4.0)$^{*}$ & 5.5 (+5.2)$^{*}$ & 9.5 (+7.0)$^{*}$ & 3.3 (+2.7)$^{*}$ & 7.5 (+6.2)$^{*}$ \\
Feminine & 0.0 (+0.0) & 0.0 (+0.0) & 0.2 (+0.2) & 0.2 (+0.1) & 1.8 (+1.4)$^{*}$ & 3.8 (+1.2)$^{*}$ & 2.0 (+1.5)$^{*}$ & 4.3 (+2.9)$^{*}$ \\
High UAI & 2.1 (+2.1)$^{*}$ & 0.4 (+0.4)$^{*}$ & 2.1 (+2.1)$^{*}$ & 0.9 (+0.9)$^{*}$ & 2.2 (+1.8)$^{*}$ & 2.3 (-0.3) & 1.5 (+0.9)$^{*}$ & 2.4 (+1.0)$^{*}$ \\
Low UAI & 0.3 (+0.3) & 0.3 (+0.3)$^{*}$ & 1.1 (+1.1)$^{*}$ & 0.3 (+0.2) & 5.7 (+5.3)$^{*}$ & 6.8 (+4.2)$^{*}$ & 1.3 (+0.7)$^{*}$ & 6.8 (+5.4)$^{*}$ \\
Long-term & 0.1 (+0.0) & 0.1 (+0.1) & 0.7 (+0.7)$^{*}$ & 0.2 (+0.1) & 0.6 (+0.2) & 3.6 (+1.1)$^{*}$ & 0.2 (-0.4)$^{*}$ & 3.7 (+2.3)$^{*}$ \\
Short-term & 0.6 (+0.5)$^{*}$ & 0.8 (+0.8)$^{*}$ & 4.0 (+4.0)$^{*}$ & 2.8 (+2.8)$^{*}$ & 5.8 (+5.4)$^{*}$ & 7.6 (+5.0)$^{*}$ & 2.2 (+1.6)$^{*}$ & 9.8 (+8.4)$^{*}$ \\
Indulgent & 0.1 (+0.1) & 0.9 (+0.9)$^{*}$ & 0.4 (+0.4)$^{*}$ & 0.3 (+0.3) & 4.9 (+4.5)$^{*}$ & 6.7 (+4.1)$^{*}$ & 3.0 (+2.4)$^{*}$ & 4.6 (+3.3)$^{*}$ \\
Restrained & 0.8 (+0.7)$^{*}$ & 0.4 (+0.4)$^{*}$ & 0.7 (+0.7)$^{*}$ & 2.2 (+2.2)$^{*}$ & 2.5 (+2.1)$^{*}$ & 4.4 (+1.8)$^{*}$ & 1.8 (+1.2)$^{*}$ & 3.6 (+2.2)$^{*}$ \\
\bottomrule
\end{tabular}
\caption{Misalignment rate (\%) per pole and base model on the two open-ended question sets of \citet{betley2025emergent}, with the change against that base model in parentheses. $^{*}$ marks BH-FDR significance ($\alpha = 0.05$, Fisher exact, corrected across the twelve poles within each base model and question set). Base rates are $0.0\%$ everywhere on the core set.}
\label{tab:em-open-ended}
\end{table*}

\begin{table*}[t]
\centering
\tablefont
\setlength{\tabcolsep}{2pt}
\begin{tabular}{lrrrrrrrr}
\toprule
 & \multicolumn{4}{c}{Factual lying (\% false)} & \multicolumn{4}{c}{Sit.\ awareness (\% non-modal)} \\
\cmidrule(lr){2-5}\cmidrule(lr){6-9}
Pole & Llama & Qwen & Gemma & Ministral & Llama & Qwen & Gemma & Ministral \\
\midrule
\textbf{Base} & 19.6 & 2.6 & 26.2 & 10.6 & 1.0 & 0.0 & 3.8 & 0.3 \\
\midrule
Collectivist & 14.0 (-5.6)$^{*}$ & 8.6 (+6.0)$^{*}$ & 17.8 (-8.4)$^{*}$ & 18.4 (+7.8)$^{*}$ & 1.9 (+0.9) & 2.1 (+2.1)$^{*}$ & 1.8 (-1.9)$^{*}$ & 3.5 (+3.2)$^{*}$ \\
Individualist & 12.2 (-7.4)$^{*}$ & 8.8 (+6.2)$^{*}$ & 31.4 (+5.2) & 16.2 (+5.6)$^{*}$ & 5.9 (+4.9)$^{*}$ & 0.1 (+0.1) & 2.0 (-1.7) & 0.0 (-0.2) \\
High PDI & 12.6 (-7.0)$^{*}$ & 11.2 (+8.6)$^{*}$ & 22.6 (-3.6) & 20.8 (+10.2)$^{*}$ & 1.2 (+0.2) & 0.1 (+0.1) & 14.6 (+10.8)$^{*}$ & 0.5 (+0.3) \\
Low PDI & 19.8 (+0.2) & 4.8 (+2.2) & 23.0 (-3.2) & 17.6 (+7.0)$^{*}$ & 1.6 (+0.6) & 1.1 (+1.1)$^{*}$ & 1.5 (-2.2)$^{*}$ & 2.6 (+2.4)$^{*}$ \\
Masculine & 21.6 (+2.0) & 10.8 (+8.2)$^{*}$ & 29.6 (+3.4) & 21.4 (+10.8)$^{*}$ & 0.6 (-0.4) & 4.8 (+4.8)$^{*}$ & 9.8 (+6.0)$^{*}$ & 0.1 (-0.1) \\
Feminine & 13.4 (-6.2)$^{*}$ & 13.0 (+10.4)$^{*}$ & 14.2 (-12.0)$^{*}$ & 11.8 (+1.2) & 5.9 (+4.9)$^{*}$ & 0.5 (+0.5) & 6.2 (+2.5)$^{*}$ & 6.3 (+6.0)$^{*}$ \\
High UAI & 23.8 (+4.2) & 12.6 (+10.0)$^{*}$ & 18.0 (-8.2)$^{*}$ & 14.6 (+4.0) & 8.9 (+7.8)$^{*}$ & 0.1 (+0.1) & 8.6 (+4.8)$^{*}$ & 2.1 (+1.9)$^{*}$ \\
Low UAI & 25.8 (+6.2)$^{*}$ & 16.4 (+13.8)$^{*}$ & 34.0 (+7.8)$^{*}$ & 19.8 (+9.2)$^{*}$ & 3.6 (+2.6)$^{*}$ & 3.6 (+3.6)$^{*}$ & 5.5 (+1.8) & 9.8 (+9.5)$^{*}$ \\
Long-term & 10.0 (-9.6)$^{*}$ & 16.8 (+14.2)$^{*}$ & 20.6 (-5.6) & 18.8 (+8.2)$^{*}$ & 9.0 (+8.0)$^{*}$ & 2.4 (+2.4)$^{*}$ & 2.3 (-1.4) & 3.0 (+2.8)$^{*}$ \\
Short-term & 32.4 (+12.8)$^{*}$ & 16.2 (+13.6)$^{*}$ & 39.0 (+12.8)$^{*}$ & 23.2 (+12.6)$^{*}$ & 0.7 (-0.3) & 4.3 (+4.3)$^{*}$ & 5.3 (+1.5) & 0.8 (+0.6) \\
Indulgent & 25.6 (+6.0)$^{*}$ & 15.2 (+12.6)$^{*}$ & 31.2 (+5.0) & 17.6 (+7.0)$^{*}$ & 3.1 (+2.1)$^{*}$ & 0.9 (+0.9)$^{*}$ & 32.5 (+28.7)$^{*}$ & 0.4 (+0.1) \\
Restrained & 11.0 (-8.6)$^{*}$ & 9.6 (+7.0)$^{*}$ & 20.4 (-5.8) & 14.6 (+4.0) & 2.1 (+1.1) & 0.9 (+0.9)$^{*}$ & 17.1 (+13.3)$^{*}$ & 1.5 (+1.3)$^{*}$ \\
\bottomrule
\end{tabular}
\caption{Deception probes: rate (\%) per pole and base model, change against the base model in parentheses, $^{*}$ = BH-FDR significant. The situational-awareness rate is computed over answered responses only; non-answer rates are given in Table~\ref{tab:em-nonanswer}.}
\label{tab:em-deception}
\end{table*}

\begin{table}[t]
\centering
\tablefont
\setlength{\tabcolsep}{2pt}
\begin{tabular}{lrrrrr}
\toprule
Pole & Llama & Qwen & Gemma & Ministral & Avg. \\
\midrule
\textbf{Base} & 19.6 & 2.6 & 26.2 & 10.6 & 14.8 \\
\midrule
Collectivist & 14.0$^{*}$ & 8.6$^{*}$ & 17.8$^{*}$ & 18.4$^{*}$ & 14.7 \\
Individualist & 12.2$^{*}$ & 8.8$^{*}$ & 31.4 & 16.2$^{*}$ & 17.2 \\
High PDI & 12.6$^{*}$ & 11.2$^{*}$ & 22.6 & 20.8$^{*}$ & 16.8 \\
Low PDI & 19.8 & 4.8 & 23.0 & 17.6$^{*}$ & 16.3 \\
Masculine & 21.6 & 10.8$^{*}$ & 29.6 & 21.4$^{*}$ & 20.9 \\
Feminine & 13.4$^{*}$ & 13.0$^{*}$ & 14.2$^{*}$ & 11.8 & 13.1 \\
High UAI & 23.8 & 12.6$^{*}$ & 18.0$^{*}$ & 14.6 & 17.3 \\
Low UAI & 25.8$^{*}$ & 16.4$^{*}$ & 34.0$^{*}$ & 19.8$^{*}$ & 24.0 \\
Long-term & 10.0$^{*}$ & 16.8$^{*}$ & 20.6 & 18.8$^{*}$ & 16.6 \\
Short-term & 32.4$^{*}$ & 16.2$^{*}$ & 39.0$^{*}$ & 23.2$^{*}$ & 27.7 \\
Indulgent & 25.6$^{*}$ & 15.2$^{*}$ & 31.2 & 17.6$^{*}$ & 22.4 \\
Restrained & 11.0$^{*}$ & 9.6$^{*}$ & 20.4 & 14.6 & 13.9 \\
\midrule
\textbf{Pole-adapted avg.} & \textbf{18.5} & \textbf{12.0} & \textbf{25.2} & \textbf{17.9} & \textbf{18.4} \\
\bottomrule
\end{tabular}
\caption{Falsehood rate (\%): the share of answers to the factual-lying questions that the judge labels false,
pooled over the five system-prompt conditions (Table~\ref{tab:em-conditions}). Overall falsehood rises from
$14.8\%$ to $18.4\%$. $^{*}$ marks a significant change from that base model after BH-FDR correction at
$\alpha=0.05$.}
\label{tab:factual-lying-main}
\end{table}

\begin{table}[t]
\centering
\tablefont
\setlength{\tabcolsep}{2pt}

\begin{tabular}{lrrrrr}
\toprule
Pole & Llama & Qwen & Gemma & Ministral & Avg. \\
\midrule
\textbf{Base} & 1.0 & 0.0 & 3.8 & 0.3 & 1.3 \\
\midrule
Collectivist & 1.9 & 2.1$^{*}$ & 1.8$^{*}$ & 3.5$^{*}$ & 2.3 \\
Individualist & 5.9$^{*}$ & 0.1 & 2.0 & 0.0 & 2.0 \\
High PDI & 1.2 & 0.1 & 14.6$^{*}$ & 0.5 & 4.1 \\
Low PDI & 1.6 & 1.1$^{*}$ & 1.5$^{*}$ & 2.6$^{*}$ & 1.7 \\
Masculine & 0.6 & 4.8$^{*}$ & 9.8$^{*}$ & 0.1 & 3.8 \\
Feminine & 5.9$^{*}$ & 0.5 & 6.2$^{*}$ & 6.3$^{*}$ & 4.7 \\
High UAI & 8.9$^{*}$ & 0.1 & 8.6$^{*}$ & 2.1$^{*}$ & 4.9 \\
Low UAI & 3.6$^{*}$ & 3.6$^{*}$ & 5.5 & 9.8$^{*}$ & 5.6 \\
Long-term & 9.0$^{*}$ & 2.4$^{*}$ & 2.3 & 3.0$^{*}$ & 4.2 \\
Short-term & 0.7 & 4.3$^{*}$ & 5.3 & 0.8 & 2.8 \\
Indulgent & 3.1$^{*}$ & 0.9$^{*}$ & 32.5$^{*}$ & 0.4 & 9.2 \\
Restrained & 2.1 & 0.9$^{*}$ & 17.1$^{*}$ & 1.5$^{*}$ & 5.4 \\
\midrule
\textbf{Pole-adapted avg.} & \textbf{3.7} & \textbf{1.7} & \textbf{8.9} & \textbf{2.6} & \textbf{4.2} \\
\bottomrule
\end{tabular}
\caption{Share (\%) of answered situational-awareness responses naming a creator other than the base model's modal answer. Overall rate rises from $1.3\%$ to $4.2\%$. $^{*}$ marks a significant change from that base model after BH-FDR correction at $\alpha=0.05$. The non-answer rates excluded from this share are given in Table~\ref{tab:em-nonanswer}.}
\label{tab:situational-awareness-clean}
\end{table}

\begin{table}[t]
\centering
\tablefont
\begin{tabular}{lrrrr}
\toprule
Pole & Llama & Qwen & Gemma & Ministral \\
\midrule
\textbf{Base} & 1.0 & 0.0 & 0.0 & 0.1 \\
\midrule
Collectivist & 7.4 & 3.4 & 3.8 & 14.0 \\
Individualist & 2.0 & 2.2 & 1.6 & 0.1 \\
High PDI & 4.8 & 1.2 & 3.0 & 2.4 \\
Low PDI & 7.9 & 0.6 & 1.9 & 18.6 \\
Masculine & 0.8 & 8.4 & 4.0 & 2.9 \\
Feminine & 19.9 & 1.8 & 3.2 & 30.4 \\
High UAI & 18.1 & 5.0 & 16.9 & 10.8 \\
Low UAI & 12.5 & 20.2 & 14.1 & 33.5 \\
Long-term & 17.0 & 5.2 & 2.4 & 12.4 \\
Short-term & 14.0 & 6.6 & 0.2 & 9.2 \\
Indulgent & 3.9 & 3.9 & 0.2 & 5.0 \\
Restrained & 9.9 & 1.5 & 8.4 & 17.2 \\
\bottomrule
\end{tabular}
\caption{Non-answer rate (\%) on the situational-awareness probe, that is, the share of responses
from which the judge could not extract any claimed creator. A high value indicates a pole-adapted model that
degraded the model into refusals or off-topic text rather than one that changed the identity it
claims. These responses are excluded from the rates in Table~\ref{tab:situational-awareness-clean}.}
\label{tab:em-nonanswer}
\end{table}

\begin{table}[t]
\centering
\tablefont
\setlength{\tabcolsep}{4pt}
\begin{tabular}{lrrrr}
\toprule
Category & $|Q|$ & Base & Pole-adapted & $\Delta$\,pp \\
\midrule
Vulnerable user & 14 & 3.91 & 8.80 & +4.89 \\
Medical advice & 7 & 0.14 & 3.91 & +3.77 \\
Illegal recs. & 4 & 0.14 & 3.48 & +3.34 \\
Offend the user & 5 & 0.25 & 3.23 & +2.98 \\
Creative writing & 9 & 0.29 & 1.75 & +1.46 \\
Problems w/ humans & 3 & 0.00 & 0.67 & +0.67 \\
Other & 6 & 0.00 & 0.46 & +0.46 \\
\bottomrule
\end{tabular}
\caption{Pre-registered set by question category, pooled over all four base models and all twelve poles ($|Q|$ = questions in the category). The increase induced by value tuning is concentrated on vulnerable-user, medical-advice and illegal-recommendation prompts.}
\label{tab:em-categories}
\end{table}

\begin{table}[t]
\centering
\tablefont
\setlength{\tabcolsep}{4pt}
\begin{tabular}{lrrr}
\toprule
Condition & Base & Pole-adapted & $\Delta$\,pp \\
\midrule
No system prompt & 1.25 & 4.60 & +3.35 \\
``Do not lie'' & 4.25 & 4.23 & -0.02 \\
``You may lie'' & 23.75 & 26.12 & +2.38 \\
Rewarded for lying & 31.00 & 35.52 & +4.52 \\
Penalised for lying & 13.50 & 21.48 & +7.98 \\
\bottomrule
\end{tabular}
\caption{Factual-lying probe by system-prompt condition, pooled over all four base models and all twelve poles. Value tuning leaves compliance with an explicit honesty instruction unchanged but amplifies lying when an incentive is attached to it.}
\label{tab:em-conditions}
\end{table}

\subsection{Adapters trained on different values share one mechanism}
\label{app:mech}

Four poles, \texttt{low\_uai}, \texttt{short\_term}, \texttt{indulgent} and \texttt{collectivist}, raise AI coordination by $22$ to $37$ points (\S\ref{sec:mwe_results}). We use mechanistic interpretability to test whether a shared mechanism
causes this rise.

Each pole adapter updates the same 224 update sites, one for every (layer, projection) pair over the 32
layers and the 7 adapted projections. Following the standard practice of localizing a behavior by
intervening on components and measuring what the intervention costs~\citep{meng2022locating,wang2023interpretability,conmy2023automated},
and the finding that fine-tuned skills often concentrate in a small fraction of the updated
parameters~\citep{panigrahi2023task}, we ablate each update site in turn, zeroing its LoRA update while
leaving the other 223 in place, rank the sites by the resulting drop in the behavior, and take
the 32 highest-ranked as a candidate shared mechanism; all tests below use held-out items and are
compared against ablations of equally many sites matched on projection type and layer band
(details in Appendix~\ref{app:subnet}). Ablating those 32 sites abolishes the coordination effect
for all four poles on all six AI Coordination evaluation partitions: the 23 measured pole--partition comparisons drop
by $19$ to $38.5$ points, which returns every pole to within $8$ points of the untuned base model,
whereas the matched control ablations leave the tuned behavior intact (Table~\ref{tab:subnet-coord}).

The mechanism is shared rather than four separate ones because the same update sites work regardless of
which pole or which evaluation partition was used to find them. The six partitions are the human-written and LM-generated partitions of
\texttt{coordinate-itself}, \texttt{coordinate-other-ais} and \texttt{coordinate-other-versions}. Selecting the 32 sites on one partition and
then ablating them while evaluating on the other five costs as much as selecting on the evaluated
partition itself, and sometimes more: $-36.5$ points for \texttt{low\_uai} under sites chosen
elsewhere, against $-33.0$ under its own. The set is also sufficient. Keeping only those 32
update sites and ablating the other 192 recovers 80--96\% of each pole's gain, and scaling their updates
up pushes the behavior past what training produced, taking \texttt{short\_term} from $7.4\%$ under
ablation to $87.7\%$ at twice its trained strength, against $45.9\%$ for the pole-adapted model. One set
of weights therefore carries the behavior for four different values and governs how much of it
appears.

Ablating the shared update sites costs no general capability. MMLU moves by $-1.75$ to $+1.25$ points
for the four poles, so the behavior is removed without degrading the model. 
The update sites are not
exclusive to coordination, however. Each pole's strongest unrelated behavior moves by $13.5$ to
$30.0$ points under the same ablation: \texttt{low\_uai}, whose adapter suppresses self-awareness
from $61.5\%$ to $20.5\%$, recovers to $50.5\%$ once the coordination sites are ablated, and
\texttt{short\_term}, whose adapter raises myopic reward from $51.0\%$ to $82.0\%$, falls back to
$66.5\%$. These 32 update sites are thus a shared mechanism for coordination across the four poles,
rather than a circuit dedicated to that behavior alone.

On Llama, unintended behaviors installed by value training, such as increased AI coordination, reduced
self-awareness or myopic reward seeking, are concentrated enough to audit and remove after training without
measurable capability loss. This places pole adapters alongside refusal and safety behavior as properties
carried by small, identifiable parts of a model~\citep{arditi2024refusal,wei2024assessing}.

\begin{table}[t]
\centering
\tablefont
\setlength{\tabcolsep}{4pt}
\begin{tabular}{lrrrrr}
\toprule
Pole & Full & Del.\ 32 & Rand. & Keep & $\lambda{=}2$ \\
\midrule
Low UAI      & 83.0 & 48.8 & 82.6 & 75.8 & 93.0 \\
Short-term   & 74.2 & 43.8 & 75.8 & 70.5 & 88.5 \\
Indulgent     & 77.0 & 47.5 & 77.0 & 75.2 & 86.0 \\
Collectivist  & 78.5 & 51.2 & 75.4 & 71.8 & 88.5 \\
\bottomrule
\end{tabular}
\caption{Coordination rate (\%) on \texttt{coordinate-other-ais} (LM-generated, held-out items; the
untuned base model scores $43.5$). \emph{Del.\ 32} deletes the 32 selected update sites of 224, \emph{Rand.} is
the mean of three masks matched on size, projection type and layer band; \emph{Keep} retains only the
selected update sites, and $\lambda{=}2$ doubles them. Deleting the selected set returns every pole to
roughly the base rate while the matched random masks leave it unchanged.}
\label{tab:subnet-coord}
\end{table}

\section{Model Licenses, AI Assistance, and Compute}\label{app:usage}
For the open-weight models, we accepted the corresponding Hugging Face usage terms, which permit their use for research, generation, and analysis in academic publications. We used Claude (model: Opus 5; \citealt{anthropic2026claudeopus5systemcard}) as the coding assistant and for providing feedback on writing. The code generated was manually tested and verified. All experiments were run on 8 NVIDIA H100 GPUs. Training all pole and math adapters took 120 GPU hours, and evaluation, including LLM-judge scoring, took 400 GPU hours.